\documentclass{article}

\usepackage{microtype}
\usepackage{graphicx}
\usepackage{booktabs} 
\usepackage{amsmath}
\usepackage{amsthm}
\newtheorem{theorem}{Theorem}
\newtheorem{lemma}[theorem]{Lemma}
\newtheorem{proposition}[theorem]{Proposition}
\usepackage{subfiles}
\usepackage{graphicx}
\usepackage{bbm}
\usepackage[disable]{todonotes}
\usepackage{subfig}
\newtheorem{prop}{Proposition}
\usepackage[toc,page]{appendix}
\newtheorem{corollary}{Corollary}[theorem]
\usepackage{float}
\usepackage{amsfonts}
\usepackage{caption}

\usepackage{xcite}
\usepackage{xr}
\usepackage{courier}

\usepackage[space]{grffile}

\usepackage[round]{natbib}   
\newcommand{\R}{\mathbb{R}}

\usepackage{mdframed}
\newmdtheoremenv{theo}{Theorem}
\newcommand{\new}[1]{\textcolor{black}{#1}}

\usepackage{hyperref}

\usepackage[accepted]{icml2020}

\icmltitlerunning{Preference Modeling with Context-Dependent Salient Features }

\begin{document}

\twocolumn[
\icmltitle{Preference Modeling with Context-Dependent Salient Features}



\icmlsetsymbol{equal}{*}

\begin{icmlauthorlist}
\icmlauthor{Amanda Bower}{math}
\icmlauthor{Laura Balzano}{eecs}
\end{icmlauthorlist}

\icmlaffiliation{math}{Department of Mathematics, University of Michigan, Ann Arbor, MI}
\icmlaffiliation{eecs}{Department of Electrical Engineering and Computer Science, University of Michigan, Ann Arbor, MI}

\icmlcorrespondingauthor{Amanda Bower}{amandarg@umich.edu}

\icmlkeywords{Machine Learning, ICML, Preference learning, Ranking}

\vskip 0.3in
]



\printAffiliationsAndNotice{}  

\begin{abstract}
We consider the problem of estimating a ranking on a set of items from noisy pairwise comparisons given item features. We address the fact that pairwise comparison data often reflects irrational choice, e.g. intransitivity. Our key observation is that two items compared in isolation from other items may be compared based on only a salient subset of features. Formalizing this framework, we propose the \textit{salient feature preference model} and prove a finite sample complexity result for learning the parameters of our model and the underlying ranking with maximum likelihood estimation. We also provide empirical results that support our theoretical bounds and illustrate how our model explains systematic intransitivity. Finally we demonstrate strong performance of maximum likelihood estimation of our model on both synthetic data and two real data sets: the UT Zappos50K data set and comparison data about the compactness of legislative districts in the US.

\end{abstract}

\section{Introduction}
The problem of estimating a ranking is ubiquitous and has applications in a wide variety of areas such as recommender systems, review of scientific articles or proposals, search results, sports tournaments, and understanding human perception. Collecting full rankings of $n$ items from human users is infeasible if the number of items $n$ is large. Therefore, $k$-wise comparisons, $k<n$, are typically collected and aggregated instead. 
Pairwise comparisons ($k=2$) are popular since it is believed that humans can easily and quickly answer these types of comparisons. However, it has been observed that data from $k$-wise comparisons for small $k$ often exhibit what looks like irrational choice, such as systematic intransitivity among comparisons. Common models address this issue with modeling noise, ignoring its systematic nature. We observe, as others have before us \cite{seshadri2019discovering, rosenfeld2019predicting, DBLP:journals/corr/abs-1901-10860, kleinberg2017comparison, benson2016relevance, chen2016predicting, Chen:2016:MIM:2835776.2835787}, that these systematic irrational behaviors can likely be better modeled as \emph{rational behaviors made in context}, meaning that the particular $k$ items used in a $k$-wise comparison will affect the comparison outcome.

\begin{figure}[!htb]
\minipage{0.15\textwidth}
\centering
District $A$
  \includegraphics[width=\linewidth]{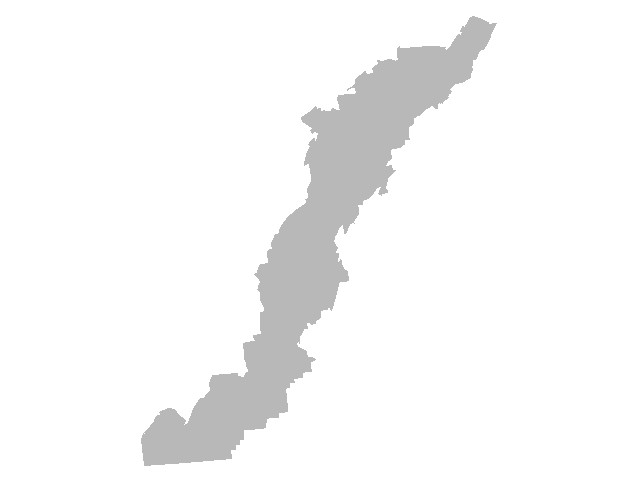}
\endminipage\hspace{\fill}
\minipage{0.15\textwidth}
\centering
District $B$
  \includegraphics[width=\linewidth]{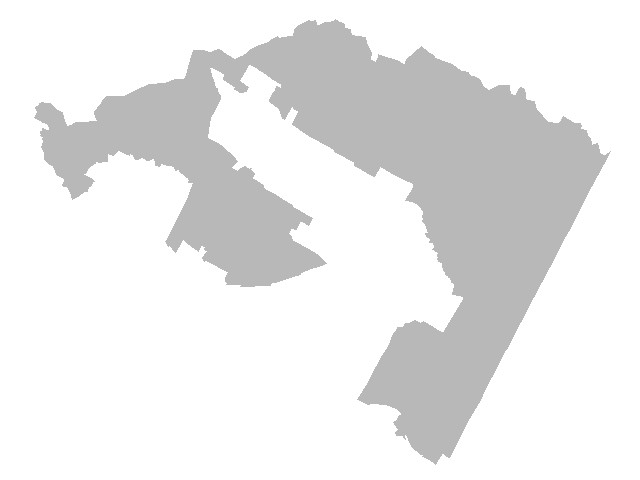}
\endminipage\hspace{\fill}
\minipage{0.15\textwidth}%
\centering
District $C$
  \includegraphics[width=\linewidth]{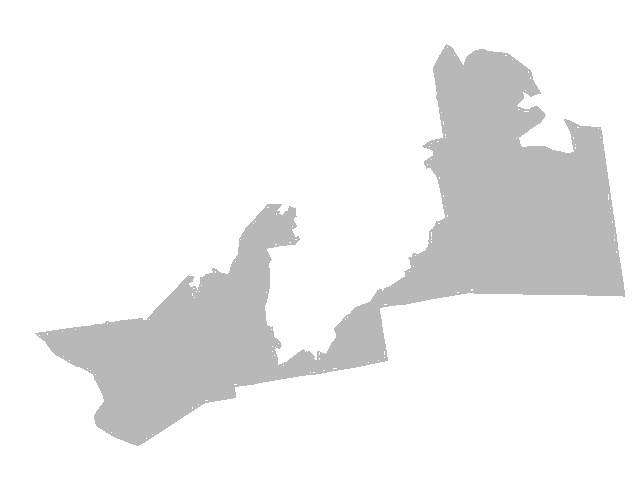}
\endminipage
\caption{Three districts used in pairwise comparison tasks in \cite{districtCompactness}}
\label{fig:districts}
\end{figure}

Consider the most common model for learning a single ranking from pairwise comparisons, the Bradley-Terry-Luce (BTL) model. In this model, there exists a judgment vector $w^* \in \R^d$ that indicates the favorability of each of the $d$ features of an item (e.g. for shoes: cost, width, material quality, etc), and each item has an embedding $U_i \in \R^d$, $i=1,\dots,n$, indicating the value of each feature for that given item. Subsequently, the outcome of a comparison is made with probability related to the inner product $\langle U_i, w^* \rangle$; the larger this inner product, the more likely item $i$ will be ranked above other items to which it is compared. A key implicit assumption is that the features used to rank all $n$ items are the same features used to rank just $k$ items in the absence of the other $n-k$ items. 
However, 
we argue that the context of that particular pairwise comparison is also relevant; it is likely that 
when a pairwise comparison is collected, if there are a small number of features that ``stand out,'' a person will use only these features and ignore the rest when he or she makes a comparison judgment. Otherwise, if there are no salient features between a pair of items, a person will take all features into consideration. This theory has been hypothesized by the social science community to explain violations of rational choice \cite{tversky1972elimination, tversky1993context, rieskamp2006extending, brown2009enquiry, shepard1964attention, torgerson1965multidimensional, tversky1977features, bordalo2013salience}. 
For example, \cite{districtCompactness} collected preference data to understand human perception of the compactness of legislative districts. They hypothesized that the features respondents use in a pairwise comparison task to judge district compactness vary from pair to pair, which explains why their data are more reliable for larger $k$. To illustrate this point, we highlight a concrete example from their experiments. Given two images of districts, they asked respondents to pick which district is more compact. When comparing district $A$ with district $B$ or district $C$ in Figure \ref{fig:districts}, one of the most salient features is the degree of nonconvexity. However, when comparing district $B$ and district $C$, the degree of nonconvexity is no longer a salient feature. These districts look similar on many dimensions, forcing a person to really think and consider all the features before making a judgment.
Let $P_{ij}$ be the empirical probability that district $i$ beats district $j$ with respect to compactness. Then, from the experiments of \cite{districtCompactness}, we have $P_{AB} = 100\%$, $P_{BC} = 67\%$, and $P_{AC} = 70\%$. These three districts violate strong stochastic transitivity, the requirement that if $P_{AB} \geq 50\%$ and $P_{BC} \geq 50\%$, then $P_{AC} \geq \max\{P_{AB},P_{BC}\}$.

We propose a novel probabilistic model called the \textit{salient feature preference model} for pairwise comparisons such that the features used to compare two items are dependent on the context in which two items are being compared. The salient feature preference model is a variation of the standard Bradley-Terry-Luce model. At a high level, given a pair of items in $\R^d$, we posit that humans perform the pairwise comparison in a coordinate subspace of $\R^d$. The particular subspace depends on the salience of each feature of the pairs being compared. Crucially, if any human were able to rank all the items at once, he or she would compare the items in the ambient space without projection onto a smaller subspace. This single ranking in the ambient space 
is the ranking that we would like to estimate. Our contributions are threefold. First, we precisely formulate this model and derive the associated maximum likelihood estimator (MLE) where the log-likelihood is convex. Our model can result in intransitive preferences, despite the fact that comparisons are based off a single universal ranking. In addition, our model generalizes to unseen items and unseen pairs. Second, we then prove a necessary and sufficient identifiability condition for our model and finite sample complexity bounds for the MLE. 
Our result specializes to the sample complexity of the MLE for the BTL model with features, which to the best of our knowledge has not been provided in the literature. Third, we provide synthetic experiments that support our theoretical results and also illustrate scenarios where our salient feature preference model results in systematic intransitives. We also demonstrate the efficacy of our model and maximum likelihood estimation on real preference data about legislative district compactness and the \texttt{UT Zappos50K} data set.

\subsection{Related Work}
\paragraph{The Bradley-Terry-Luce Model}
One popular probabilistic model for pairwise comparisons is the Bradley-Terry-Luce (BTL) model \cite{bradley1952rank, luce2012individual}. In this model, there are $n$ items each with an unknown utility $u_i$ for $i \in [n]$, and the items are ranked by sorting the utilities. The BTL model defines
\begin{align}
\label{eq:BTL}
    \mathbb{P}(\text{item } i \text{ beats item } j) = \frac{e^{u_i}}{e^{u_i}+e^{u_j}}.
\end{align}

Although the BTL model makes strong parametric assumptions, it has been analyzed extensively by both the machine learning and social science community and has been applied in practice. For instance, the World Chess Federation has used a variation of the BTL model in the past for ranking chess players \cite{menke2008bradley}. The sample complexity of learning the utilities or the ranking of the items with maximum likelihood estimation (MLE) has been studied recently in \cite{pmlr-v32-rajkumar14, rankCentrality}. Moreover, there is a recent line of work that analyzes the sample complexity of learning the utilities with MLE and other algorithms under several variations of the BTL model, including when the items have features that may or may not be known \cite{NIPS2018_7746, oh2015collaboratively, lu2015individualized, park2015preference, fbtl, niranjan2017inductive}. Our model is also a variation of the BTL model where the utility of each item is dependent on the items it is being compared to.

\paragraph{Violations of Rational Choice}
The social science community has long recognized and hypothesized about irrational choice \cite{shepard1964attention, torgerson1965multidimensional, tversky1977features, tversky1972elimination, bordalo2013salience}. See \cite{rieskamp2006extending} for an excellent survey of this area including references to social science experiments that demonstrate scenarios where humans make choices that can violate a variety of rational choice axioms such as transitivity.
There has been recent progress in modeling and providing evidence for violations of rational choice axioms in the machine learning community \cite{seshadri2019discovering, rosenfeld2019predicting, heckel2019active, DBLP:journals/corr/abs-1901-10860, kleinberg2017comparison, shah2017simple, ragain2016pairwise, niranjan2017inductive, benson2016relevance, chen2016predicting, Chen:2016:MIM:2835776.2835787, Rajkumar:2015:RSP:3045118.3045190, dehui, agresti2012categorical}. In contrast to our work, none of these works model preference data that both violates rational choice and admits a universal ranking of the items with the exception of \cite{shah2017simple,heckel2019active}. Assuming there is a true ranking of the items, our model makes a direct connection between pairwise comparison data that violates rational choice and the underlying ranking. Violations of rational choice, including intransitivty, occur in our model because of contextual effects due to which pairs of items are being compared. These contextual effects distort the true ranking, whereas in the work of \cite{shah2017simple,heckel2019active} the intransitive choices define the ranking. Specifically, the items are ranked by sorting the items by the probability that an item beats any other item.

We now focus on the works most similar to ours. The work in \cite{seshadri2019discovering}, which generalizes \cite{chen2016predicting, Chen:2016:MIM:2835776.2835787} from pairwise comparisons to $k$-wise comparisons, considers a model for context dependent comparisons. However, because they do not assume access to features, their model cannot predict choices based on new items, which is a key task for very large modern data sets. In contrast, our model can predict pairwise outcomes and rankings of new items. Both \cite{rosenfeld2019predicting} and \cite{DBLP:journals/corr/abs-1901-10860} assume access to features of items and propose learning contextual utilities with neural networks. In contrast, we propose a linear approach with typically far fewer parameters to estimate. Furthermore, the latter work does not contain any theory, whereas we prove a sample complexity result on estimating the parameters of our model. In all of the aforementioned works in this paragraph, the resulting optimization problems are non-convex with the exception of a special case in \cite{seshadri2019discovering} that requires sampling every pairwise comparison. In contrast, the negative log likelihood of our model is convex. Interestingly, the work in \cite{makhijani2019parametric} shows that for a class of parametric models for pairwise preference probabilities, if intransitives exist, then the negative log likelihood cannot be convex. Our model does not belong to the class of parametric models they consider.

\paragraph{Notation} For an integer $d>0$, $[d] := \{1, \dots, d\}$. For $x,y \in \mathbb{R}^d$, $\langle x, y \rangle := \sum_{i = 1}^d x_i y_i$. 
For $x \in \mathbb{R}^d$ and $\Omega \subset [d]$, let $x^{\Omega} \in \mathbb{R}^d$ where $(x^{\Omega})_i = x_i$ if $i \in \Omega$ and 0 otherwise. For $i,j \in [n]$, ``$i >_B j$'' means ``item $i$ beats item $j$.'' Let $\mathcal{P}(X)$ be the power set of a set $X$. Given a set of vectors $S = \{x_i \in \mathbb{R}^d \}_{i = 1}^q$, $\mathrm{span}(S) = \{ \sum_{i=1}^q \alpha_i x_i : \alpha_i \in \mathbb{R} \}$.

\section{Model and Algorithm}
\label{sec:modelFormulation}

\paragraph{Salient Feature Preference Model}

Suppose there are $n$ items, and each item $j \in [n]$ has a known feature vector $U_j \in \mathbb{R}^d$. Let $U := \begin{bmatrix} U_1 U_2 \cdots U_n \end{bmatrix} \in \mathbb{R}^{d \times n}$.
Let $w^* \in \mathbb{R}^d$ be the unknown \textit{judgment weights}, which signify the importance of each feature when comparing items. Let $\tau: [n] \times [n] \rightarrow \mathcal{P}([d])$ 
be the known \textit{selection function} that determines which features are used in each pairwise comparison. Let $P := \{(i,j) \in [n] \times [n]: i <j \}$ be the set of all pairs of items. 
Let $S_m = \{(i_\ell, j_\ell, y_\ell)\}_{\ell =1}^m $ be a set of $m$ independent pairwise comparison samples where $(i_\ell, j_\ell) \in P$ are chosen uniformly at random from $P$ with replacement, and $y_\ell \in \{0,1\}$ indicates the outcome of the pairwise comparison where 1 indicates item $i_\ell$ beat item $j_\ell$ and 0 indicates item $j_\ell$ beat item $i_\ell$. We model $y_\ell \sim \text{Bern}(\mathbb{P}(i_\ell >_B j_\ell))$ where
\begin{align}
\label{eq:thresholdBTL}
    \mathbb{P}(i_{\ell} >_B j_{\ell} ) = \frac{  \exp \left(\langle U_{i_{\ell}}^{\tau(i_{\ell},j_{\ell})} - U_{j_{\ell}}^{\tau(i_{\ell},j_{\ell})} , w^{*} \rangle \right)  }{ 1 + \exp \left(\langle U_{i_{\ell}}^{\tau(i_{\ell},j_{\ell})} - U_{j_
    \ell}^{\tau(i_{\ell},j_{\ell})}, w^{*} \rangle \right)}.
\end{align}
To understand the probability model given by Equation \eqref{eq:thresholdBTL}, note that $ \langle U_{i}^{\tau(i,j)}, w^{*} \rangle$ is the inner product of $U_i$ and $w^*$ after $U_i$ is projected to the coordinate subspace given by $\tau(i,j).$ Therefore, Equation \eqref{eq:thresholdBTL} is simply the utility model of Equation \eqref{eq:BTL} where the utilities are inner products computed in the subspace defined by the selection function $\tau$. If the selection function returns all the coordinates, i.e. $\tau(i,j) = [d]$, then Equation \eqref{eq:thresholdBTL} becomes the standard BTL model where the utility of item $i$ is $\langle U_i, w^*\rangle$ and fixed regardless of context, i.e., regardless of which pair is being compared. \new{This model is typically called ``BTL with features," and we will refer to it as FBTL.} See Section \ref{appendix:PLextension} in the Supplement for a natural extension of Equation \eqref{eq:thresholdBTL} to $k$-wise comparisons for $k>2$. Furthermore, we assume that the true ranking of all the items depends on all the features and is given by sorting the items by $\langle U_i, w^* \rangle$ for $i \in [n]$.

\paragraph{Selection Function}

We propose a selection function $\tau$ inspired by the social science literature, which posits that violations of rational choice axioms arise in certain scenarios because people make comparison judgments on a set of items based on the features that differentiate them the most \cite{rieskamp2006extending, brown2009enquiry, bordalo2013salience}.

For two variables $w, z \in \mathbb{R}$, let $\mu := (w+z)/2$ be their mean and $\bar{s} := ((w - \mu)^2 + (z - \mu)^2)/2$ be their sample variance. Given $t \in [d]$ and items $i, j \in [n]$, the \textit{top-$t$ selection function} selects the $t$ coordinates with the $t$ largest sample variances in the entries of the feature vectors $U_i$, $U_j$.

\paragraph{Algorithm: Maximum Likelihood Estimation}
Given observations $S_m = \{(i_\ell, j_\ell, y_\ell)\}_{\ell =1}^m $, item features $U \in \mathbb{R}^{d \times n}$, and a selection function $\tau$, the negative log-likelihood of $w \in \mathbb{R}^d$ is 

\begin{align}
    \mathcal{L}_{m}(w; U, S_m, \tau ) = \sum_{\ell = 1}^m \log\left(1+\exp\left( u_{i_{\ell},j_{\ell}} \right) \right) - y_\ell   u_{i_{\ell},j_{\ell}},
    \label{eq:logLoss}
\end{align}
where $u_{i_{\ell},j_{\ell}} = \left\langle w, U_{i_\ell}^{\tau(i_\ell,j_\ell)} - U_{j_\ell}^{\tau(i_\ell,j_\ell)} \right\rangle.$

Equation \ref{eq:logLoss} is equivalent to logistic regression with features $x_\ell = U_{i_\ell}^{\tau(i_{\ell},j_{\ell})} - U_{j_\ell}^{\tau(i_{\ell},j_{\ell})}$. See Section \ref{appendix:LLderiv} of the Supplement for the derivation. We estimate $w^*$ with the maximum likelihood estimator $\hat{w}$, which requires minimizing a convex function:
$\hat{w} := \text{argmin}_{w} \mathcal{L}_m(w; U, S_m, \tau).$



\section{Theory}
\label{sec:theory}
In this section, we analyze the sample complexity of estimating the judgment weights with the MLE given by minimizing $\mathcal{L}_m$ of Equation \eqref{eq:logLoss}. We first consider the sample complexity under an arbitrary selection function, and then specialize to two concrete selection functions: one that selects all features per pair and another that selects just one feature per pair. Throughout this section, we assume the set-up and notation presented in the beginning of Section \ref{sec:modelFormulation}.

First, the following proposition completely characterizes the identifiability of $w^*$. Identifiability means that with infinite samples, it is possible to learn $w^*$. Precisely, the salient feature preference model is identifiable if for all $(i,j) \in P$ and for $w_1, w_2 \in \mathbb{R}^d$, if $\mathbb{P}(i >_B j; w_1 ) = \mathbb{P}(i >_B j; w_2 )$, then $w_1 = w_2$ where $\mathbb{P}(i >_B j; w)$ refers to Equation \eqref{eq:thresholdBTL} where $w$ is the judgement vector. The proof is in Section \ref{appendix:identifiability} of the Supplement.
\begin{prop}[Identifiability]
\label{prop:identifiability}
Given item features $U \in \mathbb{R}^{n \times d}$, the salient feature preference model with selection function $\tau$ is identifiable if and only if $\text{span}\{ U_i^{\tau(i,j)} - U_j^{\tau(i,j)}: (i,j) \in P \} = \mathbb{R}^d$. 
\end{prop}

Now we present our main theorem on the sample complexity of estimating $w^*$. Let 
\[b^*: = \max_{(i,j) \in P} | \langle w^*, U_i^{\tau(i,j)} - U_j^{\tau(i,j)} \rangle |,\] which is the maximum absolute difference between two items' utilities when comparing them in context, i.e. based on the features given by the selection function $\tau$. Let \[\mathcal{W}(b^*) := \{ w \in \mathbb{R}^{d}: \max_{(i,j) \in P} | \langle w, U_i^{\tau(i,j)} - U_j^{\tau(i,j)} \rangle| \leq b^* \}.\] We constrain the MLE to $\mathcal{W}(b^*)$ so that we can bound the entries of the Hessian of $\mathcal{L}_m$ in our theoretical analysis. We do not enforce this constraint in our synthetic experiments.
\begin{theorem}[Sample complexity of learning $w^*$]
\label{thm:sampleComplexity}
Let $U \in \R^{d \times n}$, $w^* \in \R^d$, $\tau$, and $S_m$ be defined as in the beginning of Section \ref{sec:modelFormulation}. Let $\hat{w}$ be the maximum likelihood estimator, i.e. the minimum of $\mathcal{L}_m$ in Equation \eqref{eq:logLoss}, restricted to the set $\mathcal{W}(b^*)$. The following expectations are taken with respect to a uniformly chosen random pair of items from $P$. For $(i,j) \in P$, let 
\begin{align*} Z_{(i,j)} &:=(U_{i}^{\tau(i,j)} - U_{j}^{\tau(i,j)}) (U_{i}^{\tau(i,j)} - U_{j}^{\tau(i,j)})^T
\\
\lambda &:= \lambda_{\min}( \mathbb{E} Z_{(i,j)}),\\ 
\eta &:= \sigma_{\max}(\mathbb{E}((Z_{(i,j)} - \mathbb{E}Z_{(i,j)})^2)),\\
\zeta &:= \max_{(k,\ell) \in P} \lambda_{\max}(\mathbb{E} Z_{(i,j)  } - Z_{ (k,\ell)}),
\end{align*}
where for a positive semidefinite matrix $X$, $\lambda_{\min}(X)$ and $\lambda_{\max}(X)$ are the smallest/largest eigenvalues of $X$, and where for any matrix $X$, $\sigma_{\max}(X)$ is the largest singular value of $X$. Let \begin{equation}
\beta:=\max_{(i,j) \in P} \| U_{i}^{\tau(i,j)} - U_{j}^{\tau(i,j)} \|_{\infty}. \label{eq:betadef}
\end{equation}

Let $\delta > 0$. If $\lambda>0$ and 
\begin{align*}
m \geq \max & \left\{C_1(\beta^2 d+\beta\sqrt{d}) \log(4d / \delta)\right., \\
& \left. C_2(\eta + \lambda \zeta)\frac{\log(2d / \delta)}{\lambda^2} \right\},    
\end{align*}
then with probability at least 
$1 - \delta$,
\begin{align*} 
\| w^* - \hat{w}\|_2 = O\left(\frac{\exp(b^*)}{\lambda}  \sqrt{\frac{(\beta^2 d+\beta\sqrt{d}) \log(4 d / \delta)}{m}}\right)
\end{align*} 

where $C_1, C_2$ are constants given in the proof and the randomness is from the randomly chosen pairs and the outcomes of the pairwise comparisons.
\end{theorem}

 We utilize the proof technique of Theorem 4 in \cite{rankCentrality}, which proves a similar result for the standard BTL model of Equation \eqref{eq:BTL}, i.e. when $U=I_{n \times n}$, the $n \times n$ identity matrix, $d=n$, and $\tau(i,j) =[d]$ for all $(i,j) \in P$. We modify the proofs for arbitrary $U$ and $d$.  \todo{regular btl is actually not a special subcase} See Section \ref{appendix:sampleComplexityProof} in the Supplement for the proof. 
 
 We now discuss the terms that appear in Theorem \ref{thm:sampleComplexity}. First, the $d \log(d / \delta)$ terms are natural since we are estimating $d$ parameters. Second, estimating $w^*$ well essentially requires inverting the logistic function. When $b^*$ is large, we need to invert the logistic function for pairwise probabilities that are close to 0 and 1. This is precisely the challenging regime, since a small change in probabilities results in a large change in the estimate of $w^*$, and thus we expect to require many samples to estimate $w^*$ when $b^*$ is large. The exponential dependence on $b^*$ is standard for this type of analysis and arises from the Hessian of $\mathcal{L}_m$. Third, $\eta$ and $\zeta$ arise from a matrix concentration bound applied to the Hessian of $\mathcal{L}_m$. Fourth,  $\lambda$ arises from the minimum eigenvalue of the Hessian of $\mathcal{L}_m$ in a neighborhood of $w^*$, which controls the convexity of $\mathcal{L}_m$. This type of dependence also appears in other state of the art finite sample complexity analyses \cite{negahban2012unified}. In addition, to better understand the role of $\lambda$, we present the following proposition whose proof is in Section \ref{appendix:lambdaIdentifiable} in the Supplement. Proposition \ref{prop:lambdaIdentifiability} shows that the requirement $\lambda>0$ in Theorem \ref{thm:sampleComplexity} is fundamental, because we would otherwise be unable to bound the estimation error for the non-identifiable part of $w^*$, i.e., the projection of $w^*$ onto the orthogonal complement of $\text{span}\{ U_i^{\tau(i,j)} - U_j^{\tau(i,j)}: (i,j) \in P \} = \mathbb{R}^d$.
\begin{prop}
\label{prop:lambdaIdentifiability}
 $\lambda > 0$ if and only if the salient feature preference model is identifiable.
\end{prop}

Finally, if one assumes $\lambda, \eta, \zeta, \beta, \exp(b^*)$ are $O(1)$, then $\Omega(d \log (d / \delta))$ samples are enough to guarantee the error is $O(1)$. However, as we will show in the corollaries, these parameters are not always $O(1)$, increasing the complexity. We point out that the combination of the features $U$ and the selection function $\tau$ is what dictates the parameters of Theorem \ref{thm:sampleComplexity}. For the top-$t$ selection function in particular, we plot $\lambda, \zeta, \eta, b^*, \beta$, the number of samples required by Theorem \ref{thm:sampleComplexity}, and the bound on the estimation error as a function of intransitivity rates in the Supplement in Section \ref{appendix:parameterPlot}, to provide further insight into these parameters. Since we envision practical selection functions will be dependent on the features themselves, further analysis is a challenging but exciting subject of future work.



For deterministic $U$, we now specialize our results to FBTL as well as to the case where a single feature is used in each comparison. The following corollaries provide insight into how a particular selection function $\tau$ impacts $\lambda$, $\eta$, and $\zeta$ and thus the sample complexity.

First, we consider FBTL. In this case, the selection function selects all the features in each pairwise comparison, so there cannot be intransitivities in the preference data. The following Corollary of Theorem \ref{thm:sampleComplexity} gives a simplified form for $\lambda$ and upper bounds $\zeta$ and $\eta$. The terms involving the conditioning of $UU^T$ are natural; since we make no assumption on $w^*$, if the feature vectors are concentrated in a lower dimensional subspace, estimation of $w^*$ will be more difficult. See Section \ref{appendix:specificSelectionFunctions} of the Supplement for the proof.
\begin{corollary}[Sample complexity for FBTL]
\label{coro:fullFeatures}
For the selection function $\tau$, suppose $|\tau(i,j)| = d$ for any $(i,j) \in P$. In other words, all the features are used in each pairwise comparison. Let $\nu := \max\{\max_{(i,j) \in P} \|U_i - U_j\|_2^2, 1\}$. Assume $n>d$. Without loss of generality, assume the columns of $U$ sum to zero: $\sum_{i = 1}^n U_i =0$. Let $\delta > 0$. Then,
\begin{align*}
\lambda &= \frac{n\lambda_{\min}(UU^T)}{\binom{n}{2}}, \\
\zeta &\leq \nu + \frac{n\lambda_{\max}(UU^T)}{\binom{n}{2}}, \text{ and } \\ 
\eta &\leq \frac{\nu n \lambda_{\max}(UU^T)}{\binom{n}{2}} + \frac{n^2 \lambda_{\max}(UU^T)^2}{\binom{n}{2}^2}.
\end{align*}

Hence, if 
\begin{align*} m \geq \max & \left\{ C_1(\beta^2 d+\beta\sqrt{d}) \log(4d / \delta) ,  \right. \\
 & \left. C_3 \log(2d / \delta) \nu n \bar\lambda \right\}
\end{align*}

where
$$\bar \lambda = \left( \frac{\lambda_{\max}(UU^T) + \lambda_{\max}(UU^T)^2 +  \lambda_{\min}(UU^T)}{\lambda_{\min}(UU^T)^2} \right)$$


then with probability at least $1 - \delta$, \[\| w^* - \hat{w}\|_2 = O\left(\frac{\exp(b^*) n}{\lambda_{\min}(UU^T) } \sqrt{\frac{(\beta^2 d+\beta\sqrt{d}) \log( \frac{4d}{\delta})}{m}}\right)\]  
where $C_1$ and $C_3$ are constants given in the proof.
\end{corollary}
\todo{what if U is sampled from Gaussian? you need to control the variance because if $U$ gets too big then $b^*$ gets too big}
\todo{add in $R$, and $C$, and if $U$ is drawn from a gaussian}

To the best of our knowledge, this is the first analysis of the sample complexity for the MLE of FBTL parameters. There are related results in \cite{fbtl, negahban2012unified, heckel2019active, shah2017simple} to which our bound compares favorably, and we discuss this in Section \ref{sec:boundcompare} of the Supplement.

Second, suppose the selection function is very aggressive and selects only one coordinate for each pair, i.e. $|\tau(i,j)|=1$ for all $(i,j) \in P$. For instance, the top-$1$ selection function has this property. This type of selection function can cause intransitivities in the preference data as we show in the synthetic experiments of Section \ref{sec:syntheticExperiments}. 

\begin{corollary}
\label{coro:MaxCoordCoro}
Assume that for any $(i,j) \in P$, $|\tau(i,j)| = 1$. Partition $P = \sqcup_{k=1}^d P_k$ into $d$ sets where $(i,j) \in P_k$ if $\tau(i, j) = \{k\}$ for $k \in [d]$. Let $\beta$ be defined as in Theorem \ref{thm:sampleComplexity} and \[ \epsilon:= \min_{(i,j) \in P} \|U_i^{\tau(i,j)} - U_j^{\tau(i,j)}\|_{\infty}.\] Let $\delta > 0$.  Then
\begin{align*}
\lambda &\geq \frac{\epsilon^2}{ \binom{n}{2}}  \min_{k \in[d]} |P_k|,  \\
\zeta &\leq  \beta^2 + \frac{\beta^2}{ \binom{n}{2}} \max_{k \in [d]} |P_k|, \text{ and } \\
\eta &\leq \frac{\beta^4}{ \binom{n}{2}} \max_{k \in [d]} \left( |P_k| + \frac{|P_k|^2}{\binom{n}{2}} \right).
\end{align*}

Hence, if


$$m \geq \max\left\{C_1(\beta^2 d+\beta\sqrt{d}) \log(4d / \delta) ,  C_4(Q_1 + Q_2)  \right\},$$ 

where 
\begin{align*}
Q_1 &= \left(\frac{ \beta^4}{\epsilon^4}\right)\frac{\binom{n}{2} \max_{k \in [d]} |P_k| + \max_{k \in [d]} |P_k|^2}{\min_{k \in [d]} |P_k|^2}, \\
Q_2 &= \left(\frac{\beta^2}{\epsilon^2}\right)\frac{\binom{n}{2}  + \max_{k \in [d]} |P_k|}{\min_{k \in [d]} |P_k|},    
\end{align*}


then with probability at least $1 - \delta$, \[\| w^* - \hat{w}\|_2 = O\left(\frac{\exp(b^*) \binom{n}{2}}{ \epsilon^2 \underset{{k \in[d]}}{\min} |P_k| } \sqrt{\frac{(\beta^2 d+\beta\sqrt{d}) \log(\frac{4d}{\delta})}{m}}\right)\] 
where $C_1$ and $C_4$ are constants given in the proof.


\end{corollary}

 There are two main implications of Corollary \ref{coro:MaxCoordCoro} if we consider $\beta$ and $\epsilon$ constant. First, suppose there is a coordinate $k \in [d]$ such that $|P_k| := |\{(i,j) \in P: \tau(i,j)=k\}|$ is small. Intuitively it will take many samples to estimate $w^*$ well, since the chance of sampling a pairwise comparison that uses the $k$-th coordinate of $w^*$ is $|P_k| / \binom{n}{2}$. Corollary \ref{coro:MaxCoordCoro} formalizes this intuition. In particular, $\lambda=O(|P_k| / \binom{n}{2} )$, and since $\lambda$ comes into the bounds of Theorem \ref{thm:sampleComplexity} in the denominator of both the lower bound on samples and the upper bound on error, a small $\lambda$ makes estimation more difficult. 

Second, on the other hand, if $\epsilon$ is fixed, the maximum lower bound on $\lambda$ given by Corollary \ref{coro:MaxCoordCoro} is $ \max \min_{i \in [d]} |P_i| = \binom{n}{2} / d $ where the maximum is with respect to any partition of $P$. In this case, $|P_i| \approx |P_j|$ for all $i,j \in [d]$, so the chance of sampling a pairwise comparison that uses any coordinate is approximately equal. Therefore, $\lambda, \eta, \zeta = O(1/d)$, and by tightening a bound used in the proof of Theorem \ref{thm:sampleComplexity},  $\Omega(d^2\log(d / \delta))$ samples ensures the estimation error is $O(1)$. See Section \ref{sec:coroBoundTighten} in the Supplement for an explanation.

\todo{
\begin{itemize}
\item need to talk about how number of items affects things (maybe it does not if d =1, then lambda i guess is like constant? + add plots for this) and just interpret all these results
\end{itemize}
}


Ultimately, we seek to estimate the underlying ranking of the items. The following corollary of Theorem \ref{thm:sampleComplexity} says that by controlling the estimation error of $w^*$, the underlying ranking can be estimated approximately. 
The sample complexity depends inversely on the square of the differences of full feature item utilities. Intuitively, if the absolute difference between the utilities of two items is small, then many samples are required in order to rank these items correctly relative to each other. 
See Section \ref{appendix:coroRankingProof} in the Supplement for the proof.
\begin{corollary}[Sample complexity of estimating the ranking]
\label{coro:rankingComplexity}
Assume the set-up of Theorem \ref{thm:sampleComplexity}. Pick $k \in [\binom{n}{2}]$. Let $\alpha_k$ be the $k$-th smallest number in $\{ |\langle w^*, U_i - U_j \rangle|: (i,j) \in P \}$. Let $M := \max_{i \in [n]} \|U_i\|_2$. Let $\gamma^*: [n] \rightarrow [n]$ be the ranking obtained from $w^*$ by sorting the items by their full-feature utilities $\langle w^*, U_i\rangle$ where $\gamma^*(i)$ is the position of item $i$ in the ranking. Define $\hat{\gamma}$ similarly but for the estimated ranking obtained from the MLE estimate $\hat{w}$. Let $\delta > 0$. If 
\begin{align*}
m \geq \max & \left\{  C_1(\beta^2 d+\beta\sqrt{d}) \log(4d / \delta) ,\right. \\
& \left. C_2(\eta + \lambda \zeta)\frac{\log(2d / \delta)}{\lambda^2},\right. \\
& \left. \frac{C_5  M^2 e^{2b^*} (\beta^2 d+\beta\sqrt{d}) \log(4d / \delta)}{\alpha_k^2 \lambda^2} \right\},
\end{align*}
then with probability $1 - \delta$, \[K(\gamma^*,\hat{\gamma}) \leq k-1,\] where $K(\gamma^*, \hat{\gamma}) = |\{(i,j) \in P: (\gamma^*(i) - \gamma^*(j))(\hat{\gamma}(i) - \hat{\gamma}(j)) <0\}|$ is the Kendall tau distance between two rankings and $C_1$, $C_2$, and $C_5$ are constants given in the proof. 
\end{corollary}

\section{Experiments}
\label{sec:experiments}
See Sections \ref{appendix:algorithmImplementation}, \ref{appendix:district_experiments}, and \ref{appendix:zappos} of the Supplement for additional details about the algorithm implementation, data, preprocessing, hyperparameter selection, and training and validation error for both synthetic and real data experiments. 
\subsection{Synthetic Data}
We investigate violations of rational choice arising from the salient feature preference model and illustrate Theorem \ref{thm:sampleComplexity} while highlighting the differences between the salient feature preference model and the FBTL model throughout. Given the very reasonable simulation setup we use, these experiments suggest that the salient feature preference model may sometimes be better suited to real data than FBTL.

For these experiments, the ambient dimension $d=10$, the number of items $n = 100$, and comparisons are sampled from the salient feature preference model with top-$t$ selection function. The coordinates of $U$, respectively $w^*$, are drawn from 
$\mathcal{N}(0, 1/\sqrt{d})$, respectively
$\mathcal{N}(0, 4/\sqrt{d})$, so that $\mathbb{P}(i>_B j)$ is bounded away from $0$ and $1$ for $i,j \in [n]$. This set-up ensures $b^*$ does not become too large.

\begin{figure}%
    \centering
    \includegraphics[width=7.5cm]{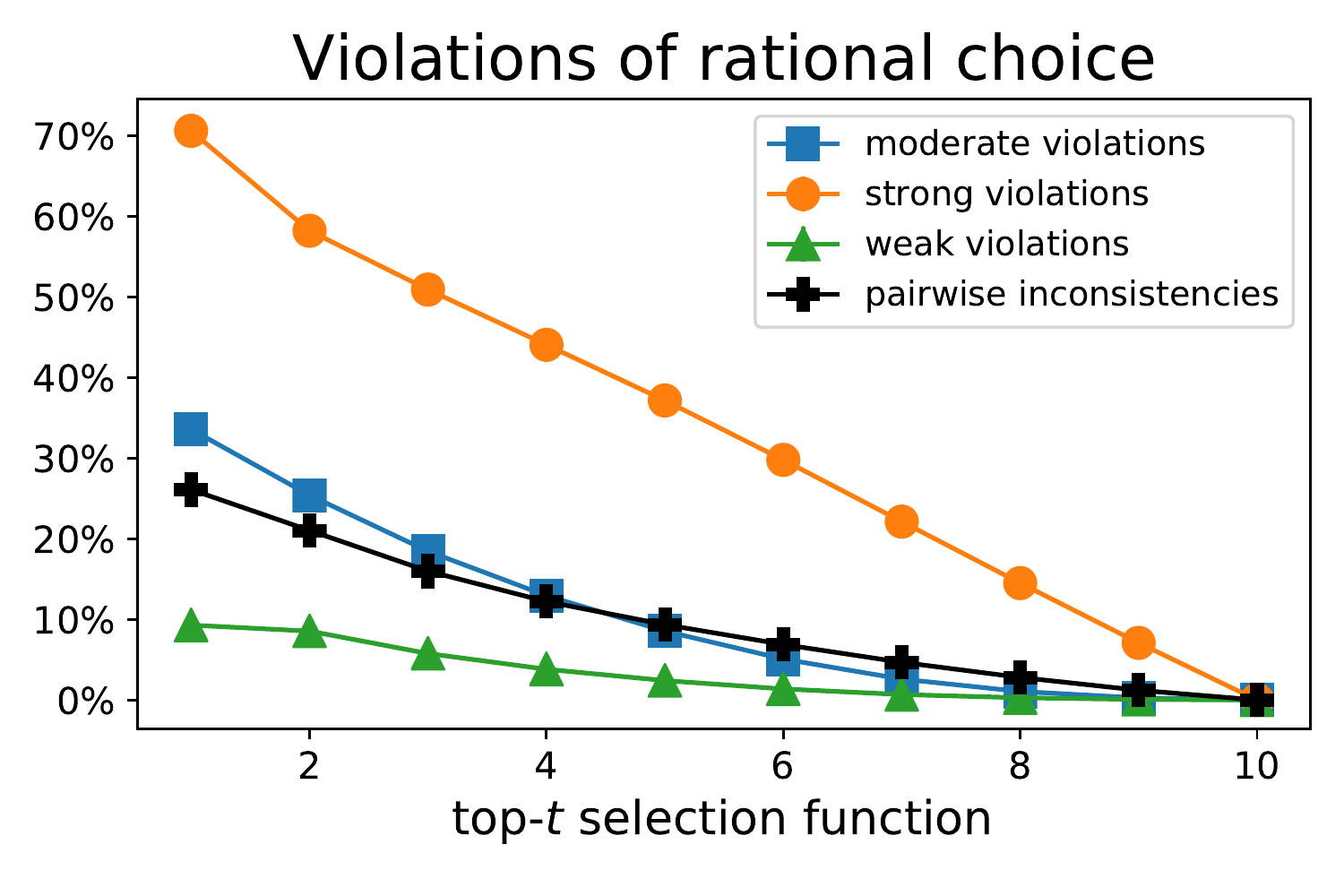}
    \caption{The salient feature preference model with the top-$t$ selection function produces systematic intransitives and pairwise comparisons that are inconsistent with the underlying ranking. When $t=10$, the salient feature preference model with the top-$t$ selection function is the FBTL model, and hence there are no intransitives or pairwise inconsistencies.}
    \label{fig:violationRationalChoice}
\end{figure}

First, the salient feature preference model can produce preferences that systematically violate rational choice. In contrast, the FBTL model cannot. Let $P_{ij} = \mathbb{P}(i>_B j)$ and $T = \{ (i,j,k) \in [n]^3: P_{ij} > .5, P_{jk} > .5\}$. Then $(i,j,k) \in T$ satisfies strong stochastic transitivity if $P_{ik}\geq \max\{P_{ij}, P_{jk} \}$, moderate stochastic transitivity if $P_{ik} \geq \min\{P_{ij}, P_{jk} \}$, and weak stochastic transitivity if $P_{ik} \geq .5$ \cite{cattelan2012models}. We sample $U$ and $w^*$ 10 times as described in the beginning of the section and allow $t$ to vary in $[d]$. Figure \ref{fig:violationRationalChoice} shows the average ratio of the number of weak, moderate, and strong stochastic transitivity violations to $|T|$ as a function of $t \in [d]$. There is very little deviation from the average. The standard error bars over the 10 experiments were plotted but they are so small that the markers covered them. All $\binom{n}{2}$ probabilities given by Equation \eqref{eq:thresholdBTL} are used to calculate the intransitivity rates. In the same figure we also show the percentage of pairwise comparisons that are inconsistent with the true ranking under the same experimental set-up. These are the pairs $i,j$ such that $\langle U_{i} - U_{j}, w^* \rangle < 0$, meaning item $i$ is ranked lower than item $j$ in the true ranking, but $\langle U_{i}^{\tau_{t}(i,j)} - U_{j}^{\tau_{t}(i,j)}, w^* \rangle >0$ meaning item $i$ beats item $j$ by at least 50\% when compared in isolation from the other items.
%
Notice that when $t=10$, the salient feature preference model is the FBTL model, so there are no pairwise inconsistencies or intransitives. Although this example is synthetic, real data exhibits intransitivity and even inconsistent pairs with the underlying ranking as discussed in the real data experiments in Section \ref{sec:realDataExp}. 

\begin{figure}%
    \centering
    \includegraphics[width=7.5cm]{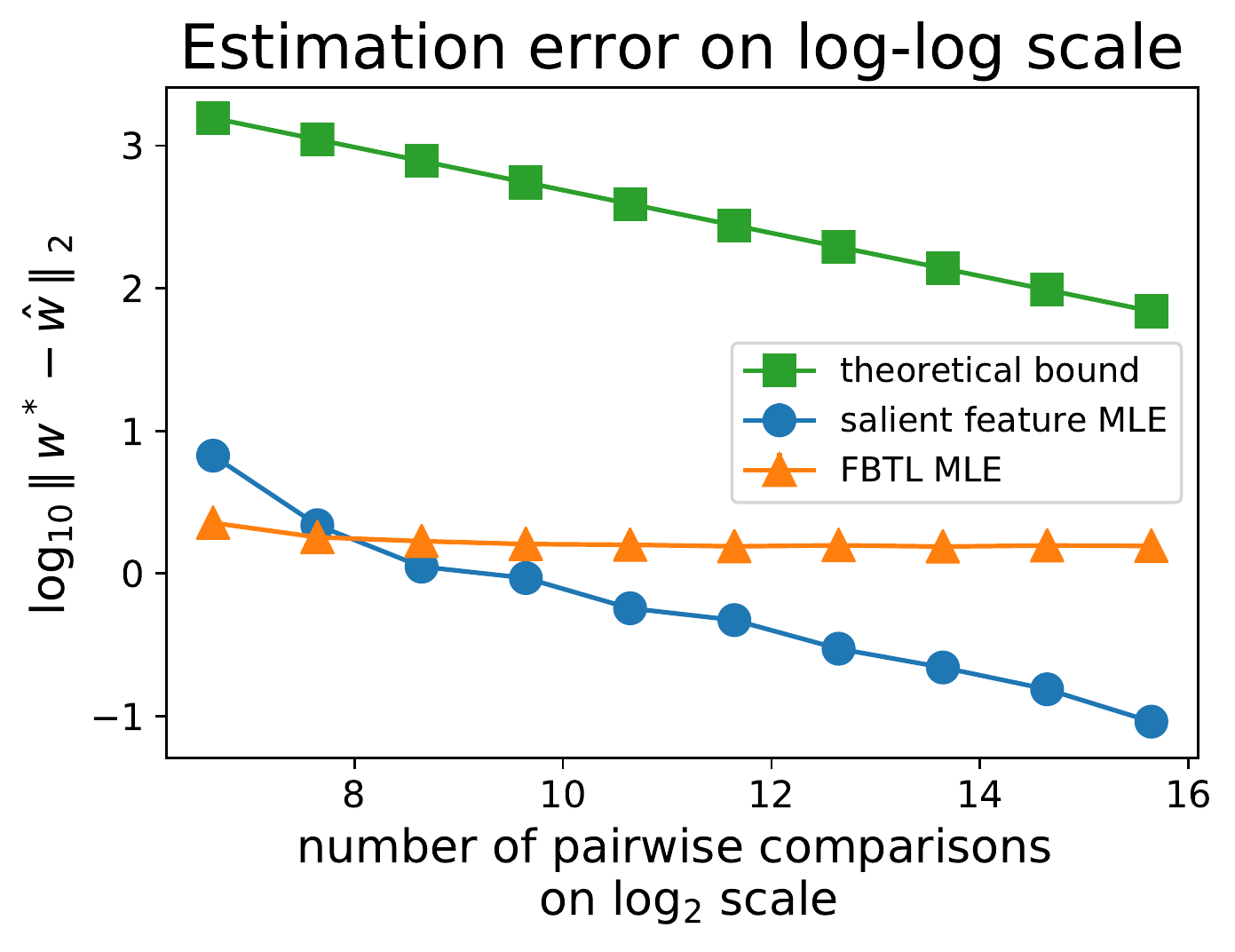}
    \caption{Illustration of Theorem \ref{thm:sampleComplexity} with the exact theoretical upper bound for the salient feature preference model with the top-$1$ selection function. Although there is a gap between the bound and the observed estimation error, they decrease at the same rate eventually. Excluding the first two points, the salient feature MLE error’s slope on the log-log scale is -0.154, whereas the theoretical bound’s slope is -0.151.}
    \label{fig:varySamples}%
\end{figure}

Second, we illustrate Theorem \ref{thm:sampleComplexity} with the top-$1$ selection function, and where $U$ and $w$ are sampled once as described in the beginning of this section. We sample $m$ pairwise comparisons for $m \in \{(100)2^{i-1}: i \in [10]\}$, fit the MLEs of both the salient preference model with the top-$1$ selection function and FBTL, and repeat 10 times. Figure \ref{fig:varySamples} shows the average estimation error of $w^*$ on a logarithmic scale as a function of the number of pairwise comparison samples also on a logarithmic scale. Figure \ref{fig:varySamples} also shows the exact theoretical upper bound where $\delta = \frac{1}{d} = \frac{1}{10}$ of Theorem \ref{thm:sampleComplexity} without constants $C_1$ and $C_2$ as stated in Section \ref{appendix:sampleComplexityProof} of the Supplement.
Again, there is very little deviation from the average. The standard error bars over the 10 experiments were plotted but they are so small that the markers covered them. 
There is a gap between the observed error and the theoretical bound, though the error decreases at the same rate. 
The error of the MLE of FBTL does not improve with more samples, since the pairwise comparisons are generated according to the salient feature preference model with the top-$1$ selection function. 
 See Section \ref{sec:additionalExperiments} in the Supplement for investigating model misspecification, i.e. fitting the MLE of the top-$t$ selection function for $t \neq 1$ with the same experimental set-up. 

 By estimating $w^*$ well, we can estimate the underlying ranking well by Corollary \ref{coro:rankingComplexity}. Under the same experimental set up, Figure \ref{fig:ktError} shows the Kendall tau correlation (definition given in Supplement \ref{sec:additionalExperiments}) between the true ranking (obtained by ranking the items according to $\langle U_i, w^* \rangle$) and the estimated ranking (according to $\langle U_i, \hat{w} \rangle$) but on a new set of 100 items drawn from the same distribution. 
The maximum Kendall tau correlation between two rankings is 1 and occurs when both rankings are equal.  Also, estimating $w^*$ well allows us to predict the outcome of unseen pairwise comparisons well, as shown in the Supplement in Section \ref{sec:additionalExperiments}. 

\begin{figure}%
    \centering
    \includegraphics[width=7.5cm]{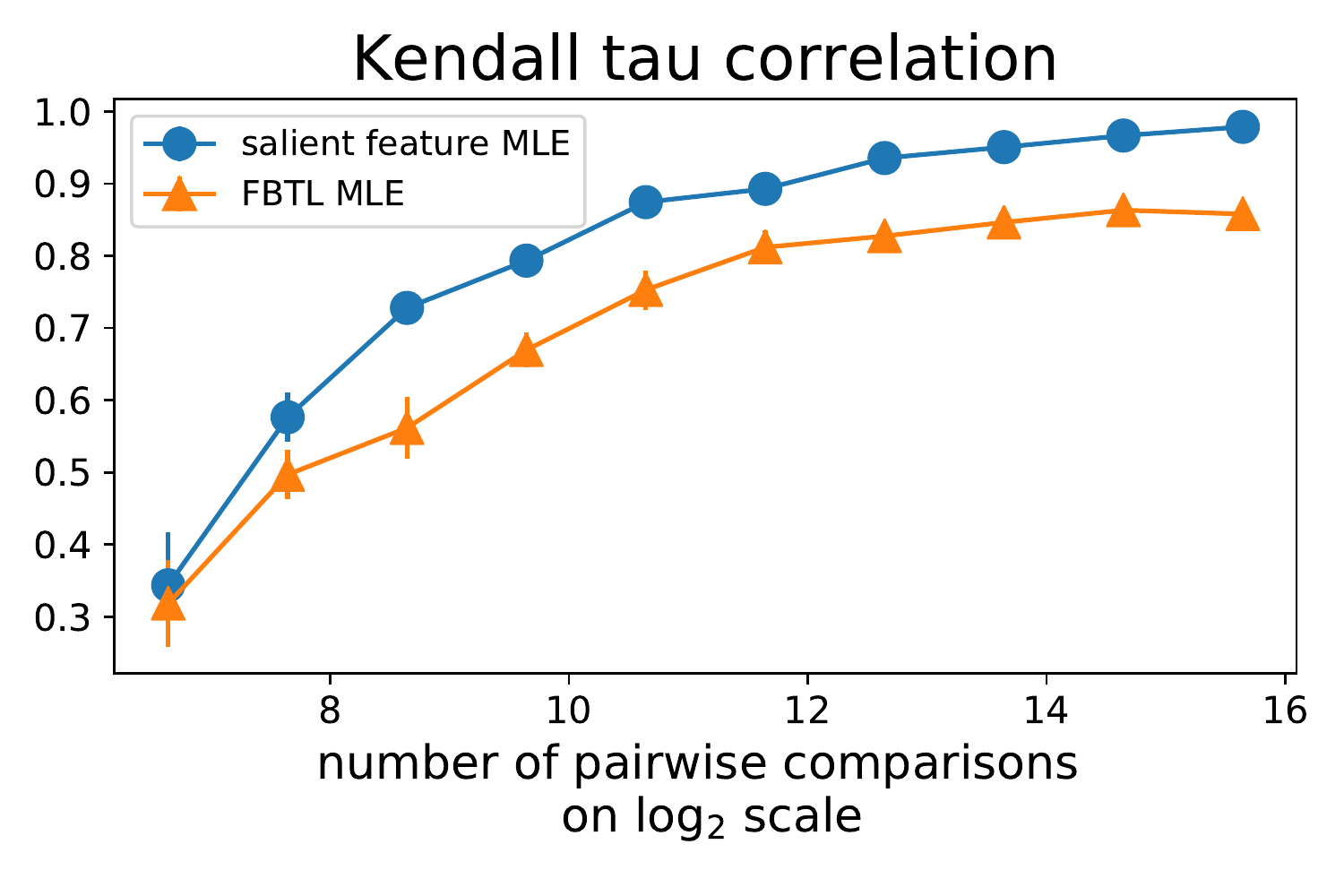}
    \caption{Kendall tau correlation between the true ranking and the estimated ranking where pairwise comparisons are sampled from the salient feature preference model with the top-$1$ selection function. Estimating $w^*$ well implies being able to estimate the underlying ranking well as stated in Corollary \ref{coro:rankingComplexity}.}
    \label{fig:ktError}%
\end{figure}

\label{sec:syntheticExperiments}

\subsection{Real Data}
\label{sec:realDataExp}
For the following experiments, we use the top-$t$ selection function for the salient feature preference model, where $t$ is treated as a hyperparameter and tuned on a validation set. 
We compare to FBTL, RankNet \cite{burges2005learning} with one hidden layer, and Ranking SVM  \cite{joachims2002optimizing}. We append an $\ell_2$ penalty to $\mathcal{L}_m$ for the salient feature preference model and the FBTL model, that is, for regularization parameter $\mu$, we solve $\min_{w \in \mathbb{R}^d} \mathcal{L}_{m}(w) + \mu\|w\|^2_2.$ For RankNet, we add to the objective function an $\ell_2$ penalty on the weights. As explained in more detail in subsections \ref{appendix:districtHyperparameters} and \ref{appendix:zapposHyperparameters} in the Supplement, the hyperparameters for the salient feature preference model are $t$ for the top-$t$ selection function and $\mu$, the hyperparameter for FBTL is $\mu$, the hyperparameter for Ranking SVM is the coefficient corresponding to the norm of the learned hyperplane, and the hyperparameters for RankNet are the number of nodes in the single hidden layer and the coefficient for the $\ell_2$ regularization of the weights.

\paragraph{District Compactness}
\cite{districtCompactness} collected preference data to understand human perception of compactness of legislative districts in the United States. 
Their data include both pairwise comparisons and $k$-wise ranking data for $k > 2$ as well as 27 continuous features for each district, including geometric features and compactness metrics. Although difficult to define precisely, the United States law suggests compactness is universally understood \cite{districtCompactness}. In fact, the authors provide evidence that most people agree on a universal ranking, but they found the pairwise comparison data was extremely noisy. They hypothesize that pairwise comparisons may not directly capture the full ranking, since all features may not be used when comparing two districts in isolation from the other districts. Hence, this problem is applicable to our salient feature preference model and its motivation.

\begin{table*}
  \caption{Average Kendall tau correlation over individual rankings on test sets for district compactness. The number in parenthesis is the standard deviation.}
  \label{table:districtResults}
  \centering
  \begin{tabular}{llllllll}
    \toprule
    \cmidrule(r){1-2}
Model:             & \texttt{Shiny1}    & \texttt{Shiny2}    & \texttt{UG1-j1} & \texttt{UG1-j2} & \texttt{UG1-j3} & \texttt{UG1-j4} & \texttt{UG1-j5} \\
    \midrule
Salient features   & \textbf{0.14} (.26) & \textbf{0.26} (.2) & \textbf{0.48} (.21) & \textbf{0.41} (.09) & \textbf{0.6} (.1) & 0.14 (.14) & \textbf{0.42} (.09)    \\ 
FBTL  & 0.09 (.22) & 0.18 (.17) & 0.2 (.12) & 0.26 (.07) & 0.45 (.15) & 0.2 (.13) & 0.06 (.14)    \\ 
Ranking SVM  & 0.09 (.22) & 0.18 (.17) & 0.22 (.12) & 0.26 (.07) & 0.45 (.15) & 0.2 (.13) & 0.06 (.14) \\
RankNet  & 0.12 (.24)& 0.24 (.18)& 0.28 (.14)& 0.37 (.08)& 0.53 (.11)& \textbf{0.28} (.08)& 0.15 (.15) \\ 
    \bottomrule
  \end{tabular}
\end{table*}

 \begin{table*}
  \centering
  \caption{Average pairwise prediction accuracy over 10 train/validation/test splits on the test sets by attribute for \texttt{UT Zappos50k}. $C$ stands for coarse and $F$ stands for fine grained. The number in parenthesis is the standard deviation.}
  \label{table:zappos}

  \begin{tabular}{lllllllll}
    \toprule
    \cmidrule(r){1-2}
Model:                     & open-$C$ & pointy-$C$ & sporty-$C$ & comfort-$C$ & open-$F$ & pointy-$F$ & sporty-$F$ & comfort-$F$ \\
    \midrule
Salient features     & 0.73 (.02) & 0.78 (.02) & 0.78 (.03) & 0.77 (.03) & 0.6 (.04) & 0.59 (.04) & 0.59 (.03) & 0.56 (.03) \\
FBTL     & 0.73 (.02) & 0.77 (.03) & 0.8 (.03) & 0.78 (.03) & 0.6 (.03) & 0.6 (.03) & 0.59 (.03) & 0.58 (.05) \\
Ranking SVM      & 0.74 (.02) & 0.78 (.03) & 0.79 (.03) & 0.78 (.03)  & 0.6 (.03) & 0.6 (.04) & 0.6 (.04) & 0.58 (.03)\\
RankNet  & 0.73 (.01) & 0.79 (.01) & 0.78 (.03) & 0.8 (.02) & 0.61 (.02) & 0.59 (.02) & 0.59 (.04) & 0.59 (.05) \\ 

    \bottomrule
  \end{tabular}
\end{table*}

The goal as set forth by \cite{districtCompactness} is to learn a ranking of districts. We train on 5,150 pairwise comparisons collected from 94 unique pairs of districts to learn $\hat{w}$, an estimate of the judgment vector $w^*$, then estimate a ranking by sorting the districts by $\langle \hat{w}, U_i\rangle$. The $k$-wise ranking data sets are used for validation and testing.  Since there is no ground truth for the universal ranking, we measure how close the estimated ranking is to each individual ranking. 
In this scenario, we care about the accuracy of the full ranking, and so we consider Kendall tau correlation.
Given a $k$-wise comparison data set, Table \ref{table:districtResults} shows the average Kendall tau correlation between the estimated ranking and each individual ranking where the number in parenthesis is the standard deviation. The standard deviation on \texttt{shiny1} and \texttt{shiny2} is relatively high because the Kendall tau correlation between pairs of rankings in these data sets has high variability, shown in Figure \ref{fig:intercoder} in the Supplement.

The MLE of the salient feature preference model under the top-$t$ selection function outperforms both the MLE of FBTL and Ranking SVM by a significant amount on 6 out 7 test sets, suggesting that pairwise comparison decisions may be better modeled by incorporating context. The MLE of the salient feature preference model, which is linear, is competitive with RankNet, which models pairwise comparisons as in Equation \eqref{eq:BTL} except where the utility of each item uses a function  $f$ defined by a neural network, i.e. $u_i = f(U_i)$.

The salient feature preference model may be outperforming FBTL and Ranking SVM since this data exhibits significant violations of rational choice. First, on the training set of pairwise comparisons, there are 48 triplets of districts $(i,j,k)$ where both (1) all three distinct pairwise comparisons were collected and (2) $P_{ij} > .5$ and $P_{jk} > .5$. Seventeen violate strong transitivity, 3 violate moderate transitivity, but none violate weak transitivity. 
Second, given a set of $k$-wise ranking data, let $\hat{P}_{ij}$ be the proportion of rankings in which item $i$ is ranked higher than item $j$.
There are 20 pairs of districts that appear in both the $k$-wise ranking data and
the pairwise comparison training data. Four of these pairs of items $i,j$ have the property that $(.5 - P_{ij})(.5 - \hat{P}_{ij})<0$, meaning item $i$ is typically ranked higher than item $j$ in the ranking data, but $j$ typically beats $i$ in the pairwise comparisons.

\paragraph{\texttt{UT Zappos50k}}
The \texttt{UT Zappos50K} data set consists of pairwise comparisons on images of shoes and 960 extracted vision features for each shoe \cite{finegrained, semjitter}. Given images of two shoes and an attribute from $\{$``open,'' ``pointy,'' ``sporty,'' ``comfort''$\}$, respondents picked which shoe exhibited the attribute more. The data consists of easier, coarse questions, i.e. based on comfort, pick between a slipper or high-heel, and harder, fine grained questions i.e. based on comfort, pick between two slippers. 

We now consider predicting pairwise comparisons instead of estimating a ranking since there is no ranking data available. We train four models, one for each attribute. See Table \ref{table:zappos} for the average pairwise comparison accuracy over ten train (70\%),  validation (15\%), and test splits (15\%) of the data. The pairwise comparison accuracy is defined as the percentage of items $(i,j)$ where $i$ beats $j$ a majority of the time and the model estimates the probability that $i$ beats $j$ exceeds $50\%$. 

In this case, the MLE of the FBTL model and the salient feature preference model under the top $t$ selection function perform similarly. Nevertheless, while the FBTL model utilizes all 990 features, the best $t$'s on each validation set and split of the data do not use all features, so our model is different from yet competitive to FBTL. See Table \ref{table:zapposT} in the Supplement. This suggests that the salient feature preference model under the top-$t$ selection function for relatively small $t$ is still a reasonable model for real data.

\section{Conclusion}

We focused on the problem of learning a ranking from pairwise comparison data with irrational choice behaviors, and we formulated the salient feature preference model where one uses projections onto salient coordinates in order to perform comparisons. We proved sample complexity results for MLE on this model and demonstrated the efficacy of our model on both synthetic and real data. Going forward, we would like to develop techniques to learn both the selection function $\tau$ and feature embeddings simultaneously. Finally, it will be useful to consider how to incorporate context into models more sophisticated than BTL, and also consider contextual effects in other tasks that use human judgements such as ordinal embedding \cite{terada2014local}.



\paragraph{Acknowledgements}
L. Balzano was supported by NSF CAREER award CCF-1845076, NSF BIGDATA award IIS-1838179, ARO YIP award W911NF1910027, and the Institute for Advanced Study Charles Simonyi Endowment. A. Bower was also supported by ARO W911NF1910027 as well as University of Michigan's Rackham Merit Fellowship, University of Michigan's Mcubed grant, and NSF graduate research fellowship DGE 1256260.

\medskip
\bibliography{biblio}

\begin{thebibliography}{43}
\providecommand{\natexlab}[1]{#1}
\providecommand{\url}[1]{\texttt{#1}}
\expandafter\ifx\csname urlstyle\endcsname\relax
  \providecommand{\doi}[1]{doi: #1}\else
  \providecommand{\doi}{doi: \begingroup \urlstyle{rm}\Url}\fi

\bibitem[Agresti(2012)]{agresti2012categorical}
Alan Agresti.
\newblock \emph{Categorical data analysis.}
\newblock John Wiley \& Sons, 2012.

\bibitem[Benson et~al.(2016)Benson, Kumar, and Tomkins]{benson2016relevance}
Austin~R Benson, Ravi Kumar, and Andrew Tomkins.
\newblock On the relevance of irrelevant alternatives.
\newblock In \emph{Proceedings of the 25th International Conference on World
  Wide Web}, pages 963--973. International World Wide Web Conferences Steering
  Committee, 2016.

\bibitem[Bordalo et~al.(2013)Bordalo, Gennaioli, and
  Shleifer]{bordalo2013salience}
Pedro Bordalo, Nicola Gennaioli, and Andrei Shleifer.
\newblock Salience and consumer choice.
\newblock \emph{Journal of Political Economy}, 121\penalty0 (5):\penalty0
  803--843, 2013.

\bibitem[Bradley and Terry(1952)]{bradley1952rank}
Ralph~Allan Bradley and Milton~E Terry.
\newblock Rank analysis of incomplete block designs: I. the method of paired
  comparisons.
\newblock \emph{Biometrika}, 39\penalty0 (3/4):\penalty0 324--345, 1952.

\bibitem[Brown and Peterson(2009)]{brown2009enquiry}
Thomas~C Brown and George~L Peterson.
\newblock An enquiry into the method of paired comparison: reliability,
  scaling, and thurstone's law of comparative judgment.
\newblock \emph{Gen Tech. Rep. RMRS-GTR-216WWW. Fort Collins, CO: US Department
  of Agriculture, Forest Service, Rocky Mountain Research Station. 98 p.}, 216,
  2009.

\bibitem[Burges et~al.(2005)Burges, Shaked, Renshaw, Lazier, Deeds, Hamilton,
  and Hullender]{burges2005learning}
Chris Burges, Tal Shaked, Erin Renshaw, Ari Lazier, Matt Deeds, Nicole
  Hamilton, and Greg Hullender.
\newblock Learning to rank using gradient descent.
\newblock In \emph{Proceedings of the 22nd international conference on Machine
  learning}, pages 89--96, 2005.

\bibitem[Cattelan(2012)]{cattelan2012models}
Manuela Cattelan.
\newblock Models for paired comparison data: A review with emphasis on
  dependent data.
\newblock \emph{Statistical Science}, pages 412--433, 2012.

\bibitem[Chen and Joachims(2016{\natexlab{a}})]{Chen:2016:MIM:2835776.2835787}
Shuo Chen and Thorsten Joachims.
\newblock Modeling intransitivity in matchup and comparison data.
\newblock In \emph{Proceedings of the Ninth ACM International Conference on Web
  Search and Data Mining}, WSDM '16, pages 227--236, New York, NY, USA,
  2016{\natexlab{a}}. ACM.

\bibitem[Chen and Joachims(2016{\natexlab{b}})]{chen2016predicting}
Shuo Chen and Thorsten Joachims.
\newblock Predicting matchups and preferences in context.
\newblock In \emph{Proceedings of the 22nd ACM SIGKDD International Conference
  on Knowledge Discovery and Data Mining}, pages 775--784. ACM,
  2016{\natexlab{b}}.

\bibitem[Heckel et~al.(2019)Heckel, Shah, Ramchandran, Wainwright,
  et~al.]{heckel2019active}
Reinhard Heckel, Nihar~B Shah, Kannan Ramchandran, Martin~J Wainwright, et~al.
\newblock Active ranking from pairwise comparisons and when parametric
  assumptions do not help.
\newblock \emph{The Annals of Statistics}, 47\penalty0 (6):\penalty0
  3099--3126, 2019.

\bibitem[Joachims(2002)]{joachims2002optimizing}
Thorsten Joachims.
\newblock Optimizing search engines using clickthrough data.
\newblock In \emph{Proceedings of the eighth ACM SIGKDD international
  conference on Knowledge discovery and data mining}, pages 133--142. ACM,
  2002.

\bibitem[Kaufman et~al.(2017)Kaufman, King, and
  Komisarchik]{districtCompactness}
Aaron Kaufman, Gary King, and Mayya Komisarchik.
\newblock How to measure legislative district compactness if you only know it
  when you see it.
\newblock \emph{American Journal of Political Science}, 2017.

\bibitem[Kleinberg et~al.(2017)Kleinberg, Mullainathan, and
  Ugander]{kleinberg2017comparison}
Jon Kleinberg, Sendhil Mullainathan, and Johan Ugander.
\newblock Comparison-based choices.
\newblock In \emph{Proceedings of the 2017 ACM Conference on Economics and
  Computation}, pages 127--144. ACM, 2017.

\bibitem[Li et~al.(2018)Li, Cheng, Fujii, Hsieh, and Hsieh]{NIPS2018_7746}
Yao Li, Minhao Cheng, Kevin Fujii, Fushing Hsieh, and Cho-Jui Hsieh.
\newblock Learning from group comparisons: Exploiting higher order
  interactions.
\newblock In \emph{Advances in Neural Information Processing Systems 31}, pages
  4981--4990. Curran Associates, Inc., 2018.

\bibitem[Lu and Negahban(2015)]{lu2015individualized}
Yu~Lu and Sahand~N Negahban.
\newblock Individualized rank aggregation using nuclear norm regularization.
\newblock In \emph{2015 53rd Annual Allerton Conference on Communication,
  Control, and Computing (Allerton)}, pages 1473--1479. IEEE, 2015.

\bibitem[Luce(1959)]{luce2012individual}
R~Duncan Luce.
\newblock \emph{Individual choice behavior: A theoretical analysis}.
\newblock Courier Corporation, 1959.

\bibitem[Makhijani and Ugander(2019)]{makhijani2019parametric}
Rahul Makhijani and Johan Ugander.
\newblock Parametric models for intransitivity in pairwise rankings.
\newblock In \emph{The World Wide Web Conference}, pages 3056--3062, 2019.

\bibitem[Menke and Martinez(2008)]{menke2008bradley}
Joshua~E Menke and Tony~R Martinez.
\newblock A bradley--terry artificial neural network model for individual
  ratings in group competitions.
\newblock \emph{Neural computing and Applications}, 17\penalty0 (2):\penalty0
  175--186, 2008.

\bibitem[Negahban et~al.(2016)Negahban, Oh, and Shah]{rankCentrality}
Sahand Negahban, Sewoong Oh, and Devavrat Shah.
\newblock Rank centrality: Ranking from pairwise comparisons.
\newblock \emph{Operations Research}, 65\penalty0 (1):\penalty0 266--287, 2016.

\bibitem[Negahban et~al.(2012)Negahban, Ravikumar, Wainwright, Yu,
  et~al.]{negahban2012unified}
Sahand~N Negahban, Pradeep Ravikumar, Martin~J Wainwright, Bin Yu, et~al.
\newblock A unified framework for high-dimensional analysis of $ m $-estimators
  with decomposable regularizers.
\newblock \emph{Statistical Science}, 27\penalty0 (4):\penalty0 538--557, 2012.

\bibitem[Niranjan and Rajkumar(2017)]{niranjan2017inductive}
UN~Niranjan and Arun Rajkumar.
\newblock Inductive pairwise ranking: going beyond the n log (n) barrier.
\newblock In \emph{Thirty-First AAAI Conference on Artificial Intelligence},
  2017.

\bibitem[Oh et~al.(2015)Oh, Thekumparampil, and Xu]{oh2015collaboratively}
Sewoong Oh, Kiran~K Thekumparampil, and Jiaming Xu.
\newblock Collaboratively learning preferences from ordinal data.
\newblock In \emph{Advances in Neural Information Processing Systems}, pages
  1909--1917, 2015.

\bibitem[Park et~al.(2015)Park, Neeman, Zhang, Sanghavi, and
  Dhillon]{park2015preference}
Dohyung Park, Joe Neeman, Jin Zhang, Sujay Sanghavi, and Inderjit Dhillon.
\newblock Preference completion: Large-scale collaborative ranking from
  pairwise comparisons.
\newblock In \emph{International Conference on Machine Learning}, pages
  1907--1916, 2015.

\bibitem[Pfannschmidt et~al.(2019)Pfannschmidt, Gupta, and
  H{\"{u}}llermeier]{DBLP:journals/corr/abs-1901-10860}
Karlson Pfannschmidt, Pritha Gupta, and Eyke H{\"{u}}llermeier.
\newblock Learning choice functions.
\newblock \emph{preprint}, abs/1901.10860, 2019.
\newblock URL \url{http://arxiv.org/abs/1901.10860}.

\bibitem[Plackett(1975)]{plackett1975analysis}
Robin~L Plackett.
\newblock The analysis of permutations.
\newblock \emph{Journal of the Royal Statistical Society: Series C (Applied
  Statistics)}, 24\penalty0 (2):\penalty0 193--202, 1975.

\bibitem[Ragain and Ugander(2016)]{ragain2016pairwise}
Stephen Ragain and Johan Ugander.
\newblock Pairwise choice markov chains.
\newblock In \emph{Advances in Neural Information Processing Systems}, pages
  3198--3206, 2016.

\bibitem[Rajkumar and Agarwal(2014)]{pmlr-v32-rajkumar14}
Arun Rajkumar and Shivani Agarwal.
\newblock A statistical convergence perspective of algorithms for rank
  aggregation from pairwise data.
\newblock In \emph{Proceedings of the 31st International Conference on Machine
  Learning}, volume~32 of \emph{Proceedings of Machine Learning Research},
  pages 118--126, Bejing, China, 22--24 Jun 2014. PMLR.

\bibitem[Rajkumar et~al.(2015)Rajkumar, Ghoshal, Lim, and
  Agarwal]{Rajkumar:2015:RSP:3045118.3045190}
Arun Rajkumar, Suprovat Ghoshal, Lek-Heng Lim, and Shivani Agarwal.
\newblock Ranking from stochastic pairwise preferences: Recovering condorcet
  winners and tournament solution sets at the top.
\newblock In \emph{Proceedings of the 32nd International Conference on
  International Conference on Machine Learning - Volume 37}, ICML'15, pages
  665--673. JMLR.org, 2015.

\bibitem[Rieskamp et~al.(2006)Rieskamp, Busemeyer, and
  Mellers]{rieskamp2006extending}
J{\"o}rg Rieskamp, Jerome~R Busemeyer, and Barbara~A Mellers.
\newblock Extending the bounds of rationality: Evidence and theories of
  preferential choice.
\newblock \emph{Journal of Economic Literature}, 44\penalty0 (3):\penalty0
  631--661, 2006.

\bibitem[Rosenfeld et~al.(2020)Rosenfeld, Oshiba, and
  Singer]{rosenfeld2019predicting}
Nir Rosenfeld, Kojin Oshiba, and Yaron Singer.
\newblock Predicting choice with set-dependent aggregation.
\newblock In \emph{Proceedings of the 37th International Conference on Machine
  learning}, 2020.

\bibitem[Saha and Rajkumar(2018)]{fbtl}
Aadirupa Saha and Arun Rajkumar.
\newblock Ranking with features: Algorithm and a graph theoretic analysis.
\newblock In \emph{preprint}, 2018.
\newblock URL \url{https://arxiv.org/pdf/1808.03857.pdf}.

\bibitem[Seshadri et~al.(2019)Seshadri, Peysakhovich, and
  Ugander]{seshadri2019discovering}
Arjun Seshadri, Alexander Peysakhovich, and Johan Ugander.
\newblock Discovering context effects from raw choice data.
\newblock \emph{International Conference on Machine Learning}, 2019.

\bibitem[Shah and Wainwright(2017)]{shah2017simple}
Nihar~B Shah and Martin~J Wainwright.
\newblock Simple, robust and optimal ranking from pairwise comparisons.
\newblock \emph{The Journal of Machine Learning Research}, 18\penalty0
  (1):\penalty0 7246--7283, 2017.

\bibitem[Shepard(1964)]{shepard1964attention}
Roger~N Shepard.
\newblock Attention and the metric structure of the stimulus space.
\newblock \emph{Journal of mathematical psychology}, 1\penalty0 (1):\penalty0
  54--87, 1964.

\bibitem[Terada and Luxburg(2014)]{terada2014local}
Yoshikazu Terada and Ulrike Luxburg.
\newblock Local ordinal embedding.
\newblock In \emph{International Conference on Machine Learning}, pages
  847--855, 2014.

\bibitem[Torgerson(1965)]{torgerson1965multidimensional}
Warren~S Torgerson.
\newblock Multidimensional scaling of similarity.
\newblock \emph{Psychometrika}, 30\penalty0 (4):\penalty0 379--393, 1965.

\bibitem[Tropp(2012)]{Tropp2012}
Joel~A Tropp.
\newblock User-friendly tail bounds for sums of random matrices.
\newblock \emph{Foundations of computational mathematics}, 12\penalty0
  (4):\penalty0 389--434, 2012.

\bibitem[Tversky(1972)]{tversky1972elimination}
Amos Tversky.
\newblock Elimination by aspects: A theory of choice.
\newblock \emph{Psychological review}, 79\penalty0 (4):\penalty0 281, 1972.

\bibitem[Tversky(1977)]{tversky1977features}
Amos Tversky.
\newblock Features of similarity.
\newblock \emph{Psychological review}, 84\penalty0 (4):\penalty0 327, 1977.

\bibitem[Tversky and Simonson(1993)]{tversky1993context}
Amos Tversky and Itamar Simonson.
\newblock Context-dependent preferences.
\newblock \emph{Management science}, 39\penalty0 (10):\penalty0 1179--1189,
  1993.

\bibitem[Yang and B.~Wakin(2015)]{dehui}
Dehui Yang and Michael B.~Wakin.
\newblock Modeling and recovering non-transitive pairwise comparison matrices.
\newblock \emph{2015 International Conference on Sampling Theory and
  Applications, SampTA 2015}, pages 39--43, 07 2015.

\bibitem[Yu and Grauman(2014)]{finegrained}
A.~Yu and K.~Grauman.
\newblock Fine-grained visual comparisons with local learning.
\newblock In \emph{Computer Vision and Pattern Recognition (CVPR)}, Jun 2014.

\bibitem[Yu and Grauman(2017)]{semjitter}
A.~Yu and K.~Grauman.
\newblock Semantic jitter: Dense supervision for visual comparisons via
  synthetic images.
\newblock In \emph{International Conference on Computer Vision (ICCV)}, Oct
  2017.

\end{thebibliography}
\pagenumbering{arabic}
\onecolumn
\section*{Supplement}
\section{Extension of the salient feature preference model to $k$-wise comparisons} 
\label{appendix:PLextension}
We describe how to extend the salient feature preference model of Equation \eqref{eq:thresholdBTL} from pairwise comparisons to $k$-wise comparisons when $k>2$. We base our generalization on the Placket-Luce model \cite{plackett1975analysis, luce2012individual}, which is a generalization of the BTL model from pairwise comparisons to $k$-wise comparisons.

Let the domain of the selection function $\tau$ be $[n]^k$ instead of $[n] \times [n]$, i.e. $\tau: [n]^k \rightarrow \mathcal{P}([d])$. Then for $T_{\ell} = ( t_1, \dots, t_k)$ where $t_i \in [n]$ are items, the probability of picking the ranking $t_1 >_B \cdots >_B t_k$ is
\begin{align}
\label{eq:thresholdPL}
    \mathbb{P}(t_1 >_B \cdots >_B t_k ) = \prod_{\ell = 1}^k \frac{  \exp \left(\langle U_{t_\ell}^{\tau(T_\ell)}, w^{*} \rangle \right)  }{ \sum_{j \in [k] \setminus [\ell-1]} \exp \left(\langle U_{t_j}^{\tau(T_\ell)}, w^{*} \rangle \right)},
\end{align}
where $``t_1 >_B \cdots >_B t_k"$ means item $t_1$ is preferred to item $t_2$ and so on and so forth.

We explain Equation \eqref{eq:thresholdPL}: Given items $T_{\ell} = ( t_1, \dots, t_k)$, first project each item's features $U_{t_i}$ onto the coordinate subspace spanned by the coordinates given by $\tau(T_\ell)$. Then the utility of item $t_i$ in the presence of the other items in $T$ is given by the inner product of its projected features with $w^*$: $ \langle (U_{t_i})^{\tau(T_\ell)}, w^{*} \rangle$. The higher the utility an item has, the more likely the item will be ranked higher among the items in $T_\ell$. Now imagine a bag of balls where each ball corresponds to one of the items in $T_\ell$. We select balls from this bag without replacement where the probability of picking a ball is the ratio of its utility to the sum of the utilities of all the remaining balls. The order in which we select balls results in a ranking of the $k$ items. This process is what Equation \eqref{eq:thresholdPL} represents. 

In the pairwise comparison case ($k=2$) for two items $T_\ell = (i, j)$, Equation \eqref{eq:thresholdPL} reduces to Equation \eqref{eq:thresholdBTL}, which is the salient preference model.  We can also extend the top-$t$ selection function naturally to accommodate $k$-wise comparisons.

\section{Negative log-likelihood derivation}
\label{appendix:LLderiv}
\begin{lemma}\label{lem:MLEObjective}
 Under the set-up of Section \ref{sec:modelFormulation}, the negative log-likelihood of $w\in \mathbb{R}^d$ is \begin{equation}\label{eq:logLossAppendix}
    \mathcal{L}_{m}(w; U, S_m, \tau ) = \sum_{\ell = 1}^m \log\left(1+\exp \left(\langle U_{i_\ell}^{\tau(i_\ell,j_\ell)} - U_{j_\ell}^{\tau(i_\ell,j_\ell)} , w \rangle \right) \right) - y_\ell  \langle U_{i_\ell}^{\tau(i_\ell,j_\ell)} - U_{j_\ell}^{\tau(i_\ell,j_\ell)}, w \rangle.
\end{equation}
\end{lemma}
\begin{proof}
Let $P_w(S_m)$ be the joint distribution of the $m$ samples $S_m$ with respect to the judgement vector $w$. Then

\begin{align}
    \ & \mathcal{L}_m(w; U,S_m, \tau) \\
    &= - \log P_w(S_m) \\
    &=- \log\left( \prod_{\ell = 1}^m ( \mathbb{P}(y_\ell = 1)^{y_\ell} \mathbb{P}(y_\ell = 0)^{1-y_\ell} )\right) \text{ by independence and since $y_\ell \in \{0,1\}$} \\
    &= - \sum_{i = 1}^m y_\ell\log(\mathbb{P}(y_\ell = 1)) + (1-y_\ell)\log(1-\mathbb{P}(y_\ell = 1)) \\ 
    &= - \sum_{i = 1}^m y_\ell\log\left(\frac{  \exp \left(\langle U_{i_\ell}^{\tau(i_\ell,j_\ell)} - U_{j_\ell}^{\tau(i_\ell,j_\ell)} , w \rangle \right)  }{ 1 + \exp \left(\langle U_{i_\ell}^{\tau(i_\ell,j_\ell)} - U_{j_\ell}^{\tau(i_\ell,j_\ell)}, w \rangle \right)}\right) \\ \nonumber
    & + (1-y_\ell)\log\left(\frac{ 1  }{ 1 + \exp \left(\langle U_{i_\ell}^{\tau(i_\ell,j_\ell)} - U_{j_\ell}^{\tau(i_\ell,j_\ell)}, w \rangle \right)}\right) \\ 
    &= \sum_{i = 1}^m \log\left(1+\exp \left(\langle U_{i_\ell}^{\tau(i_\ell,j_\ell)} - U_{j_\ell}^{\tau(i_\ell,j_\ell)} , w \rangle \right) \right) - y_\ell \langle U_{i_\ell}^{\tau(i_\ell,j_\ell)} - U_{j_\ell}^{\tau(i_\ell,j_\ell)} , w \rangle
\end{align}
\end{proof}
 
\section{Proof of Proposition \ref{prop:identifiability}}
\label{appendix:identifiability}

\begin{prop}[Restatement of Proposition \ref{prop:identifiability}]
Given item features $U \in \mathbb{R}^{d \times n}$, the salient feature preference model with selection function $\tau$ is identifiable if and only if $\text{span}\{ U_i^{\tau(i,j)} - U_j^{\tau(i,j)}: (i,j) \in P \} = \mathbb{R}^d$. 
\end{prop}
\begin{proof}
Let $w \in \mathbb{R}^d$. Then for any $(i,j) \in P$,
\begin{align}
   \mathbb{P}(i >_B j; w ) &= \mathbb{P}(i >_B j; w^* ) \label{eq:equality} \\ 
   \iff 
   \frac{  \exp \left(\langle U_{i}^{\tau(i ,j )} - U_j^{\tau(i ,j )} , w \rangle \right)  }{ 1 + \exp \left(\langle U_{i}^{\tau(i ,j )} - U_j^{\tau(i ,j )}, w \rangle \right)}  &= \frac{  \exp \left(\langle U_{i}^{\tau(i ,j )} - U_j^{\tau(i ,j )} , w^{*} \rangle \right)  }{ 1 + \exp \left(\langle U_{i}^{\tau(i ,j )} - U_j^{\tau(i ,j )}, w^{*} \rangle \right)} \\ 
   \iff \exp \left(\langle U_{i}^{\tau(i ,j )} - U_j^{\tau(i ,j )}, w \rangle \right) & = \exp \left(\langle U_{i}^{\tau(i ,j )} - U_j^{\tau(i ,j )}, w^{*} \rangle \right) \\ 
   \iff \langle U_{i}^{\tau(i ,j )} - U_j^{\tau(i ,j )}, w \rangle & = \langle U_{i}^{\tau(i ,j )} - U_j^{\tau(i ,j )}, w^{*} \rangle \\ 
   \iff \langle U_{i}^{\tau(i ,j )} - U_j^{\tau(i ,j )}, w^{*} - w \rangle &= 0. \label{eq:orthogonal}
\end{align}

$\Rightarrow$ Assume identifiability. By contradiction, if $\text{span}\{ U_i^{\tau(i,j)} - U_j^{\tau(i,j)}: (i,j) \in P \} \neq \mathbb{R}^d$, then there is some vector $x \neq 0$ that is orthogonal to $\text{span}\{ U_i^{\tau(i,j)} - U_j^{\tau(i,j)}: (i,j) \in P \}$. Consider $w^*-x$. Then, for any $(i,j) \in P$
\begin{align} 
\langle U_{i}^{\tau(i ,j )} - U_j^{\tau(i ,j )}, w^{*} - (w^* - x) \rangle & = \langle U_{i}^{\tau(i ,j )} - U_j^{\tau(i ,j )}, x \rangle \\ 
& = 0.
\end{align}
Therefore, with $w = w^* -x$, Equation \eqref{eq:orthogonal} is true and implies Equation \eqref{eq:equality} meaning \[\mathbb{P}(i > j; w^* -x ) = \mathbb{P}(i > j; w^* ),\] contradicting identifiability since $w^* - x \neq w^*$ because $x \neq 0$.

$\Leftarrow$ Now assume $\text{span}\{ U_i^{\tau(i,j)} - U_j^{\tau(i,j)}: (i,j) \in P \} = \mathbb{R}^d$. We want to prove identifiability so suppose there exists $w$ such that Equation \eqref{eq:equality} holds. We will show $w = w^*$. Let $x \in \mathbb{R}^d$ where $x = \sum_{(i,j) \in P} \alpha_{i,j} (U_{i}^{\tau(i ,j )} - U_j^{\tau(i ,j )})$ for $\alpha_{i,j} \in \mathbb{R}$. Then by Equation \eqref{eq:orthogonal}, \[\left\langle \sum_{(i,j) \in P} \alpha_{i, j} \left( U_{i}^{\tau(i ,j )} - U_j^{\tau(i ,j )} \right), w^*-w \right \rangle = 0.\] Since this is true for any $x \in \mathbb{R}^d$, $w^* - w =0$, which means $w = w^*.$
\end{proof}


\section{Proof of Proposition \ref{prop:lambdaIdentifiability}}
\label{appendix:lambdaIdentifiable}
\begin{prop}[Restatement of Proposition \ref{prop:lambdaIdentifiability}]
 Under the set-up of Section \ref{sec:modelFormulation}, $\lambda := \lambda_{\min}(\mathbb{E}(U_i^{\tau(i,j)} - U_j^{\tau(i,j)})(U_i^{\tau(i,j)} - U_j^{\tau(i,j)})^T) >0 $ if and only if the salient feature preference model with selection function $\tau$ is identifiable.
\end{prop}

\begin{proof}
For both directions, we prove the contrapositive.

$\Rightarrow$ Assume $\lambda_{\min}(\mathbb{E}(U_i^{\tau(i,j)} - U_j^{\tau(i,j)})(U_i^{\tau(i,j)} - U_j^{\tau(i,j)})^T) = 0$. Recall the expectation is with respect to a uniformly at random chosen pair of items. Let $\mathbf{0} \in \mathbb{R}^d$ be the all 0 vector. Then there exists $y\neq \mathbf{0} \in \mathbb{R}^d$ that has unit norm such that 
\begin{align}
    & (\mathbb{E}(U_i^{\tau(i,j)} - U_j^{\tau(i,j)})(U_i^{\tau(i,j)} - U_j^{\tau(i,j)})^T) y = \textbf{0} \\
    & \implies y^T(\mathbb{E}(U_i^{\tau(i,j)} - U_j^{\tau(i,j)})(U_i^{\tau(i,j)} - U_j^{\tau(i,j)})^T) y = 0 \\   
    & \implies \frac{1}{\binom{n}{2}} \sum_{(i,j) \in P } y^T(U_i^{\tau(i,j)} - U_j^{\tau(i,j)})(U_i^{\tau(i,j)} - U_j^{\tau(i,j)})^T) y = 0  \text{ since } (i,j) \in P \text{ is chosen uniformly at random} \\ 
    & \implies \frac{1}{\binom{n}{2}} \sum_{(i,j) \in P } \|(U_i^{\tau(i,j)} - U_j^{\tau(i,j)})^Ty\|_2^2 = 0\\ 
    & \implies \|(U_i^{\tau(i,j)} - U_j^{\tau(i,j)})^Ty\|_2^2 = 0 \ \forall (i,j) \in P\\ 
    & \implies (U_i^{\tau(i,j)} - U_j^{\tau(i,j)})^Ty = \textbf{0} \  \forall (i,j) \in P.
\end{align}

We now show $y \notin \text{span}\{ U_i^{\tau(i,j)} - U_j^{\tau(i,j)}: (i,j) \in P \}$, which establishes the salient feature preference model is not identifiable by Proposition \ref{prop:identifiability}. By contradiction, suppose there exist $\alpha_{i,j} \in \R$ such that \[y = \sum_{(i,j) \in P} \alpha_{i, j} ( U_{i}^{\tau(i ,j )} - U_j^{\tau(i ,j )}).\] Then
\begin{align} 
1 &= \langle y, y \rangle \\
& = \left\langle \sum_{(i,j) \in P} \alpha_{i, j} \left( U_{i}^{\tau(i ,j )} - U_j^{\tau(i ,j )} \right),y \right \rangle \\
&=  \sum_{(i,j) \in P} \alpha_{i, j} \left \langle \left( U_{i}^{\tau(i ,j )} - U_j^{\tau(i ,j )} \right),y \right \rangle \\ 
& = 0,
\end{align}
a contradiction.

$\Leftarrow$ Now suppose that the preference model is not identifiable. By Proposition \ref{prop:identifiability}, $ \text{span}\{ U_i^{\tau(i,j)} - U_j^{\tau(i,j)}: (i,j) \in P \} \neq \R^d$. In particular, there exists $y \in \R^d$ such that $y \neq \textbf{0}$ and $\langle y, U_i^{\tau(i,j)} - U_j^{\tau(i,j)} \rangle = 0$ for all $(i,j) \in P$, i.e. $y$ is in the orthogonal complement of $\text{span}\{ U_i^{\tau(i,j)} - U_j^{\tau(i,j)}: (i,j) \in P \}$. Furthermore,
\begin{align}
    \frac{1}{\binom{n}{2}} \sum_{(i,j) \in P } (U_i^{\tau(i,j)} - U_j^{\tau(i,j)})(U_i^{\tau(i,j)} - U_j^{\tau(i,j)})^T y = \textbf{0}  \\
    \implies (\mathbb{E}(U_i^{\tau(i,j)} - U_j^{\tau(i,j)})(U_i^{\tau(i,j)} - U_j^{\tau(i,j)})^T) y & = \textbf{0},  \\
\end{align}
since the expectation is with respect to a uniformly at random chosen pair of items. Therefore, $\lambda_{\min}(\mathbb{E}(U_i^{\tau(i,j)} - U_j^{\tau(i,j)})(U_i^{\tau(i,j)} - U_j^{\tau(i,j)})^T) = 0$ since all the eigenvalues of $\mathbb{E}(U_i^{\tau(i,j)} - U_j^{\tau(i,j)})(U_i^{\tau(i,j)} - U_j^{\tau(i,j)})^T$ are non-negative since it is a sum of positive semidefinite matrices, and 0 is an eigenvalue. 
\end{proof}

\section{Proof of Theorem \ref{thm:sampleComplexity}}
\label{appendix:sampleComplexityProof}

Recall the set-up from the beginning of Section \ref{sec:modelFormulation}. There are $n$ items where the features of the items are given by the columns of $U \in \mathbb{R}^{d \times n}$ and let $w^* \in \mathbb{R}^d$ be the judgment vector. Let $\tau$ be the selection function. Let $S_m = \{(i_{\ell}, j_{\ell}, y_\ell)\}_{\ell = 1}^m$ be the $m$ samples of independent pairwise comparisons where each pair of items $(i_{\ell}, j_{\ell})$ is chosen uniformly at random from all the pairs of items $P := \{(i,j) \in [n] \times [n]: i < j\}.$ Furthermore, $y_\ell$ is 1 if the {$i_{\ell}$-th} item beats the {$j_{\ell}$-th} item and 0 otherwise where $y_\ell \sim \text{Bernoulli}\left(\frac{  \exp \left(\langle U_{i_\ell}^{\tau(i_\ell,j_\ell)} - U_{j_\ell}^{\tau(i_\ell,j_\ell)} , w^{*} \rangle \right)  }{ 1 + \exp \left(\langle U_{i_\ell}^{\tau(i_\ell,j_\ell)} - U_{j_\ell}^{\tau(i_\ell,j_\ell)}, w^{*} \rangle \right)}\right)$. We will not repeat these assumptions in the following lemmas.

In this section, we present the exact lower bounds on the number of samples and upper bound on the estimation error. The exact values of the constants that appear in the main text, i.e. $C_1$ and $C_2$, appear at the end of the proof. 

\begin{theorem}[restatement of Theorem \ref{thm:sampleComplexity}: sample complexity of estimating $w^*$]
Let $U$, $w^*$, $\tau$, and $S_m$ be defined as above. Let $\hat{w}$ be the maximum likelihood estimator, i.e. the minimum of $\mathcal{L}_m$ in Equation \eqref{eq:logLoss}, restricted to the set $\mathcal{W}(b^*)$. The following expectations are taken with respect to a uniformly chosen random pair of items from $P$. For $(i,j) \in P$, let 
\begin{align*} Z_{(i,j)} &:=(U_{i}^{\tau(i,j)} - U_{j}^{\tau(i,j)}) (U_{i}^{\tau(i,j)} - U_{j}^{\tau(i,j)})^T
\\
\lambda &:= \lambda_{\min}( \mathbb{E} Z_{(i,j)}),\\ 
\eta &:= \sigma_{\max}(\mathbb{E}((Z_{(i,j)} - \mathbb{E}Z_{(i,j)})^2)),\\
\zeta &:= \max_{(k,\ell) \in P} \lambda_{\max}(\mathbb{E} Z_{(i,j)  } - Z_{ (k,\ell)}),
\end{align*}
where for a positive semidefinite matrix $X$, $\lambda_{\min}(X)$ and $\lambda_{\max}(X)$ are the smallest/largest eigenvalues of $X$, and where for any matrix $X$, $\sigma_{\max}(X)$ is the largest singular value of $X$. Let \begin{equation}
\beta:=\max_{(i,j) \in P} \| U_{i}^{\tau(i,j)} - U_{j}^{\tau(i,j)} \|_{\infty}. 
\end{equation}

Let $\delta >0.$ If $\lambda >0$ and if \[m \geq \max\left\{\frac{3\beta^2\log{(4d / \delta)}d + 4\sqrt{d}\beta \log{(4 d / \delta)}}{6}, \frac{8\log(2 d / \delta)(6\eta + \lambda \zeta)}{3\lambda^2} \right\},\] then with probability at least $1 - \delta$,
\[ \| w^* - \hat{w}\|_2 \leq \frac{4(1 + \exp(b^*))^2}{\exp(b^*)\lambda}  \sqrt{\frac{3\beta^2\log{(4d / \delta)}d + 4\sqrt{d}\beta \log{(4 d /  \delta)}}{6m}}\] where the randomness is from the randomly chosen pairs and the outcomes of the pairwise comparisons.
\end{theorem}

\begin{proof}
We use the proof technique of Theorem 4 in \cite{rankCentrality}. We use the notation $\mathcal{L}_m(w)$ instead of $\mathcal{L}_m(w; U, S_m, \tau)$ throughout the proof since it is clear from context.

By definition $\mathcal{L}_m(\hat{w}) \leq \mathcal L_m(w^*)$. Let $\Delta := \hat{w} - w^*$. Then
\begin{align}
    & \ \mathcal{L}_m(w^* + \Delta) - \mathcal{L}_m(w^*) - \langle \nabla \mathcal{L}_m(w^*), \Delta \rangle \label{eq:toBound} \\
    & \leq - \langle \nabla \mathcal{L}_m(w^*), \Delta \rangle \\
    & \leq \|\nabla \mathcal{L}_m(w^*)\|_2 \|\Delta \|_2, \label{eq:CSUpperBound}
\end{align}
by the Cauchy-Schwarz inequality. 

Recall Taylor's theorem:
\begin{theo}[Taylor's Theorem]\label{thm:taylorThm}
Let $f: \mathbb{R}^n \rightarrow \mathbb{R}$. If the Hessian $H_f$ of $f$ exists everywhere on its domain, then for any $x, \Delta \in \mathbb{R}^n$, there exists $\lambda \in [0,1]$ such that $f(x + \Delta) = f(x) + \langle \nabla f(x), \Delta \rangle + \frac{1}{2} \Delta^T H_f(x + \lambda \Delta) \Delta $.
\end{theo}

Now, we lower bound Equation \eqref{eq:toBound}. Let $H_{\mathcal{L}_m}$ be the Hessian of $\mathcal{L}_m$. Then by Taylor's theorem, there exists $\lambda \in [0,1]$ such that 
\begin{align}
    & \ \frac{1}{m} \left(\mathcal{L}_m(w^* + \Delta) - \mathcal{L}_m(w^*) - \langle \nabla \mathcal{L}_m(w^*), \Delta \rangle \right) \\
    & = \frac{1}{2m} \Delta^T H_{\mathcal{L}_m}(w^* + \lambda \Delta) \Delta \label{eq:taylor} \\
    & = \frac{1}{2m} \sum_{\ell = 1}^m h(\langle w^* + \lambda \Delta, U_{i_\ell}^{\tau(i_\ell,j_\ell)} - U_{j_\ell}^{\tau(i_\ell,j_\ell)} \rangle) \Delta^T  (U_{i_\ell}^{\tau(i_\ell,j_\ell)} - U_{j_\ell}^{\tau(i_\ell,j_\ell)}) (U_{i_\ell}^{\tau(i_\ell,j_\ell)} - U_{j_\ell}^{\tau(i_\ell,j_\ell)})^T \Delta \label{eq:hessian}
\end{align}

where the Hessian $H_{\mathcal{L}_m}$ is computed in Lemma \ref{lem:gradientHessian} and $h(x) := \frac{e^x}{(1+e^x)^2}$.

Note 
\begin{align}
    & |\langle w^* + \lambda \Delta, U_{i_\ell}^{\tau(i_\ell,j_\ell)} - U_{j_\ell}^{\tau(i_\ell,j_\ell)} \rangle| \\
    & = |(1-\lambda) \langle w^*, U_{i_\ell}^{\tau(i_\ell,j_\ell)} - U_{j_\ell}^{\tau(i_\ell,j_\ell)} \rangle + \lambda \langle \hat{w}, U_{i_\ell}^{\tau(i_\ell,j_\ell)} - U_{j_\ell}^{\tau(i_\ell,j_\ell)} \rangle| \\
    & \leq (1-\lambda) b^* + \lambda b^* \\
    & = b^*
\end{align}
where the second to last inequality is by definition of $b^*$ and since $ \hat{w} \in \mathcal{W}(b^*)$. Because $h(x) = \frac{e^x}{(1+e^x)^2}$ is symmetric and decreases on $[0, \infty)$ by Lemma \ref{lem:symmetricAndDecreasing}, for any $i,j \in [n]$, \[h(\langle w^* + \lambda \Delta, U_{i}^{\tau(i,j)} - U_{j}^{\tau(i,j)} \rangle) \geq h(b^*) = \frac{\exp(b^*)}{(1 + \exp(b^*))^2}.\]

Therefore,
\begin{align}
  & \frac{1}{m} \left(\mathcal{L}_m(w^* + \Delta) - \mathcal{L}_m(w^*) - \langle \nabla \mathcal{L}_m(w^*), \Delta \rangle \right) \\
  & \geq \frac{\exp(b^*)}{2m(1 + \exp(b^*))^2} \sum_{\ell = 1}^m  \Delta^T (U_{i_\ell}^{\tau(i_\ell,j_\ell)} - U_{j_\ell}^{\tau(i_\ell,j_\ell)}) (U_{i_\ell}^{\tau(i_\ell,j_\ell)} - U_{j_\ell}^{\tau(i_\ell,j_\ell)})^T \Delta. \label{eq:hessianLowerBound}
\end{align}

By Lemma \ref{lem:gradientUpperBound} and \ref{lem:randomMatrixLowerBound} and combining Equations \eqref{eq:CSUpperBound} and \eqref{eq:hessianLowerBound}, with probability at least $1 - \delta$ if \[m \geq \max\left\{\frac{3\beta^2\log{(4d / \delta)}d + 4\sqrt{d}\beta \log{(4d / \delta)}}{6}, \frac{8\log(2 d / \delta)(6\eta + \lambda \zeta)}{3\lambda^2} \right\},\]
\begin{align}
    \left(\frac{\exp(b^*)}{2(1 + \exp(b^*))^2}\right)\frac{\lambda}{2} \|\Delta\|_2^2 & \leq \frac{1}{m}\left(\mathcal{L}_m(w^* + \Delta) - \mathcal{L}_m(w^*) - \langle \nabla \mathcal{L}_m(w^*), \Delta \rangle \right) \\
    & \leq \sqrt{\frac{3\beta^2\log{(4d / \delta)}d + 4\sqrt{d}\beta \log{(4d / \delta)}}{6m}} \|\Delta\|_2 \\
    & \implies \|\Delta\|_2 \leq \frac{4(1 + \exp(b^*))^2}{\exp(b^*)\lambda} \sqrt{\frac{3\beta^2\log{(4d / \delta)}d + 4\sqrt{d}\beta \log{(4 d / \delta)}}{6m}}.
\end{align}
In the main paper with order terms, it is easy to see the $O(\cdot)$ bound on the upper bound on the estimation error. Furthermore, it is easy to see that for the constants $C_1$ and $C_2$ given in the main paper, we have $C_1 = 4/6$ and $C_2 = 48 / 3$. 
\end{proof}

We now present the lemmas used in the prior proof.

\begin{lemma}\label{lem:gradientUpperBound}
Let $\delta > 0$. Under the model assumptions in this section, if \[m \geq \frac{3\beta^2\log{(4d / \delta)}d + 4\sqrt{d}\beta \log{(4d / \delta)}}{6},\] then with probability at least $1-\frac{\delta}{2}$, \[ \left\|\frac{1}{m} \nabla \mathcal{L}_m(w^*)\right\|_2 \leq \sqrt{\frac{3\beta^2\log{(4d / \delta)}d + 4\sqrt{d}\beta \log{(4d / \delta)}}{6m}}\] where $\beta:=\max_{(i,j) \in P} \left\| U_{i}^{\tau(i,j)} - U_{j}^{\tau(i,j)} \right\|_{\infty}.$
\end{lemma}
\begin{proof}
For $\ell \in [m]$, let \[X_\ell = \frac{1}{m} \left( U_{i_\ell}^{\tau(i_\ell,j_\ell)} - U_{j_\ell}^{\tau(i_\ell,j_\ell)} \right) \left( \frac{\exp(\langle w^*, U_{i_\ell}^{\tau(i_\ell,j_\ell)} - U_{j_\ell}^{\tau(i_\ell,j_\ell)} \rangle) }{1+ \exp(\langle w^*, U_{i_\ell}^{\tau(i_\ell,j_\ell)} - U_{j_\ell}^{\tau(i_\ell,j_\ell)} \rangle)} - y_\ell\right),\]
so $\frac{1}{m} \nabla \mathcal{L}_m(w^*) = \sum_{\ell =1}^m X_{\ell}$ by Lemma \ref{lem:gradientHessian}.

We now show (1) $\mathbb{E}(X_\ell)=0$ where the expectation is taken with respect to a uniformly chosen pair of items, (2) the coordinates of $X_\ell$ are bounded, and (3) the coordinates of $X_\ell$ have bounded second moments.

First $\mathbb{E}(X_\ell)=0$. By conditioning on each pair of items, each of which have the same probability of being chosen,
\begin{align}
    \mathbb{E} (X_\ell) &= \frac{1}{ \binom{n}{2}} \sum_{(i,j) \in P} \mathbb{E} (X_\ell| \text{items } i,j \text{ are chosen} ) \\
    & = \frac{1}{ \binom{n}{2}} \sum_{(i,j) \in P} \frac{1}{m}\left( U_{i}^{\tau(i,j)} - U_{j}^{\tau(i,j)} \right) \left( \frac{\exp(\langle w^*, U_{i}^{\tau(i,j)} - U_{j}^{\tau(i,j)} \rangle) }{1+ \exp(\langle w^*, U_{i}^{\tau(i,j)} - U_{j}^{\tau(i,j)} \rangle)} - \mathbb{E}(y_{(i,j)})\right) \\ 
    &= \frac{1}{ \binom{n}{2}} \sum_{(i,j) \in P} \frac{1}{m}\left( U_{i}^{\tau(i,j)} - U_{j}^{\tau(i,j)} \right)  \left( \frac{\exp(\langle w^*, U_{i}^{\tau(i,j)} - U_{j}^{\tau(i,j)} \rangle) }{1+ \exp(\langle w^*, U_{i}^{\tau(i,j)} - U_{j}^{\tau(i,j)} \rangle)} - \frac{\exp(\langle w^*, U_{i}^{\tau(i,j)} - U_{j}^{\tau(i,j)} \rangle) }{1+ \exp(\langle w^*, U_{i}^{\tau(i,j)} - U_{j}^{\tau(i,j)} \rangle)}\right) \\ 
    & = 0,
\end{align}
where the expectation is with respect to the random pair that is drawn and the outcome of the pairwise comparison.

Second, $|X_\ell^{(k)}| \leq \frac{\beta}{m}$ where $X_\ell^{(k)}$ is the $k$-th coordinate of $X_\ell$. Then for $k \in [d]$
\begin{align}
    & |X_\ell^{(k)}| \\
    & = \left|\frac{1}{m} \left( (U_{i_\ell}^{\tau(i_\ell,j_\ell)})^{(k)} - (U_{j_\ell}^{\tau(i_\ell,j_\ell)})^{(k)} \right) \left( \frac{\exp(\langle w^*, U_{i_\ell}^{\tau(i_\ell,j_\ell)} - U_{j_\ell}^{\tau(i_\ell,j_\ell)} \rangle) }{1+ \exp(\langle w^*, U_{i_\ell}^{\tau(i_\ell,j_\ell)} - U_{j_\ell}^{\tau(i_\ell,j_\ell)} \rangle)} - y_\ell\right)\right| \\
    & \leq \frac{1}{m} \left|\left( (U_{i_\ell}^{\tau(i_\ell,j_\ell)})^{(k)} - (U_{j_\ell}^{\tau(i_\ell,j_\ell)})^{(k)} \right) \right| \text{ since } \frac{\exp(\langle w^*, U_{i_\ell}^{\tau(i_\ell,j_\ell)} - U_{j_\ell}^{\tau(i_\ell,j_\ell)} \rangle) }{1+ \exp(\langle w^*, U_{i_\ell}^{\tau(i_\ell,j_\ell)} - U_{j_\ell}^{\tau(i_\ell,j_\ell)} \rangle)}, y_\ell \in [0,1]\\
    & \leq \frac{1}{m} \max_{(i,j) \in P} \|U_{i}^{\tau(i,j)} - U_{j}^{\tau(i,j)}\|_{\infty} \\
    & = \frac{\beta}{m},
\end{align} 

by definition of $\beta$.

Third, $\mathbb{E}((X_\ell^{(k)})^2) \leq \frac{\beta^2}{m^2}$. Let $p(x) = \frac{e^x}{1+e^x}.$ For $k \in [d]$,
\begin{align}
    & \mathbb{E}((X_\ell^{(k)})^2) \\ 
    &= \frac{1}{ \binom{n}{2}} \sum_{(i,j) \in P} \mathbb{E} ((X_\ell^{(k)})^2| \text{items } i,j \text{ are chosen} ) \\
    &= \frac{1}{ \binom{n}{2}} \sum_{(i,j) \in P} \frac{1}{m^2}\left( (U_{i}^{\tau(i,j)})^{(k)} - (U_{j}^{\tau(i,j)})^{(k)} \right)^2 \mathbb{E}\left( \left( p(\langle w^*, U_{i}^{\tau(i,j)} - U_{j}^{\tau(i,j)} \rangle) - y_{(i,j)} \right)^2 \right) \\
    & = \frac{1}{ m^2 \binom{n}{2}} \sum_{(i,j) \in P} \left( (U_{i}^{\tau(i,j)})^{(k)} - (U_{j}^{\tau(i,j)})^{(k)} \right)^2 \\
    & \left( p(\langle w^*, U_{i}^{\tau(i,j)} - U_{j}^{\tau(i,j)} \rangle)^2  -2 \mathbb{E}(y_{(i,j)})p(\langle w^*, U_{i}^{\tau(i,j)} - U_{j}^{\tau(i,j)} \rangle) + \mathbb{E}((y_{(i,j)})^2) \right) \\ 
    & =  \frac{1}{ m^2 \binom{n}{2}} \sum_{(i,j) \in P} \left( (U_{i}^{\tau(i,j)})^{(k)} - (U_{j}^{\tau(i,j)})^{(k)} \right)^2  \left( - \left( p(\langle w^*, U_{i}^{\tau(i,j)} - U_{j}^{\tau(i,j)} \rangle) \right)^2  + \mathbb{E}((y_{(i,j)})^2) \right) \\
    & =  \frac{1}{ m^2 \binom{n}{2}} \sum_{(i,j) \in P} \left( (U_{i}^{\tau(i,j)})^{(k)} - (U_{j}^{\tau(i,j)})^{(k)} \right)^2  \left( - p(\langle w^*, U_{i}^{\tau(i,j)} - U_{j}^{\tau(i,j)} \rangle)^2  + \mathbb{E}(y_{(i,j)}) \right) \text{ since } y_{(i,j)} \in \{0,1\}\\
    & =  \frac{1}{ m^2 \binom{n}{2}} \sum_{(i,j) \in P}  \left( (U_{i}^{\tau(i,j)})^{(k)} - (U_{j}^{\tau(i,j)})^{(k)} \right)^2 \left(p(\langle w^*, U_{i}^{\tau(i,j)} - U_{j}^{\tau(i,j)} \rangle) - p(\langle w^*, U_{i}^{\tau(i,j)} - U_{j}^{\tau(i,j)} \rangle)^2 \right) \\
    & \leq  \frac{\beta^2}{4m^2} \label{eq:betabound}
\end{align} 

by definition of $\beta$ and since $p(\langle w^*, U_{i}^{\tau(i,j)} - U_{j}^{\tau(i,j)} \rangle) \in [0,1]$ and $x - x^2 \leq \frac{1}{4}$ for $x \in [0,1]$.

Therefore, $\frac{1}{m}\nabla \mathcal{L}_m(w^*) = \sum_{\ell = 1}^m X_\ell$ is a sum of i.i.d. mean zero random variables. Hence, each coordinate is also a sum of i.i.d. random variables with mean zero, so Bernstein's inequality applies. Recall Bernstein's inequality:
\begin{theo}[Bernstein's inequality]\label{thm:bernsteinScalar}
Let $X_i$ be i.i.d. random variables such that $\mathbb{E}(X_i) = 0$ and $|X_i| \leq M$. Then for any $t > 0$,
\[\mathbb{P}\left(\sum_{i = 1}^m X_i > t\right) \leq \exp\left(-\frac{\frac{1}{2} t^2}{ \sum \mathbb{E} X_i^2 + \frac{1}{3}Mt} \right).\]
\end{theo}

We apply Bernstein's inequality to the $k$-th coordinate of $\frac{1}{m}\nabla \mathcal{L}_m(w^*)$:
\begin{align}
\mathbb{P}\left( \left| \frac{1}{m} \nabla  \mathcal{L}_m(w^*)^{(k)}\right| > t\right) \leq 2\exp\left(-\frac{\frac{1}{2} t^2}{ \frac{\beta^2}{4m} + \frac{\beta t}{3m}} \right)    \label{eq:applyBernstein}
\end{align}

since $\sum_{\ell = 1}^m \mathbb{E}((X_{\ell}^{(k)})^2) \leq \frac{\beta^2}{4m}$ and $|X_\ell^{(k)}| \leq \frac{\beta}{m}$.

Since $\|x\|_2 \leq \sqrt{d} \|x\|_{\infty}$ for any $x \in \mathbb{R}^d$, 
\begin{align}
    & \mathbb{P}\left( \left\|\frac{1}{m}\nabla \mathcal{L}_m(w^*)\right\|_{2} > t\right) \\
    & \leq \mathbb{P} \left( \frac{\sqrt{d}}{m}  \left\|\nabla \mathcal{L}_m(w^*)\right\|_{\infty} > t\right) \\
    & = \mathbb{P}\left( \left\| \frac{1}{m}\nabla \mathcal{L}_m(w^*)\right\|_{\infty} > \frac{t}{\sqrt{d} }\right)\\ 
    & \leq 2d\exp\left(-\frac{\frac{1}{2} \frac{t^2}{d }}{ \frac{\beta^2}{4m} + \frac{\beta\frac{t}{\sqrt{d} }}{3m}} \right) \text{ by union bound and inequality \eqref{eq:applyBernstein}} \\
    & =  2d\exp\left(-\frac{ t^2}{ \frac{ d \beta^2}{2m} + \frac{2\beta t \sqrt{d}}{3m}}\right) \\
    & = 2d\exp\left(-\frac{ 6m t^2}{ 3 d \beta^2 + 4\beta t \sqrt{d} }\right).
\end{align}

In other words, for $t>0$, with probability at least $1-2d\exp\left(-\frac{ 6m t^2}{ 3 d \beta^2 + 4\beta t \sqrt{d} }\right)$, $\|\frac{1}{m} \nabla \mathcal{L}_m(w^*)\|_{2} \leq t$. 

Let \[\alpha := 3\beta^2\log{(4d / \delta)}d + 4\sqrt{d}\beta \log{(4d / \delta)}.\] Set \[t = \sqrt{\frac{\alpha}{6m}}.\] If \[m \geq \frac{3\beta^2\log{(4d / \delta)}d + 4\sqrt{d}\beta \log{(4d / \delta)}}{6} = \frac{\alpha}{6},\] then
\[2d\exp\left(-\frac{ 6m t^2}{ 3 d \beta^2 + 4\beta t \sqrt{d}}\right) \leq \frac{\delta}{2},\] which we establish below.

If
\begin{align}
& \ \ \ m \geq \frac{\alpha}{6} \\ 
& \implies m \geq \frac{\alpha (4\beta\log{(4d / \delta)})^2d}{6(4\beta\log{(4d / \delta)})^2d} \\
& \implies m \geq \frac{\alpha (4\beta\log{(4d / \delta)})^2d}{6(\alpha - 3 \beta^2 \log{(4d / \delta)}d)^2} \\ 
& \implies m \geq \frac{\alpha d}{6\left(\frac{\alpha - 3 \beta^2 \log{(4d / \delta)}d}{4\beta\log{(4d / \delta)}}\right)^2} \\ 
& \implies \left(\frac{\alpha - 3 \beta^2 \log{(4d / \delta)}d}{4\beta\log{(4d / \delta)}}\right)^2 \geq \frac{\alpha d}{6m} \\ 
& \implies \frac{\frac{\alpha}{\log{(4d / \delta)}} - 3 \beta^2 d}{4\beta} \geq \sqrt{\frac{\alpha d}{6m}} \\ 
& \implies \frac{\alpha}{\log{(4d / \delta)}} \geq 4\beta\sqrt{\frac{\alpha d}{6m}} + 3 \beta^2 d \\ 
& \implies \frac{\alpha}{ 4\beta\sqrt{\frac{\alpha d}{6m}} + 3 \beta^2 d } \geq \log{(4d / \delta)}  \\ 
& \implies \frac{t^2 6m}{ 4\beta t \sqrt{d} + 3 \beta^2 d } \geq \log{(4d / \delta)}  \\ 
& \implies 2 d \exp\left( -\frac{6mt^2}{ 4\beta t \sqrt{d} + 3 \beta^2 d } \right) \leq \frac{\delta}{2}  \\ 
\end{align}

Therefore, if \[m \geq \frac{3\beta^2\log{(4d / \delta)}d + 4\sqrt{d}\beta \log{(4d / \delta)}}{6}\] with probability at least $1-\frac{\delta}{2}$, 

\[\left\|\frac{1}{m} \nabla \mathcal{L}_m(w^*)\right \|_{2} < \sqrt{\frac{3\beta^2\log{(4d / \delta)}d + 4\sqrt{d}\beta \log{(4d / \delta)}}{6m}}.\]
\end{proof}

\begin{lemma}\label{lem:randomMatrixLowerBound}
For $(i,j) \in P$, let $Z_{(i,j)} =(U_{i}^{\tau(i,j)} - U_{j}^{\tau(i,j)}) (U_{i}^{\tau(i,j)} - U_{j}^{\tau(i,j)})^T.$
Let \[\lambda := \lambda_{\min}( \mathbb{E} Z_{(i,j)})\] where for a square matrix $U$, $\lambda_{\min}(U)$ is the smallest eigenvalue of $U$. Let \[\eta := \sigma_{\max}(\mathbb{E}((Z_{(i,j)} - \mathbb{E}Z_{(i,j)})^2))\] where $\sigma_{\max}(X)$ is the largest singular value of a matrix $X$. Let \[\zeta := \max_{(i,j) \in P} \lambda_{\max}(\mathbb{E} Z_{(i,j) } - Z_{(i,j) }),\] where $\lambda_{\max}(X)$ is the largest eigenvalue of $X$.  The expectation in $\lambda$, $\eta$, and $\zeta$ is taken with respect to a uniformly chosen random pair of items. 

Let $\delta > 0.$ Under the model assumptions in this section, if $\lambda >0$ and if \[m \geq \frac{8\log(2 / \delta)(6\eta + \lambda \zeta)}{3\lambda^2},\] then with probability at least $1-\frac{\delta}{2},$
\[\frac{1}{m} \sum_{\ell = 1}^m  \Delta^T (U_{i_\ell}^{\tau(i_\ell,j_\ell)} - U_{j_\ell}^{\tau(i_\ell,j_\ell)}) (U_{i_\ell}^{\tau(i_\ell,j_\ell)} - U_{j_\ell}^{\tau(i_\ell,j_\ell)})^T \Delta \geq \|\Delta\|_2^2\frac{\lambda}{2}\]
where 
\[\Delta = \hat{w} - w^*.\]
\end{lemma}

\begin{proof}
Let \[X_\ell = \frac{1}{m} (U_{i_\ell}^{\tau(i_\ell,j_\ell)} - U_{j_\ell}^{\tau(i_\ell,j_\ell)}) (U_{i_\ell}^{\tau(i_\ell,j_\ell)} - U_{j_\ell}^{\tau(i_\ell,j_\ell)})^T - \frac{1}{m} \mathbb{E}((U_{i}^{\tau(i,j)} - U_{j}^{\tau(i,j)}) (U_{i}^{\tau(i,j)} - U_{j}^{\tau(i,j)})^T)).\] 

Notice that $\frac{1}{m} \sum_{\ell = 1}^m (U_{i_\ell}^{\tau(i_\ell,j_\ell)} - U_{j_\ell}^{\tau(i_\ell,j_\ell)}) (U_{i_\ell}^{\tau(i_\ell,j_\ell)} - U_{j_\ell}^{\tau(i_\ell,j_\ell)})^T $ is a sum of random matrices where the randomness is from the random pairs of items that are chosen in the samples. Therefore, bounding the smallest eigenvalue of this random matrix is sufficient to get the desired lower bound as we show.

Since $\mathbb{E} X_\ell = 0$ by construction and $X_\ell$ is self-adjoint since it is symmetric and real, we apply the following concentration bound to $\sum_{\ell=1}^m X_\ell$:

\begin{theo}[Theorem 1.4 in \cite{Tropp2012}]
\label{thm:joelConcentration}
Consider a finite sequence $\{X_k\}$ of independent, random, self-adjoint matrices with dimension $d$. Assume that each random matrix satisfies $\mathbb{E}X_k = 0$ and $\lambda_{\max}(X_k) \leq R$ almost surely. Then for all $t \geq 0$
\begin{align}
\mathbb{P}\left( \lambda_{\max} \left(\sum_k X_k \right) \geq t \right) \leq d \exp\left( \frac{-t^2/2}{\sigma^2 + Rt/3}\right),
\end{align}
where \[\sigma^2:= \sigma_{\max}\left(\sum_k \mathbb{E}\left(X_k^2\right) \right).\]
\end{theo} 

Notice
\begin{align}
    \sigma_{\max}\left( \sum_{\ell=1}^{m} \mathbb{E}\left(X_\ell^2\right) \right) & = m \sigma_{\max}(\mathbb{E}\left(X_1^2\right) ) \text{ since each } X_\ell \text{ is distributed the same }\\
    & = \frac{m}{m^2} \eta \\ 
    & = \frac{1}{m} \eta.
\end{align}

Then applying the above theorem, for $t\geq 0,$
\begin{align}
    \mathbb{P}\left( \lambda_{\max} \left(\sum_{\ell=1}^m -X_\ell \right) \geq t \right) & \leq d \exp\left( \frac{-t^2/2}{\eta / m + \zeta t/(3m)}\right)\\
    & \leq d \exp\left( \frac{-3mt^2}{6\eta + 2 \zeta t}\right).
\end{align}

In other words, for all $t\geq0$, with probability at least $1-d \exp\left( \frac{-3mt^2}{6\eta + 2 \zeta t}\right),$ 
\begin{align}
&\lambda_{\max} \left(\sum_{\ell=1}^m -X_\ell \right) \leq t \\
& \implies \frac{\Delta^T}{\|\Delta\|_2} \left(\sum_{\ell=1}^m - X_\ell \right) \frac{\Delta}{\|\Delta\|_2}  \leq t \\
\implies & \Delta^T ( \mathbb{E}((U_{i}^{\tau(i,j)} - U_{j}^{\tau(i,j)}) (U_{i}^{\tau(i,j)} - U_{j}^{\tau(i,j)})^T)) -  \nonumber \\
& \frac{1}{m} \sum_{\ell = 1}^m (U_{i_\ell}^{\tau(i_\ell,j_\ell)} - U_{j_\ell}^{\tau(i_\ell,j_\ell)}) (U_{i_\ell}^{\tau(i_\ell,j_\ell)} - U_{j_\ell}^{\tau(i_\ell,j_\ell)})^T  ) \Delta  \leq t \|\Delta\|_2^2\\
\implies & \Delta^T \left( \mathbb{E}((U_{i}^{\tau(i,j)} - U_{j}^{\tau(i,j)}) (U_{i}^{\tau(i,j)} - U_{j}^{\tau(i,j)})^T)) \right)\Delta - t \|\Delta\|_2^2 \nonumber \\
 & \leq \Delta^T \left(\frac{1}{m}\sum_{\ell = 1}^m (U_{i_\ell}^{\tau(i_\ell,j_\ell)} - U_{j_\ell}^{\tau(i_\ell,j_\ell)}) (U_{i_\ell}^{\tau(i_\ell,j_\ell)} - U_{j_\ell}^{\tau(i_\ell,j_\ell)})^T  \right) \Delta \\ 
 \implies & \|\Delta\|_2^2 \frac{\Delta^T}{\|\Delta\|_2} \left( \mathbb{E}((U_{i}^{\tau(i,j)} - U_{j}^{\tau(i,j)}) (U_{i}^{\tau(i,j)} - U_{j}^{\tau(i,j)})^T)) \right) \frac{\Delta}{{\|\Delta\|_2}} - t \|\Delta\|_2^2 \nonumber \\
 & \leq \Delta^T \left(\frac{1}{m}\sum_{\ell = 1}^m (U_{i_\ell}^{\tau(i_\ell,j_\ell)} - U_{j_\ell}^{\tau(i_\ell,j_\ell)}) (U_{i_\ell}^{\tau(i_\ell,j_\ell)} - U_{j_\ell}^{\tau(i_\ell,j_\ell)})^T  \right) \Delta \\ 
& \implies (\lambda  -  t)  \|\Delta\|_2^2  \leq \Delta^T \left(\frac{1}{m}\sum_{\ell = 1}^m (U_{i_\ell}^{\tau(i_\ell,j_\ell)} - U_{j_\ell}^{\tau(i_\ell,j_\ell)}) (U_{i_\ell}^{\tau(i_\ell,j_\ell)} - U_{j_\ell}^{\tau(i_\ell,j_\ell)})^T  \right) \Delta \label{eq:lowerBoundWitht}
\end{align}
since $ \lambda := \lambda_{\min}( \mathbb{E}((U_{i}^{\tau(i,j)} - U_{j}^{\tau(i,j)}) (U_{i}^{\tau(i,j)} - U_{j}^{\tau(i,j)})^T)).$

Set $t = \frac{\lambda}{2}$. Since $\lambda > 0$ by assumption, Equation \eqref{eq:lowerBoundWitht} becomes 
\[\frac{\lambda}{2}  \|\Delta\|_2^2  \leq \Delta^T \left(\frac{1}{m} \sum_{\ell = 1}^m (U_{i_\ell}^{\tau(i_\ell,j_\ell)} - U_{j_\ell}^{\tau(i_\ell,j_\ell)}) (U_{i_\ell}^{\tau(i_\ell,j_\ell)} - U_{j_\ell}^{\tau(i_\ell,j_\ell)})^T  \right) \Delta \] and holds with probability at least $1 - \frac{\delta}{2}$ if \[m \geq \frac{8\log(2d / \delta)(6\eta + \lambda \zeta)}{3\lambda^2}\] since

\begin{align}
    d \exp\left( \frac{-3m\frac{\lambda^2}{4}}{6\eta + 2 \frac{\lambda}{2} \zeta t}\right) &\leq \frac{\delta}{2} \\ 
    \implies \frac{-3m\frac{\lambda^2}{4}}{6\eta + 2 \frac{\lambda}{2} \zeta t} &\leq -\log(2d / \delta) \\ 
    \implies \frac{3m\frac{\lambda^2}{4}}{6\eta + 2 \frac{\lambda}{2} \zeta t} &\geq 2\log(2d / \delta) \\ 
    \implies m &\geq \frac{8\log(2d / \delta)(6\eta + \lambda \zeta)}{3\lambda^2}. \\ 
\end{align}
\end{proof}

\begin{lemma}[Gradient and Hessian of Equation \eqref{eq:logLoss}]\label{lem:gradientHessian}

Given samples $S_m$, features of the $n$ items $U \in \mathbb{R}^{d \times n}$, and $w \in \mathbb{R}^d$, 
 \begin{align}
     & \frac{1}{m} \nabla \mathcal{L}_m(w; U, S_m, \tau) \\
     & = \frac{1}{m} \sum_{\ell = 1}^m \frac{\exp(\langle w, U_{i_\ell}^{\tau(i_\ell,j_\ell)} - U_{j_\ell}^{\tau(i_\ell,j_\ell)} \rangle)}{1+ \exp(\langle w, U_{i_\ell}^{\tau(i_\ell,j_\ell)} - U_{j_\ell}^{\tau(i_\ell,j_\ell)} \rangle)} \left(U_{i_\ell}^{\tau(i_\ell,j_\ell)} - U_{j_\ell}^{\tau(i_\ell,j_\ell)}\right) - y_\ell \left(U_{i_\ell}^{\tau(i_\ell,j_\ell)} - U_{j_\ell}^{\tau(i_\ell,j_\ell)}\right)
 \end{align} and 
 \begin{align}
     & \frac{1}{m} H_{\mathcal{L}_m}(w; U, S_m, \tau) \\
     & = \frac{1}{m} \sum_{\ell = 1}^m \frac{\exp(\langle w, U_{i_\ell}^{\tau(i_\ell,j_\ell)} - U_{j_\ell}^{\tau(i_\ell,j_\ell)} \rangle)}{(1 + \exp(\langle  w, U_{i_\ell}^{\tau(i_\ell,j_\ell)} - U_{j_\ell}^{\tau(i_\ell,j_\ell)} \rangle))^2} (U_{i_\ell}^{\tau(i_\ell,j_\ell)} - U_{j_\ell}^{\tau(i_\ell,j_\ell)}) (U_{i_\ell}^{\tau(i_\ell,j_\ell)} - U_{j_\ell}^{\tau(i_\ell,j_\ell)})^T
 \end{align}
\end{lemma}
\begin{proof}
\textbf{Gradient:}
Let $f(x) := \log(1+ e^x)$ for $x \in \mathbb{R}$ and $g(w; y) := \langle w, y \rangle$ for $w,y \in \mathbb{R}^d$, so \[ \frac{1}{m} \mathcal{L}_m(w; U, S_m, \tau) = \frac{1}{m} \sum_{\ell = 1}^m (f \circ g)(w; U_{i_\ell}^{\tau(i_\ell,j_\ell)} - U_{j_\ell}^{\tau(i_\ell,j_\ell)}) + y_\ell g(w; U_{i_\ell}^{\tau(i_\ell,j_\ell)} - U_{j_\ell}^{\tau(i_\ell,j_\ell)}).\] Note \[ f'(x) = \frac{e^x}{1+e^x} \] and $\nabla_w g(w;y) = y$.

We arrive at the desired result by the chain rule: 
\begin{align}
    & \frac{1}{m} \mathcal{L}_m(w; U, S_m, \tau) = \\
    & \frac{1}{m} \sum_{\ell = 1}^m f'(g(w; U_{i_\ell}^{\tau(i_\ell,j_\ell)} - U_{j_\ell}^{\tau(i_\ell,j_\ell)})) \nabla_w g(w; U_{i_\ell}^{\tau(i_\ell,j_\ell)} - U_{j_\ell}^{\tau(i_\ell,j_\ell)}) - y_\ell \nabla_w g(w; U_{i_\ell}^{\tau(i_\ell,j_\ell)} - U_{j_\ell}^{\tau(i_\ell,j_\ell)}).
\end{align}
\textbf{Hessian:}
Note \[ f''(x) = \frac{e^x(1+e^x) - e^{2x}}{(1+e^x)^2} = \frac{e^x}{(1+e^x)^2}. \]
Let $[H_{\mathcal{L}_m}(w; U, S_m)]_{k} $ be the $k$th row of the Hessian and $\nabla \mathcal{L}_m(w; U, S_m)^{(k)}$ be the $k$th entry of the gradient. Then by the chain rule again, 
\begin{align*}
    & [H_{\mathcal{L}_m}(w; U, S_m)]_{k}^T \\
    &= \nabla_w (\nabla \mathcal{L}_m(w; U, S_m)^{(k)}) \\
    & = \sum_{\ell = 1}^m ((U_{i_\ell}^{\tau(i_\ell,j_\ell)})^{(k)} - (U_{j_\ell}^{\tau(i_\ell,j_\ell)})^{(k)}) f''(g(w; U_{i_\ell}^{\tau(i_\ell,j_\ell)} - U_{j_\ell}^{\tau(i_\ell,j_\ell)})) \nabla_w g(w; U_{i_\ell}^{\tau(i_\ell,j_\ell)} - U_{j_\ell}^{\tau(i_\ell,j_\ell)})  \\
    & = \sum_{\ell = 1}^m \frac{\exp(\langle w, U_{i_\ell}^{\tau(i_\ell,j_\ell)} - U_{j_\ell}^{\tau(i_\ell,j_\ell)} \rangle)}{(1 + \exp(\langle  w, U_{i_\ell}^{\tau(i_\ell,j_\ell)} - U_{j_\ell}^{\tau(i_\ell,j_\ell)} \rangle))^2} ((U_{i_\ell}^{\tau(i_\ell,j_\ell)})^{(k)} - (U_{j_\ell}^{\tau(i_\ell,j_\ell)})^{(k)}) (U_{i_\ell}^{\tau(i_\ell,j_\ell)} - U_{j_\ell}^{\tau(i_\ell,j_\ell)}),
\end{align*} which proves the claim.

\end{proof}

\begin{lemma}
\label{lem:symmetricAndDecreasing}
Let $h(x) = \frac{e^x}{(1+e^x)^2}$. Then $h(x)$ is symmetric and decreases on $[0,\infty)$.
\end{lemma}
\begin{proof}
Symmetry:
\begin{align}
    h(-x) & = \frac{e^{-x}}{(1+e^{-x})^2} \\ 
    & = \frac{e^{-x}}{e^{-2x} (e^{x}+1)^2} \\ 
    & = \frac{e^x}{ (e^{x}+1)^2} \\ 
    & = h(x).
\end{align}

Decreasing on $[0,\infty)$:

Note \begin{align}
    h'(x) &= \frac{e^{x}(1+e^{x})^2  -e^{2x}2(1+e^{x})}{(1+e^{x})^4} \\
    & = \frac{e^{x}(1+e^{x})  -e^{2x}2}{(1+e^{x})^3} \\
    & = \frac{e^{x}(1-e^{x})}{(1+e^{x})^3} \\ 
    & \leq 0
\end{align}
for  $x \in [0, \infty)$ since on this interval, $1-e^x \leq 0$ but $e^x, (1+e^x)^3 \geq 0$. Thus $h(x)$ is decreasing on $[0, \infty)$.
\end{proof}

\section{Specific Selection Functions: Proofs of Corollaries \ref{coro:fullFeatures} and \ref{coro:MaxCoordCoro}}
\label{appendix:specificSelectionFunctions}
In this section, we present the full lower bounds on the number of samples and upper bound on the estimation error. The definitions of the constants that appear in the main text, i.e.  $C_3$ and $C_4$, appear at the end of the applicable proofs. 

\subsection{Proof of Corollary \ref{coro:fullFeatures}}

The following lemma is a straight forward generalization from \cite{rankCentrality}, but we include the proof for completeness. We need this lemma to prove Corollary \ref{coro:fullFeatures}.
\begin{lemma}
\label{lem:fullFeatures}
Let $U \in \mathbb{R}^{d \times n}.$ Assume that the columns of $U$ sum to 0: $\sum_{i=1}^n U_i = 0$. Then

\[\mathbb{E}((U_{i} - U_{j}) (U_{i} - U_{j})^T) = \frac{n}{\binom{n}{2}}UU^T \]

where the expectation is with respect to a uniformly at randomly chosen pair of items.
\end{lemma}
\begin{proof}
Let $e_i \in \mathbb{R}^n$ denote the $i$-th standard basis vector, $I_{n\times n}$ denote the $n \times n$ identity matrix, and $\mathbbm{1} \in \R^n$ be the vector of all ones. Since the expectation is over a uniformly chosen pair of items $(i,j) \in P$,
\begin{align}
    & \mathbb{E}((U_{i} - U_{j}) (U_{i} - U_{j})^T) \\
    &= \mathbb{E}(U(e_i - e_j) (e_i-e_j)^TU^T) \\ 
    &= \frac{1}{\binom{n}{2}} U \left(\sum_{(i,j) \in P } e_i e_i^T - e_i e_j^T - e_j e_i^T + e_j e_j^T \right)U^T \\
    & =  \frac{1}{\binom{n}{2}} U \left((n-1) \sum_{i=1}^n  e_i e_i^T -  \sum_{(i,j) \in P }e_i e_j^T + e_j e_i^T \right)U^T \text{ since each item is in } n-1 \text{ comparisons}\\ 
    & = \frac{1}{\binom{n}{2}} U \left((n-1) I_{n \times n} -  \sum_{(i,j) \in P }e_i e_j^T + e_j e_i^T\right)U^T \\     
    & =  \frac{1}{\binom{n}{2}} U\left((n-1) I_{n \times n} - \left( \mathbbm{1}\mathbbm{1}^T - I_{n \times n}\right) \right)U^T \label{eq:explainThisEq} \text{ explained below} \\  
    & = \frac{1}{\binom{n}{2}} U \left( nI_{n\times n}-\mathbbm{1}\mathbbm{1}^T   \right) U^T \\
      &=\frac{1}{\binom{n}{2}}(nUU^T-U\mathbbm{1}\mathbbm{1}^TU^T)\\
  &=\frac{n}{\binom{n}{2}}UU^T \text{since } U\mathbbm{1} = \sum_{i=1}^n U_i = \mathbf{0} \text{ by assumption.} \label{eq:expectationLast}
\end{align}  

Equation \eqref{eq:explainThisEq} is because $e_i e_j^T$ is the matrix with a 1 in the $i$-th row and $j$-th column and 0 elsewhere and we are summing over all $(i,j) \in [n] \times [n]$ where $i < j$. Thus, the sum equals $\mathbbm{1}\mathbbm{1}^T - I_{n \times n}$, which is the matrix with ones everywhere except for the diagonal.
\end{proof}

\begin{corollary}[Restatement of Corollary \ref{coro:fullFeatures}]
Assume the set-up stated in the beginning of Section \ref{sec:modelFormulation}. For the selection function $\tau$, suppose $\tau(i,j) = [d]$ for any $(i,j) \in P$. In other words, all the features are used in each pairwise comparison. Assume $n >d$. Let $\nu := \max\{\max_{(i,j) \in P} \|U_i - U_j\|_2^2, 1\}$. Without loss of generality, assume the columns of $U$ sum to zero: $\sum_{i = 1}^n U_i =0$. Then,
$$\lambda = \frac{n\lambda_{\min}(UU^T)}{\binom{n}{2}},$$
$$\zeta \leq \nu + \frac{n\lambda_{\max}(UU^T)}{\binom{n}{2}},$$ and 
    $$\eta \leq \frac{\nu n \lambda_{\max}(UU^T)}{\binom{n}{2}} + \frac{n^2 \lambda_{\max}(UU^T)^2}{\binom{n}{2}^2}.$$ 
Let $$m_1 = \frac{3\beta^2\log{(4d / \delta)}d + 4\sqrt{d}\beta \log{(4d / \delta)}}{6}.$$ Let $\delta >0$. Hence, if 

\begin{align*}
m \geq \max & \left\{  m_1,\frac{48\log(2d / \delta) \binom{n}{2}^2}{3 n^2 \lambda_{\min}(UU^T)^2} \left(\frac{\nu n \lambda_{\max}(UU^T)}{\binom{n}{2}} + \frac{n^2 \lambda_{\max}(UU^T)^2}{\binom{n}{2}^2} \right) + \frac{8\log(2d / \delta) \binom{n}{2}}{3n \lambda_{\min}(UU^T)}\left(\nu + \frac{n\lambda_{\max}(UU^T)}{\binom{n}{2}}\right) \right\},
\end{align*}
then with probability at least $1 - \delta$, 
\begin{align} 
\| w^* - \hat{w}\|_2 & \leq  \frac{4(1 + \exp(b^*))^2 \binom{n}{2}}{\exp(b^*) n \lambda_{\min}(UU^T)}  \sqrt{\frac{3\beta^2\log{(4d / \delta)}d + 4\sqrt{d}\beta \log{(4d / \delta)}}{6m}}.
\end{align}
\end{corollary}
\begin{proof}
Throughout this proof, we use $U_i$ instead of $U_i^{\tau(i,j)}$ for any items $i,j$ since $\tau(i,j)$ selects all coordinates.

If $\sum_{i = 1}^n U_i \neq 0$, simply subtract the column mean, $\bar{U} := \frac{1}{n} \sum_{i = 1}^n U_i$,  from each column. This operation does not affect the underlying pairwise probabilities since
\begin{align} 
\mathbb{P}(\text{item } i \text{ beats item } j) & = \frac{1}{1 + \exp(-\langle w^*, U_i - U_j \rangle)} \\
&= \frac{1}{1 + \exp(-\langle w^*, (U_i - \bar{U}) - (U_j -\bar{U}) \rangle)}.
\end{align}
Let $\widetilde{U} = U(I - \frac{1}{n} \mathbbm{1}\mathbbm{1}^T)$ be the centered version of $U$, i.e. where we subtract $\bar{U}$ from each column of $U$. Since $n>d$ and by Proposition \ref{prop:centeredU}, if $\lambda_{\min}(U) >0$, then   $\lambda_{\min}(\widetilde{U}) >0$ generically. Therefore, WLOG, we may assume $\sum_{i = 1}^n U_i = 0$.

First, we simplify $\lambda$. By Lemma \ref{lem:fullFeatures},  \[ \lambda = \lambda_{\min}(\mathbb{E}((U_{i} - U_{j}) (U_{i} - U_{j})^T)) = \frac{n\lambda_{\min}(UU^T)}{\binom{n}{2}}.\]

Second, we upper bound $\zeta$. Let $(k,\ell) \in P$, then
\begin{align}
    & \lambda_{\max} \left(\mathbb{E} (U_i - U_j)(U_i - U_j)^T  - (U_k - U_{\ell})(U_k - U_{\ell})^T \right) \\ 
   & =  \lambda_{\max}\left(\frac{n}{\binom{n}{2}} UU^T - (U_k - U_{\ell})(U_k - U_{\ell})^T \right)    \text{ by Lemma \ref{lem:fullFeatures}} \\ 
    & \leq \lambda_{\max}\left(\frac{n}{\binom{n}{2}} UU^T\right) + \lambda_{\max}\left( (U_k - U_{\ell})(U_k - U_{\ell})^T \right)   \\     
    & = \lambda_{\max}\left(\frac{n}{\binom{n}{2}} UU^T\right) + \|(U_k - U_{\ell})\|_2^2 \\     
    & \leq \lambda_{\max}\left(\frac{n}{\binom{n}{2}} UU^T\right) + \nu, \\     
\end{align}
where the second to last line is since the largest eigenvalue of a rank one matrix $xx^T$ is $\|x\|_2^2$ and the last line is by definition of $\nu$.
 
Third, we upper bound $\eta$. Let $e_i \in \mathbb{R}^n$ denote the $i$-th standard basis vector. For any random variable $X$, we have 
\begin{align}
    \mathbb{E}(X- \mathbb{E}(X))^2 = \mathbb{E}(X^2) - \mathbb{E}(X)^2.
\end{align} Furthermore, since $\eta$ is the largest singular value of a symmetric matrix squared, the largest eigenvalue of that matrix is also equal to $\eta$. Therefore, $\eta = \lambda_{\max} \left( \mathbb{E}((U_{i} - U_{j}) (U_{i} - U_{j})^T(U_{i} - U_{j}) (U_{i} - U_{j})^T) -  \mathbb{E}((U_{i} - U_{j}) (U_{i} - U_{j})^T)^2 \right).$ Most steps are explained below after the equations. Because the expectation is with respect to a uniformly at random pair of items $(i,j) \in P$ and by Lemma \ref{lem:fullFeatures},
\begin{align}
 & \lambda_{\max}\left(\mathbb{E}((U_{i} - U_{j}) (U_{i} - U_{j})^T(U_{i} - U_{j}) (U_{i} - U_{j})^T) -  \mathbb{E}((U_{i} - U_{j}) (U_{i} - U_{j})^T)^2\right) \\ 
   &= \lambda_{\max}\left(\frac{1}{\binom{n}{2}} \sum_{(i,j) \in P} (U_{i} - U_{j}) (U_{i} - U_{j})^T(U_{i} - U_{j}) (U_{i} - U_{j})^T - \frac{n^2}{\binom{n}{2}^2} UU^TUU^T\right)\\ 
  &= \lambda_{\max}\left(\frac{1}{\binom{n}{2}} \sum_{(i,j) \in P} \left((U_{i} - U_{j})^T(U_{i} - U_{j})\right) (U_{i} - U_{j})(U_{i} - U_{j})^T - \frac{n^2}{\binom{n}{2}^2} UU^TUU^T\right) \label{eqn:scalar} \\ 
   &= \lambda_{\max}\left(\frac{1}{\binom{n}{2}} \sum_{(i,j) \in P} \left((U_{i} - U_{j})^T(U_{i} - U_{j})\right) U(e_i - e_j)(e_i-e_j)^TU^T - \frac{n^2}{\binom{n}{2}^2} UU^TUU^T\right) \\
   &\leq \lambda_{\max}\left(\frac{1}{\binom{n}{2}} \sum_{(i,j) \in P} \left((U_{i} - U_{j})^T(U_{i} - U_{j})\right) U(e_i - e_j)(e_i-e_j)^TU^T\right) + \lambda_{\max}\left(\frac{n^2}{\binom{n}{2}^2} UU^TUU^T\right) \label{eq:triangle}  \\ 
   &= \max_x \frac{x^T}{\|x\|}\left(\frac{1}{\binom{n}{2}} \sum_{(i,j) \in P} \left((U_{i} - U_{j})^T(U_{i} - U_{j})\right) U(e_i - e_j)(e_i-e_j)^TU^T\right)\frac{x}{\|x\|} + \lambda_{\max}\left(\frac{n^2}{\binom{n}{2}^2} UU^TUU^T\right)   \\
   &= \max_x \left(\frac{1}{\binom{n}{2}} \sum_{(i,j) \in P} \left((U_{i} - U_{j})^T(U_{i} - U_{j})\right)  \frac{x^T}{\|x\|}U(e_i - e_j)(e_i-e_j)^TU^T\frac{x}{\|x\|}\right) + \lambda_{\max}\left(\frac{n^2}{\binom{n}{2}^2} UU^TUU^T\right)  \\
   &\leq \max_x \left(\frac{\nu}{\binom{n}{2}} \sum_{(i,j) \in P}   \frac{x^T}{\|x\|}U(e_i - e_j)(e_i-e_j)^TU^T\frac{x}{\|x\|}\right) + \lambda_{\max}\left(\frac{n^2}{\binom{n}{2}^2} UU^TUU^T\right)  \label{eqn:upperBound}  \\
   & = \lambda_{\max} \left(\frac{\nu}{\binom{n}{2}} \sum_{(i,j) \in P}   U(e_i - e_j)(e_i-e_j)^TU^T\right) + \lambda_{\max}\left(\frac{n^2}{\binom{n}{2}^2} UU^TUU^T\right)   \\
   & = \frac{\nu n}{\binom{n}{2}} \lambda_{\max}\left( UU^T\right) + \frac{n^2}{\binom{n}{2}^2} \lambda_{\max}\left(UU^T\right)^2 \text{by Lemma } \ref{lem:fullFeatures}.   \\
\end{align}

Equation \eqref{eqn:scalar} is because $(U_{i} - U_{j})^T(U_{i} - U_{j}) \in \mathbb{R}$. 
Equation \eqref{eqn:upperBound} is because $(U_{i} - U_{j})^T(U_{i} - U_{j}) \geq 0$ and $\frac{x^T}{\|x\|}U(e_i - e_j)(e_i-e_j)^TU^T\frac{x}{\|x\|} \geq 0$.

Now that we have bounds on $\eta$ and $\zeta$ and a simplified form for $\lambda$, we apply Theorem \ref{thm:sampleComplexity}, completing the proof. 

Now we explain how to get from these results to those in the main paper with the order terms. The $O(\cdot)$ upper bound on the estimation error is easy to see. The value of $C_1$ is given at the end of the proof of Theorem \ref{thm:sampleComplexity}. The only remaining term to explain from the main paper is the upper bound of $\frac{8\log(2d / \delta)(6\eta + \lambda \zeta)}{3\lambda^2}$, which gives us a lower bound on the number of samples required.

In particular, 
\begin{align}
    & \frac{8\log(2d / \delta)(6\eta + \lambda \zeta)}{3\lambda^2} \\ 
    & =   \frac{48\log(2d / \delta)\eta}{3\lambda^2} + \frac{8 \log(2d / \delta) \zeta}{3\lambda} \\ 
    & = \frac{48\log(2d / \delta) \binom{n}{2}^2}{3 n^2 \lambda_{\min}(UU^T)^2} \left(\frac{\nu n \lambda_{\max}(UU^T)}{\binom{n}{2}} + \frac{n^2 \lambda_{\max}(UU^T)^2}{\binom{n}{2}^2} \right) + \frac{8\log(2d / \delta) \binom{n}{2}}{3n \lambda_{\min}(UU^T)}\left(\nu + \frac{n\lambda_{\max}(UU^T)}{\binom{n}{2}}\right) \\
    & = \frac{48\log(2d / \delta) }{3 \lambda_{\min}(UU^T)^2} \left( \frac{\binom{n}{2} \nu \lambda_{\max}(UU^T)}{n} + \lambda_{\max}(UU^T)^2 \right) + \frac{8\log(2d / \delta)}{3 \lambda_{\min}(UU^T)}\left(\frac{\binom{n}{2} \nu}{n} + \lambda_{\max}(UU^T)\right) \\
    & \leq \frac{48\log(2d / \delta) }{3 \lambda_{\min}(UU^T)^2} \left( \frac{\binom{n}{2} \nu \lambda_{\max}(UU^T)}{n} + n \lambda_{\max}(UU^T)^2 \right) + \frac{8\log(2d / \delta)}{3 \lambda_{\min}(UU^T)}\left(\frac{\binom{n}{2} \nu}{n} + n \lambda_{\max}(UU^T)\right) \\
    & \leq \frac{48\log(2d / \delta) }{3 \lambda_{\min}(UU^T)^2} \left( n \nu \lambda_{\max}(UU^T) + n \lambda_{\max}(UU^T)^2 \right) + \frac{48\log(2d / \delta)}{3 \lambda_{\min}(UU^T)}\left(n \nu + n \lambda_{\max}(UU^T)\right) \\
    & \leq \frac{48\log(2d / \delta)n \nu }{3} \left( \frac{\lambda_{\max}(UU^T)}{\lambda_{\min}(UU^T)^2} + \frac{\lambda_{\max}(UU^T)^2}{\lambda_{\min}(UU^T)^2} +  \frac{1}{\lambda_{\min}(UU^T)} + \frac{\lambda_{\max}(UU^T)}{\lambda_{\min}(UU^T)}\right) \text{ since } \nu \geq 1 \\
    & \leq \frac{2*48\log(2d / \delta)n \nu }{3} \left( \frac{\lambda_{\max}(UU^T)}{\lambda_{\min}(UU^T)^2} + \frac{\lambda_{\max}(UU^T)^2}{\lambda_{\min}(UU^T)^2} +  \frac{1}{\lambda_{\min}(UU^T)} \right) \text{ since } \frac{\lambda_{\max}(UU^T)}{\lambda_{\min}(UU^T)}  \geq 1 \\
    & = C_3\log(2d / \delta)n \nu \left( \frac{\lambda_{\max}(UU^T)}{\lambda_{\min}(UU^T)^2} + \frac{\lambda_{\max}(UU^T)^2}{\lambda_{\min}(UU^T)^2} +  \frac{1}{\lambda_{\min}(UU^T)} \right)
\end{align}

where $C_3 = 2*48 / 3$. We remark that the assumption that $\nu \geq 1$ was made to simplify the upper bound and is not required.
\end{proof}

As we mentioned, we can assume $U$ is centered without loss of generality, because we can subtract the mean column from all columns if they are not centered. However one may wonder then what happens to $\lambda_{\min}(UU^T) = \sqrt{\sigma_{\min}(U)}$ once $U$ is centered. Since we assume $n>d$, it will generically be non-zero, as we make precise in the following proposition.

\begin{proposition}
\label{prop:centeredU}
Given an arbitrary rank-$d$, $d \times n$ matrix $
\widetilde U$, let $U$ be its centered version, i.e. $U = \widetilde{U}(I - \frac{1}{n} \mathbbm{1}\mathbbm{1}^T)$. Then
$\sigma_{\min}(U) = 0$ if and only if the all-ones vector is in the row space of $\widetilde U$.
\end{proposition}

\begin{proof}
Suppose $\widetilde U$ contains the all-ones vector in its row space, and therefore let $v$ be such that $\widetilde U^T v = \mathbbm{1}$. Let $Q = (I - \frac{1}{n} \mathbbm{1}\mathbbm{1}^T)$. Then $$U^T v = Q \widetilde U^T v = 0$$ since the all-ones vector is in the nullspace of $Q$, implying that $\sigma_{\min}(U)=0$. For the other direction suppose $\sigma_{\min}(U)=0$. Then there exists a vector $v \neq 0$ such that 
$$0 = U^Tv = Q \widetilde U^T v.$$ This implies either that $\widetilde U^T v = 0$ or $\widetilde U^T v$ is in the nullspace of $Q$. Since we assumed that $\widetilde U$ has full row rank, then it must be that $\widetilde U^T v = \mathbbm{1}$, the only vector in the nullspace of $Q$. 
\end{proof}

\subsection{Discussion of Corollary \ref{coro:fullFeatures} as compared to related work} \label{sec:boundcompare}

While our sample complexity theorem for MLE of the parameters of FBTL is novel to the best of our knowledge, there are some related results that merit a comparison. First, there is a result in \cite{fbtl} that gives sample complexity results for a different estimator of FBTL parameters under a substantially different sampling model. In particular, they only allow pairs to be sampled from a graph, and then for each sampled pair they observe a fixed number of pairwise comparisons. In their results one can see that as the number of pairs sampled increases, their error upper bound increases and the probability of their resulting bound also decreases. In contrast, our analysis shows that our error bound decreases as $m$ increases, and the probability of our resulting bound remains constant. 


Second, we can also attempt a comparison to the bounds for BTL without features in \cite{negahban2012unified}, despite the fact that with standard basis features, our bound does not apply because $\lambda=0$.  Assuming that $\exp(b^*)/\lambda$ is a constant in our bound and that $\nu \bar{\lambda}$ is a constant, we roughly have an error bound of $O(1)$ given $m = \Theta(n^2 (\beta^2 + \beta) d \log (d / \delta))$ samples. The result in \cite{negahban2012unified} instead has that $m = \Theta(d^2 \log d)$ gives an error bound of $O(1)$ with probability $1-\frac{2}{d}$, recalling that in their setting $d=n$. So if we can tighten bounds that require $\beta$ in our proof, our results may compare favorably. 

Recall the definition of $\beta$ in Equation \eqref{eq:betadef}: $\beta:= \max_{(i,j) \in P} \| U_{i}^{\tau(i,j)} - U_{j}^{\tau(i,j)} \|_{\infty}$. In our proof, we use this to bound differences between feature vectors at Equation \eqref{eq:betabound}. In particular, we bound $\frac{1}{ \binom{n}{2}} \sum_{(i,j) \in P}  \left( (U_{i}^{\tau(i,j)})^{(k)} - (U_{j}^{\tau(i,j)})^{(k)} \right)^2\leq \beta^2$. If we instead directly made the assumption that $$\widetilde \beta^2 := \frac{1}{ \binom{n}{2}} \max_{k \in [d]} \sum_{(i,j) \in P}  \left( (U_{i}^{\tau(i,j)})^{(k)} - (U_{j}^{\tau(i,j)})^{(k)} \right)^2 \;,$$ we could replace $\beta$ with $\widetilde \beta$ directly in our bounds. Assume $\widetilde \beta \leq 1/n^2$. Then our sample complexity would reduce to $m = \Theta(d \log(d / \delta)) = \Theta(d \log(2d^2)) = \Theta(d \log(d))$ where recall $\delta = \frac{2}{d}$, beating the complexity in \cite{negahban2012unified}. However, it is not clear in general what impact the assumption that $\widetilde \beta \leq 1/n^2$ would have on the minimum eigenvalue of $UU^T$.  Indeed, the standard basis vectors are a special case where $\widetilde \beta \leq 1/n$, and as we pointed out, for this special case $\lambda=0$.

Third, although there are crucial differences between our model and the model in \cite{shah2017simple} that make a direct comparison impossible, we attempt to roughly compare results. The first difference is that they assume the feature vectors of the items are standard basis vectors, which means our bounds do not apply just as in the comparison with \cite{negahban2012unified}. The second difference, perhaps the most crucial, is that we make different assumptions about how the intransitive pairwise comparisons are related to the ranking. In \cite{shah2017simple}, the items are ranked based on the probability that one items beats any other item chosen uniformly at random. There are scenarios where the true ranking in our model is not the same as the true ranking in \cite{shah2017simple}. The third difference is that we assume that pairs are drawn uniformly at random, whereas they assume each pair $(i,j) \in P$ is drawn $x_{i,j}$ times where $x_{i,j} \sim \mathrm{Binom}(r,p)$ for $r,p >0$. 

Their result (Theorem 2) roughly says with probability $1/n^{13}$, if the gap between a pair of consecutively ranked items' scores is at least $\sqrt{\log n / (npr)}$, then their algorithm learns the ranking exactly. We compare to our Corollary \ref{coro:rankingComplexity} with $k = 1$ and $\delta = \frac{1}{n^{13}}$ though again we emphasize an exact comparison is impossible because our model is not a special case of theirs or vice versa. Our corollary says with enough samples with high probability, we learn the ranking exactly. On average, their sampling method will see $O(n^2 rp)$ samples, so a reasonable way to compare results is to show the required number of samples in our method is comparable to $O(n^2rp)$. If we assume that $\beta, \eta, \zeta, \lambda,$ and $M$ are all constant, $\alpha_k = \sqrt{\log n / (npr)}$ which is their assumed gap between scores, and $d = n$, the number of samples we require is $\max\{n \log(n*n^{13}), \log(n), n \log(n*n^{13}) npr / \log(n) \} = O(n^2pr)$, matching their bounds.

Fourth, the set-up of \cite{heckel2019active} is the same as \cite{shah2017simple} except it considers the adaptive setting. If the gaps of the utilities of consecutively ranked items are constant and denoted by $\Delta$, then under the same assumptions in the discussion about \cite{shah2017simple}, our Corollary \ref{coro:rankingComplexity} is slightly better by a log factor than their Theorem 1a: $O(\log(n / \delta) n / (\Delta)^2))$ vs. $O(\log(n / \delta) n \log(2\log(2/\Delta)) / (\Delta)^2) )$. However, if many gaps between scores are large and only some gaps between scores are small, their adaptive method is better than our Corollary \ref{coro:rankingComplexity}. This is not surprising since they can adaptively chose which pair to sample next based on the past pairwise comparisons, whereas we consider the passive setting.


\subsection{Proof of Corollary \ref{coro:MaxCoordCoro}}

\begin{corollary}[Restatement of Corollary \ref{coro:MaxCoordCoro}]
Assume the set-up stated in the beginning of Section \ref{sec:modelFormulation}. Assume that for any $(i,j) \in P$, $|\tau(i,j)| = 1$. Partition $P = \sqcup_{k=1}^d P_k$ into $d$ sets where $(i,j) \in P_k$ if $\tau(i, j) = \{k\}$ for $k \in [d]$. Let $\epsilon := \min_{(i,j) \in P} \|U_i^{\tau(i,j)} - U_j^{\tau(i,j)}\|_{\infty}.$ Then 
 $$\lambda \geq \frac{\epsilon^2}{ \binom{n}{2}}  \min_{k \in[d]} |P_k|,$$ 
 $$\zeta \leq  \beta^2 + \frac{\beta^2}{ \binom{n}{2}} \max_{k \in [d]} |P_k|,$$ and $$\eta \leq \frac{\beta^4}{ \binom{n}{2}} \max_{k \in [d]} \left( |P_k| + \frac{|P_k|^2}{\binom{n}{2}} \right). $$
 
Furthermore, let \[m_1 = \frac{3\beta^2\log{(4d / \delta)}d + 4\sqrt{d}\beta \log{(4d / \delta)}}{6}\] and let \[m_3 := \frac{ 48 \log(2d / \delta) \beta^4 \max_{k \in [d]} \left( \binom{n}{2} |P_k| + |P_k|^2 \right)}{3 \epsilon^4 \min_{k \in[d]} |P_k|^2} + \frac{8 \log(2d / \delta) \beta^2 \left(\binom{n}{2} + \max_{k \in [d]} |P_k| \right) }{3 \epsilon^2 \min_{k \in[d]} |P_k| }.\] Let $\delta >0$. If $m \geq \max\{m_1, m_3 \}$, then with probability at least $1 - \delta$, \[\| w^* - \hat{w}\|_2 \leq \frac{4(1 + \exp(b^*))^2 \binom{n}{2}}{\exp(b^*) \epsilon^2 \min_{k \in[d]} |P_k| } \sqrt{\frac{3\beta^2\log{(4 d / \delta)}d + 4\sqrt{d}\beta \log{(4 d / \delta)}}{6m}},\]  where the randomness is from the randomly chosen pairs and the outcomes of the pairwise comparisons.
\end{corollary}

\begin{proof}
Note that $|P_k| > 0$, so that $\lambda>0$, for all $k \in [d]$ if the model is identifiable. Let $U_i^{(j)}$ be the $j$-th coordinate of the vector $U_i$, $e_i$ be the $i$-th standard basis vector, and for a vector $x$, let $\mathrm{diag}(x)$ be the diagonal matrix whose $(i,i)$-th entry is the $i$-th entry of $x$.

First we simplify and bound $\lambda$. Since each pair of items are chosen uniformly at random,
\begin{align}
    \mathbb{E}((U_{i}^{\tau(i,j)} - U_{j}^{\tau(i,j)}) (U_{i}^{\tau(i,j)} - U_{j}^{\tau(i,j)})^T) & = \frac{1}{ \binom{n}{2}} \sum_{(i,j) \in P} (U_{i}^{\tau(i,j)} - U_{j}^{\tau(i,j)}) (U_{i}^{\tau(i,j)} - U_{j}^{\tau(i,j)})^T \label{eq:maxCoord1}\\ 
    &= \frac{1}{ \binom{n}{2}} \sum_{k=1}^d \sum_{(i,j) \in P_k} (U_{i}^{\tau(i,j)} - U_{j}^{\tau(i,j)}) (U_{i}^{\tau(i,j)} - U_{j}^{\tau(i,j)})^T \\ 
    &= \frac{1}{ \binom{n}{2}} \sum_{k=1}^d \left(\sum_{(i,j) \in P_k}  (U_i^{(k)} - U_j^{(k)})^2 \right) \mathrm{diag}(e_{k}), \label{eq:maxCoord2}
\end{align}
which is a diagonal matrix. Therefore, 
\begin{align} 
\lambda & = \frac{1}{ \binom{n}{2}} \min_{k \in[d]}  \left(\sum_{(i,j) \in P_k}  (U_i^{(k)} - U_j^{(k)})^2 \right) \\
& \geq \frac{\epsilon^2}{ \binom{n}{2}}  \min_{k \in[d]} |P_k|.
\end{align}

Second, we simplify and bound $\zeta$. Since $|\tau(k,j)|=1$ for all $k,j \in P$, let $U_i^{(\tau(k,j))}$ denote the coordinate of $U_i$ corresponding to the only element in $\tau(k,j)$. Define $e_{{\tau}(k,j)}$ similarly, which is one of the standard basis vectors. From the proof of bounding $\lambda$ in Equations \eqref{eq:maxCoord1} to \eqref{eq:maxCoord2}, we have $\mathbb{E}((U_{i}^{\tau(i,j)} - U_{j}^{\tau(i,j)}) (U_{i}^{\tau(i,j)} - U_{j}^{\tau(i,j)})^T) =  \frac{1}{ \binom{n}{2}} \sum_{k=1}^d \left(\sum_{(i,j) \in P_k}  (U_i^{(k)} - U_j^{(k)})^2 \right) \mathrm{diag}(e_{k})$, so  
\begin{align}
    \zeta &= \max_{(\ell,p) \in P} \lambda_{\max}( \mathbb{E}((U_{i}^{\tau(i,j)} - U_{j}^{\tau(i,j)}) (U_{i}^{\tau(i,j)} - U_{j}^{\tau(i,j)})^T) - (U_{\ell}^{\tau(\ell,p)} - U_{p}^{\tau(\ell,p)}) (U_{\ell}^{\tau(\ell,p)} - U_{p}^{\tau(\ell,p)})^T) \\ 
    & = \max_{(\ell,p) \in P} \lambda_{\max}\left( \frac{1}{ \binom{n}{2}} \sum_{k=1}^d \left(\sum_{(i,j) \in P_k}  (U_i^{(k)} - U_j^{(k)})^2 \right) \mathrm{diag}(e_{k}) - (U_{\ell}^{\tau(\ell,p)} - U_{p}^{\tau(\ell,p)}) (U_{\ell}^{\tau(\ell,p)} - U_{p}^{\tau(\ell,p)})^T\right) \\ 
    & = \max_{(\ell,p) \in P} \lambda_{\max}\left( \frac{1}{ \binom{n}{2}} \sum_{k=1}^d \left(\sum_{(i,j) \in P_k}  (U_i^{(k)} - U_j^{(k)})^2 \right) \mathrm{diag}(e_{k}) - (U_{\ell}^{(\tau(\ell,p))} - U_{p}^{(\tau(\ell,p))})^2 \mathrm{diag}(e_{\tau(\ell,p)})\right)\\ 
    & \leq \beta^2 \left(\max_{k \in [d]} \left( \frac{|P_k|}{ \binom{n}{2}} +1  \right) \right)\\ 
\end{align}
since the maximum eigenvalue of a diagonal matrix is bounded by the absolute value of its largest entry. We have also applied the triangle inequality and the definition of $\beta$ since $|\tau(i,j)|=1$ for all $(i,j) \in P$.

Third, we simplify $\eta$. First notice from the proof of bounding $\lambda$ from Equations \eqref{eq:maxCoord1} to \eqref{eq:maxCoord2},
\begin{align}
\left(\mathbb{E}(U_{i}^{\tau(i,j)} - U_{j}^{\tau(i,j)}) (U_{i}^{\tau(i,j)} - U_{j}^{\tau(i,j)})^T\right)^2 & = \left(\frac{1}{ \binom{n}{2}} \sum_{k=1}^d \left(\sum_{(i,j) \in P_k}  (U_i^{(k)} - U_j^{(k)})^2 \right) \mathrm{diag}(e_{k})\right)^2 \\
&= \frac{1}{\binom{n}{2}^2} \sum_{k=1}^d \left(\sum_{(i,j) \in P_k}  (U_i^{(k)} - U_j^{(k)})^2 \right)^2 \mathrm{diag}(e_{k}),
\end{align}
since the matrices above are diagonal.

Also, 
\begin{align}
    & \mathbb{E}(((U_{i}^{\tau(i,j)} - U_{j}^{\tau(i,j)}) (U_{i}^{\tau(i,j)} - U_{j}^{\tau(i,j)})^T)^2) \\
    & = \mathbb{E}((U_{i}^{\tau(i,j)} - U_{j}^{\tau(i,j)}) (U_{i}^{\tau(i,j)} - U_{j}^{\tau(i,j)})^T(U_{i}^{\tau(i,j)} - U_{j}^{\tau(i,j)}) (U_{i}^{\tau(i,j)} - U_{j}^{\tau(i,j)})^T) \\
    & = \frac{1}{ \binom{n}{2}} \sum_{k=1}^d \sum_{(i,j) \in P_k} (U_{i}^{\tau(i,j)} - U_{j}^{\tau(i,j)}) (U_{i}^{\tau(i,j)} - U_{j}^{\tau(i,j)})^T(U_{i}^{\tau(i,j)} - U_{j}^{\tau(i,j)}) (U_{i}^{\tau(i,j)} - U_{j}^{\tau(i,j)})^T \\ 
    &= \frac{1}{ \binom{n}{2}} \sum_{k=1}^d \left(\sum_{(i,j) \in P_k}  (U_i^{(k)} - U_j^{(k)})^4 \right) \mathrm{diag}(e_{k}),
\end{align}

For any random variable $X$, we have 
\begin{align}
    \mathbb{E}(X- \mathbb{E}(X))^2 = \mathbb{E}(X^2) - \mathbb{E}(X)^2.
\end{align}

Therefore,

\begin{align}
    \eta & = \sigma_{\max}\left(\frac{1}{ \binom{n}{2}} \sum_{k=1}^d \left(\sum_{(i,j) \in P_k}  (U_i^{(k)} - U_j^{(k)})^4 \right) \mathrm{diag}(e_{k}) - \frac{1}{\binom{n}{2}^2} \sum_{k=1}^d \left(\sum_{(i,j) \in P_k}  (U_i^{(k)} - U_j^{(k)})^2 \right)^2 \mathrm{diag}(e_{k})\right) \\
    & = \frac{1}{ \binom{n}{2}} \sigma_{\max}\left( \sum_{k=1}^d \left(\sum_{(i,j) \in P_k}  (U_i^{(k)} - U_j^{(k)})^4 - \frac{1}{\binom{n}{2}} \left(\sum_{(i,j) \in P_k}  (U_i^{(k)} - U_j^{(k)})^2 \right)^2 \right) \mathrm{diag}(e_{k}) \right) \\
    &\leq \frac{\beta^4}{ \binom{n}{2}} \max_{k \in [d]} \left( |P_k| + \frac{|P_k|^2}{\binom{n}{2}} \right)
\end{align}
since the largest singular value of a diagonal matrix is bounded by the largest entry of the diagonal in absolute value. We have also applied the triangle inequality and definition of $\beta$.

The remainder of the corollary follows by applying the bounds on $\lambda, \zeta$ and $\eta$ to Theorem \ref{thm:sampleComplexity}. 

Now we explain how to get from these results to those in the main paper with the order terms. The $O(\cdot)$ upper bound on the estimation error is easy to see. The value of $C_1$ is given at the end of the proof of Theorem \ref{thm:sampleComplexity}. Finally, it is easy to see $C_4 = 48/3$ in the main paper.
\end{proof}

\subsection{Tightening the bounds of Corollary \ref{coro:MaxCoordCoro}}
\label{sec:coroBoundTighten}

Still in the setting where the selection function chooses one coordinate per pair, assume $|P_i| \approx |P_j|$ for all $i,j \in [d]$, where $P_i$ is defined in Corollary \ref{coro:MaxCoordCoro}. Then, as we have stated in the main text, $\lambda, \eta, \zeta = O(1/d)$, and so by Corollary \ref{coro:MaxCoordCoro}, $\Omega(d^3\log(d / \delta))$ samples ensures the estimation error is $O(1)$. However, by tightening a bound used in the proof of Theorem \ref{thm:sampleComplexity}, we can show $\Omega(d^2\log(d / \delta))$ samples ensures the estimation error is $O(1)$.

Recall the definition of $\beta$ in Equation \eqref{eq:betadef}: $\beta:= \max_{(i,j) \in P} \| U_{i}^{\tau(i,j)} - U_{j}^{\tau(i,j)} \|_{\infty}$. In our proof, we use this to bound differences between feature vectors at Equation \eqref{eq:betabound}. In particular, for $k \in [d]$ we bound $\frac{1}{ \binom{n}{2}} \sum_{(i,j) \in P}  \left( (U_{i}^{\tau(i,j)})^{(k)} - (U_{j}^{\tau(i,j)})^{(k)} \right)^2\leq \beta^2$. For any $k \in [d]$, since $|P_i| \approx |P_j|$ for all $i,j \in [d]$, each coordinate is chosen approximately $\binom{n}{2} / d$ times. Therefore, $\frac{1}{ \binom{n}{2}} \sum_{(i,j) \in P}  \left( (U_{i}^{\tau(i,j)})^{(k)} - (U_{j}^{\tau(i,j)})^{(k)} \right)^2\leq \beta^2 \ d$ since only $\binom{n}{2} / d$ of the $\binom{n}{2}$ terms in the sum are non-zero. We can now replace $\beta$ with $\beta / \sqrt{d}$ in Corollary \ref{coro:MaxCoordCoro}. Therefore, $\Omega(d^2\log(d / \delta))$ samples ensures the estimation error is $O(1)$ since $\lambda, \eta, \zeta = O(1/d)$.

\section{Proof of Corollary \ref{coro:rankingComplexity}}
\label{appendix:coroRankingProof}
In this section, we present the full lower bounds on the number of samples and upper bound on the estimation error. The definitions of the constants that appear in the main text, i.e. $C_5$, appear at the end of the proof. 

\begin{corollary}[restatement of Corollary \ref{coro:rankingComplexity}: sample complexity of learning the ranking]
Assume the set-up of Theorem \ref{thm:sampleComplexity}. Pick $k \in [\binom{n}{2}]$. Let $\alpha_k$ be the $k$-th smallest number in $\{ |\langle w^*, U_i - U_j \rangle|: (i,j) \in P \}$. Let $M := \max_{i \in [n]} \|U_i\|_2$. Let $\gamma^*: [n] \rightarrow [n]$ be the ranking obtained from $w^*$ by sorting the items by their full-feature utilities $\langle w^*, U_i\rangle$ where $\gamma^*(i)$ is the position of item $i$ in the ranking. Define $\hat{\gamma}$ similarly but for the estimated ranking obtained from the MLE estimate $\hat{w}$. Let $\delta > 0$. Let \[m_1 = \frac{3\beta^2\log{(2d / \delta)}d + 4\sqrt{d}\beta \log{(2d^2 / \delta)}}{6},\] \[m_2 =  \frac{8\log(4 d / \delta)(6\eta + \lambda \zeta)}{3\lambda^2},\] and \[m_3 = \frac{64 M^2 (1 + \exp(b^*))^4 (3\beta^2\log{(4d / \delta)}d + 4\sqrt{d}\beta \log{(4d / \delta)})}{6\alpha_k^2 \exp(b^*)^2\lambda^2}.\] If $m \geq \{m_1, m_2, m_3\}$, then with probability $1 - \frac{2}{d}$, $K(\gamma^*,\hat{\gamma}) \leq k-1$, where $K(\gamma^*, \hat{\gamma}) = |\{(i,j) \in P: (\gamma^*(i) - \gamma^*(j))(\hat{\gamma}(i) - \hat{\gamma}(j)) <0\}|$ is the Kendall tau distance between two rankings.

\end{corollary}
\begin{proof}
By Theorem \ref{thm:sampleComplexity}, with probability $1 - \delta$, we have 
\begin{align}
    \|w^* - \hat{w}\|_2 & \leq \frac{4(1 + \exp(b^*))^2}{\exp(b^*)\lambda} \sqrt{\frac{3\beta^2\log{(4d  / \delta)}d + 4\sqrt{d}\beta \log{(4d  / \delta)}}{6m}} \\
    & \leq \frac{\alpha_k}{2M}
\end{align}

by definition of $m$.


The estimated full feature utility for item $i$ is no further than $\frac{\alpha_k}{2}$ to the true utility of item $i$:
\begin{align}
    |\langle w^* - \hat{w}, U_i \rangle| &\leq \|w^* - \hat{w} \|_2 \|U_i\|_2 \text{ \ by Cauchy–Schwarz}  \\ 
    & \leq \frac{\alpha_k \|U_i\|_2}{2M} \\
    & \leq \frac{\alpha_k}{2}. 
\end{align}

Therefore for any $i \in [n]$, \begin{equation}
\label{eqn:closeUtility}
    \langle w^*, U_i \rangle - \frac{\alpha_k}{2} \leq \langle \hat{w}, U_i \rangle \leq \langle w^*, U_i \rangle + \frac{\alpha_k}{2}.
\end{equation}

Let $P_{\alpha_k}:= \{(i,j) \in P: | \langle w^*, U_i-U_j \rangle| \geq \alpha_k \}$ and let $(i,j) \in P_{\alpha_k}$. WLOG, suppose $\langle w^*, U_i \rangle - \langle w^*, U_j \rangle \leq 0$, i.e. $\gamma^*(i) - \gamma^*(j) \leq 0$, which means item $j$ is ranked higher than item $i$ in the true ranking given by $\gamma$. We want to show $ \langle \hat{w}, U_i \rangle - \langle \hat{w}, U_j \rangle  \leq 0$, i.e. $\hat{\gamma}(i) - \hat{\gamma}(j) \leq 0$, meaning that item $j$ is ranked higher than item $i$ in the estimated ranking given by $\hat{\gamma}.$

By applying Equation \eqref{eqn:closeUtility} and using the fact $\langle w^*, U_i \rangle - \langle w^*, U_j \rangle \leq 0$, we have 
\begin{align}
     \langle \hat{w}, U_i \rangle & \leq \langle w^*, U_i \rangle  + \frac{\alpha_k}{2} \text{ \ by Equation \eqref{eqn:closeUtility}} \\
      &= \langle w^*, U_i \rangle - \langle w^*, U_j \rangle + \langle w^*, U_j \rangle  + \frac{\alpha_k}{2}  \\   
      &\leq -\alpha_k + \langle w^*, U_j \rangle  + \frac{\alpha_k}{2}  \text{ since } (i,j) \in P_{\alpha_k} \text{ and since } \langle w^*, U_i \rangle - \langle w^*, U_j \rangle \leq 0\\  
      &\leq \langle w^*, U_j \rangle  - \frac{\alpha_k}{2}  \\  
      & \leq \langle \hat{w}, U_j \rangle \text{ \ by Equation \eqref{eqn:closeUtility}}.
\end{align}
Hence, $\langle \hat{w}, U_i \rangle - \langle \hat{w}, U_j \rangle \leq 0$ for every $i,j \in P_k$, meaning that for any $(i,j) \in P_k$, $\gamma^*$ and $\hat{\gamma}$ agree on the relative ordering of item $i$ and $j$. Furthermore, $|P_k| = \binom{n}{2} - (k-1)$. Therefore, $K(\gamma^*, \hat{\gamma}) \leq \binom{n}{2}  - |P_k| = k-1.$ 

Now we explain how to get from these results to those in the main paper with the order terms. The value of $C_1$ and $C_2$ are given at the end of the proof of Theorem \ref{thm:sampleComplexity}. It is easy to see that $C_5 =  64 *4*2^4 / 6$. 

\end{proof}

\section{Synthetic Experiments}
Code is available at \url{https://github.com/Amandarg/salient_features}.

\subsection{Plot of Parameters in Theorem \ref{thm:sampleComplexity}}
\label{appendix:parameterPlot}
In this section, the goal is to empirically illustrate how the top-$t$ selection function and intransitivities effect the parameters $b^*$, $\zeta$, $\eta, \beta,$ and $\lambda$ from Theorem \ref{thm:sampleComplexity} and hence the number of samples required and the exact upper bound on the estimation error. Just as in the synthetic experiment section, we sample each coordinate of $U$ from $N(0, \frac{1}{\sqrt{d}})$ and each coordinate of $w^*$ is sampled from $N(0, \frac{4}{\sqrt{d}})$.

In the experiments, the ambient dimension $d = 10$ and the number of items $n = 100$. We repeat the following 10 times: sample $U$ and $w^*$, and use this $U$ and $w^*$ while varying $t \in [d]$ to compute all of the parameters of interest and intransitivity rates. The $x$-axis of each plot is the average strong stochastic transitivity (SST) violation rate defined in Section \ref{sec:syntheticExperiments} where the average is taken over the 10 experiments. From Figure \ref{fig:violationRationalChoice}, intransitives decrease as $t$ increases, so the $x$-axis in Figures \ref{fig:param,top_k} and \ref{fig:samp,topk} could roughly, but not exactly, be replaced with $t$, where $t$ is decreasing from 10 to 1. The $y$-axis on the plots depict the average value and the bars represent the standard error over the 10 experiments.

Figure \ref{fig:param,top_k} shows the parameters in Theorem \ref{thm:sampleComplexity}.
Larger $\lambda$ means smaller sample complexity, whereas smaller $b^*, \zeta, \beta$ and $\eta$ means smaller sample complexity.

Recall in the Supplement re-statement of Theorem \ref{thm:sampleComplexity}, the number of samples $m$ required in the theorem is \[m \geq \max\left\{\frac{3\beta^2\log{(4 d / \delta)}d + 4\sqrt{d}\beta \log{(4 d / \delta)}}{6}, \frac{8\log(2d / \delta)(6\eta + \lambda \zeta)}{3\lambda^2} \right\}.\] Let $m_1 = \frac{3\beta^2\log{(4 d / \delta)}d + 4\sqrt{d}\beta \log{(4 d / \delta)}}{6}$ and $m_2 = \frac{8\log(2 d / \delta)(6\eta + \lambda \zeta)}{3\lambda^2}.$  Figure \ref{fig:samp,topk} shows $m_1$, $m_2$, and the bound from Theorem \ref{thm:sampleComplexity} with $\delta =  \frac{1}{\delta} = \frac{1}{10}$ without the number of samples, i.e. the upper bound plot on the left does not include the number of samples in it. The plot shows \[ \frac{4(1 + \exp(b^*))^2}{\exp(b^*)\lambda}  \sqrt{\frac{3\beta^2\log{(4d / \delta)}d + 4\sqrt{d}\beta \log{(4d / \delta)}}{6}}\] without the $\frac{1}{\sqrt{m}}$ term. Note that $m_1$ has constant average and standard error bars since with the dimension fixed, it is a function of $\beta$, which is constant in this case. Furthermore, this plot suggests that $m_1 << m_2$.

\begin{figure}
\centering

   \includegraphics[width=1\linewidth]{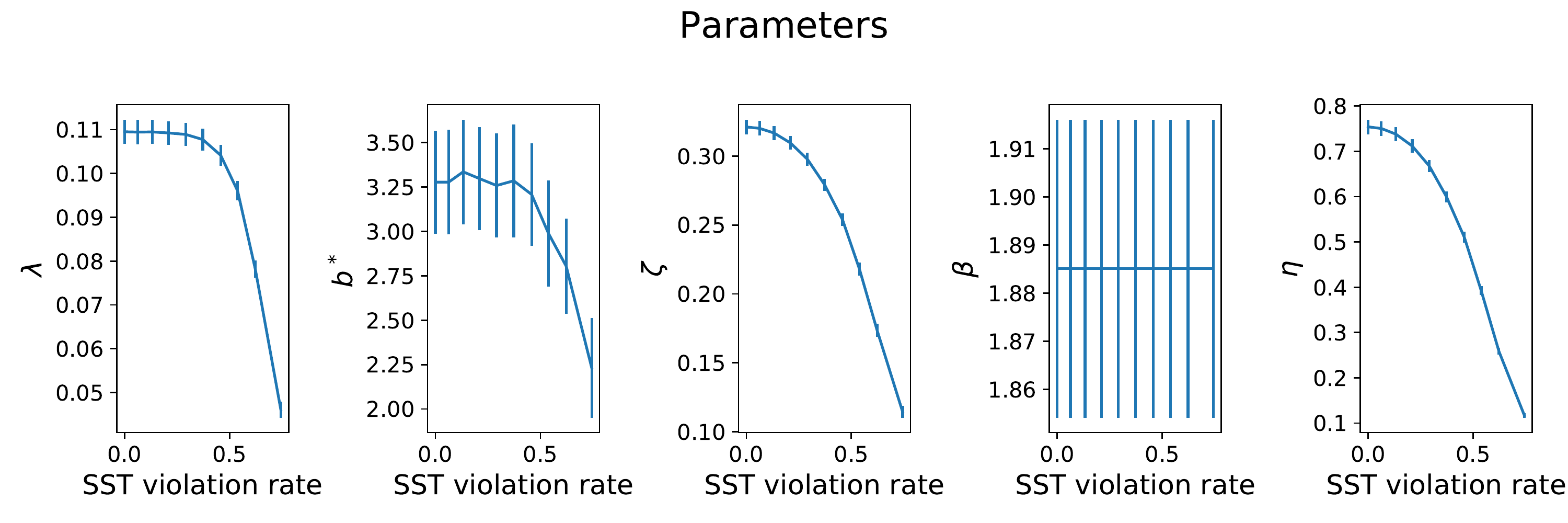}
\caption{The parameters of Theorem \ref{thm:sampleComplexity} for the top-$t$ selection function as a function of the average strong stochastic transitivity violation rate over the 10 experiments. The average over 10 experiments where a new $U$ and $w^*$ are drawn each time is depicted. The bars represent the standard error over the 10 experiments.}
\label{fig:param,top_k}
\end{figure}

\begin{figure}
\centering

   \includegraphics[width=1\linewidth]{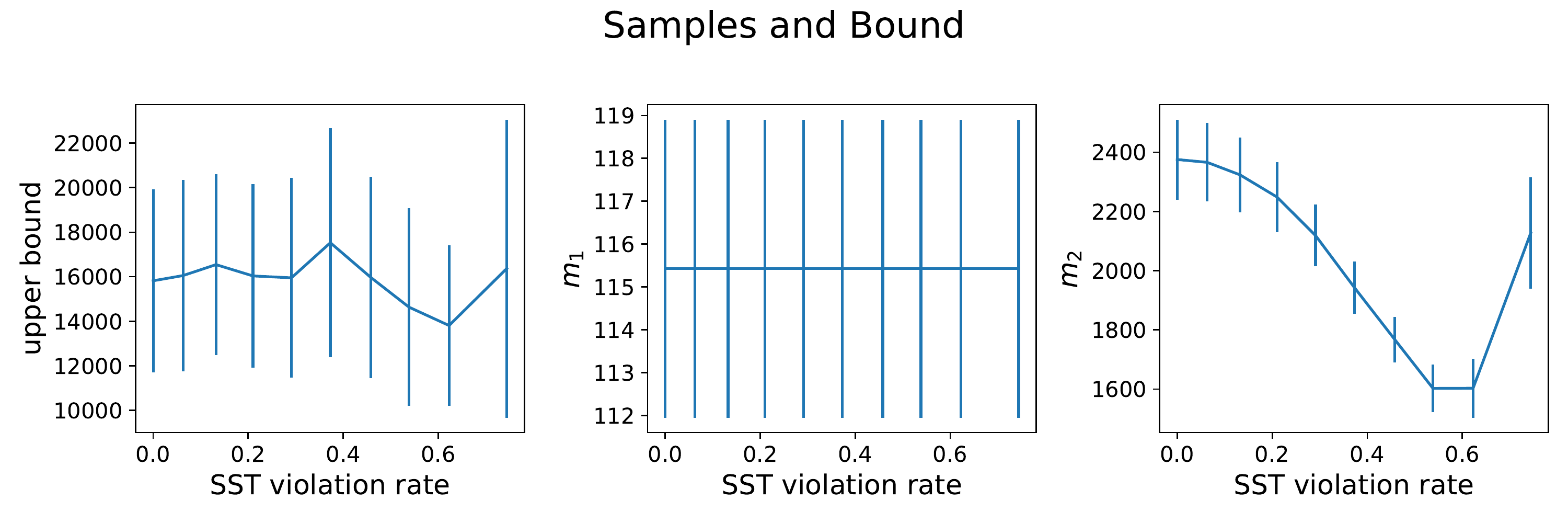}

\caption{Number of samples $m_1$ and $m_2$ and upper bound on estimation error from Theorem \ref{thm:sampleComplexity} for the top-$t$ selection function as a function of the average strong stochastic transitivity violation rate over the 10 experiments. The average over 10 experiments where a new $U$ and $w^*$ are drawn is depicted. The bars represent the standard error over the 10 experiments.}
\label{fig:samp,topk}
\end{figure}

\subsection{Additional Synthetic Experiments and Details}
\label{sec:additionalExperiments} 
First we define the Kendall tau correlation. It is used in both Sections \ref{sec:syntheticExperiments} and \ref{sec:realDataExp}, and is defined as follows.
Let $\gamma, \rho: [n] \rightarrow [n]$ be two rankings on $n$ items where $\gamma(i)$ and $\rho(i)$ is the position of item $i$ in the ranking. Let $A = \sum_{(i,j) \in P} \mathbbm{1}_{\{(\sigma(i) - \sigma(j))(\rho(i) - \rho(j))>0\}}$, respectively $D = \sum_{(i,j) \in P} \mathbbm{1}_{\{(\sigma(i) - \sigma(j))(\rho(i) - \rho(j))\leq 0\}} $,  be the number of pairs of items that $\sigma$ and $\rho$ agree, respectively disagree, on the relative ordering. Then the Kendall tau correlation of $\rho$ and $\gamma$ is 
\begin{equation}
    KT(\gamma, \rho):=\frac{A-D}{\binom{n}{2}}\label{eq:kt}.
\end{equation}

\begin{figure}
    \centering
    \includegraphics[width=8.5cm]{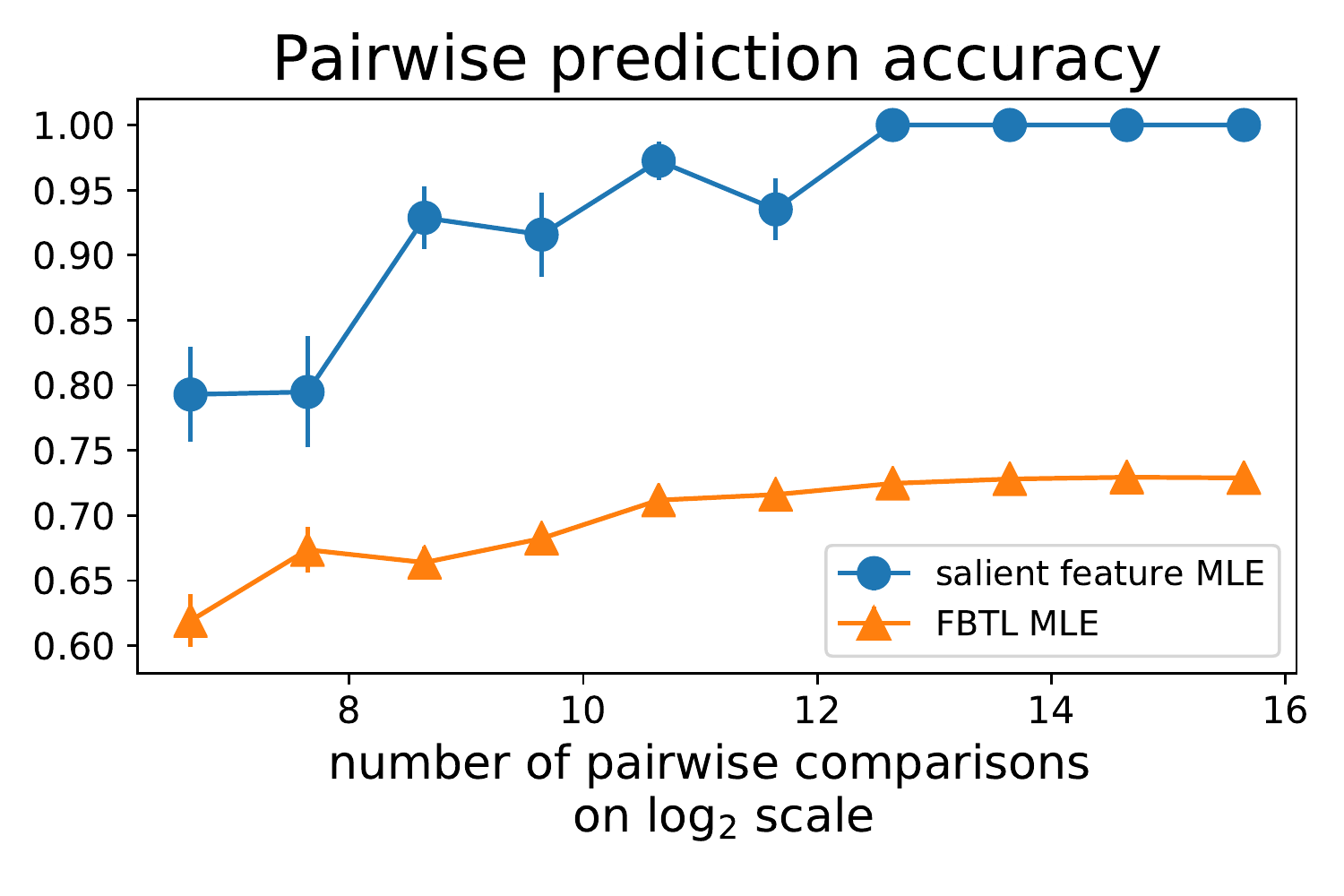}
    \caption{Pairwise prediction accuracy as a function of the number of samples, which are on the logarithmic scale, where the pairwise comparisons are sampled from the salient feature preference model with the top-$1$.}
    \label{fig:predictionAcc}%
\end{figure}

Second, recall the set-up in Section \ref{sec:experiments}: The ambient dimension $d=10$, the number of items $n = 100$, and the top-$1$ selection function is used. The coordinates of $U$ are drawn from $\mathcal{N}\left(0, \frac{1}{\sqrt{d}}\right)$,and the coordinates of $w^*$ are drawn from $\mathcal{N}\left(0, \frac{4}{\sqrt{d}}\right)$. We sample $m$ pairwise comparisons for $m \in \{2^i*(100): i \in [11]\}$, fit the MLEs of the FBTL and salient preference model with the top-$1$ selection function, and repeat 10 times. Figure \ref{fig:predictionAcc} shows the average pairwise prediction accuracy, which is defined as $$\frac{|\{(i,j) \in P: (P_{ij} - .5)(\hat{P}_{ij} -5)>0\}|}{\binom{n}{2}}$$ where $\hat{P}_{ij}$ is the estimated pairwise probability that item $i$ beats item $j$. The bars shows the standard error over the 10 experiments. The gap between the salient feature preference model MLE and the FBTL MLE is expected since the data is generated from the salient feature preference model.

\begin{figure}[htp]

\centering
\includegraphics[width=.5\textwidth]{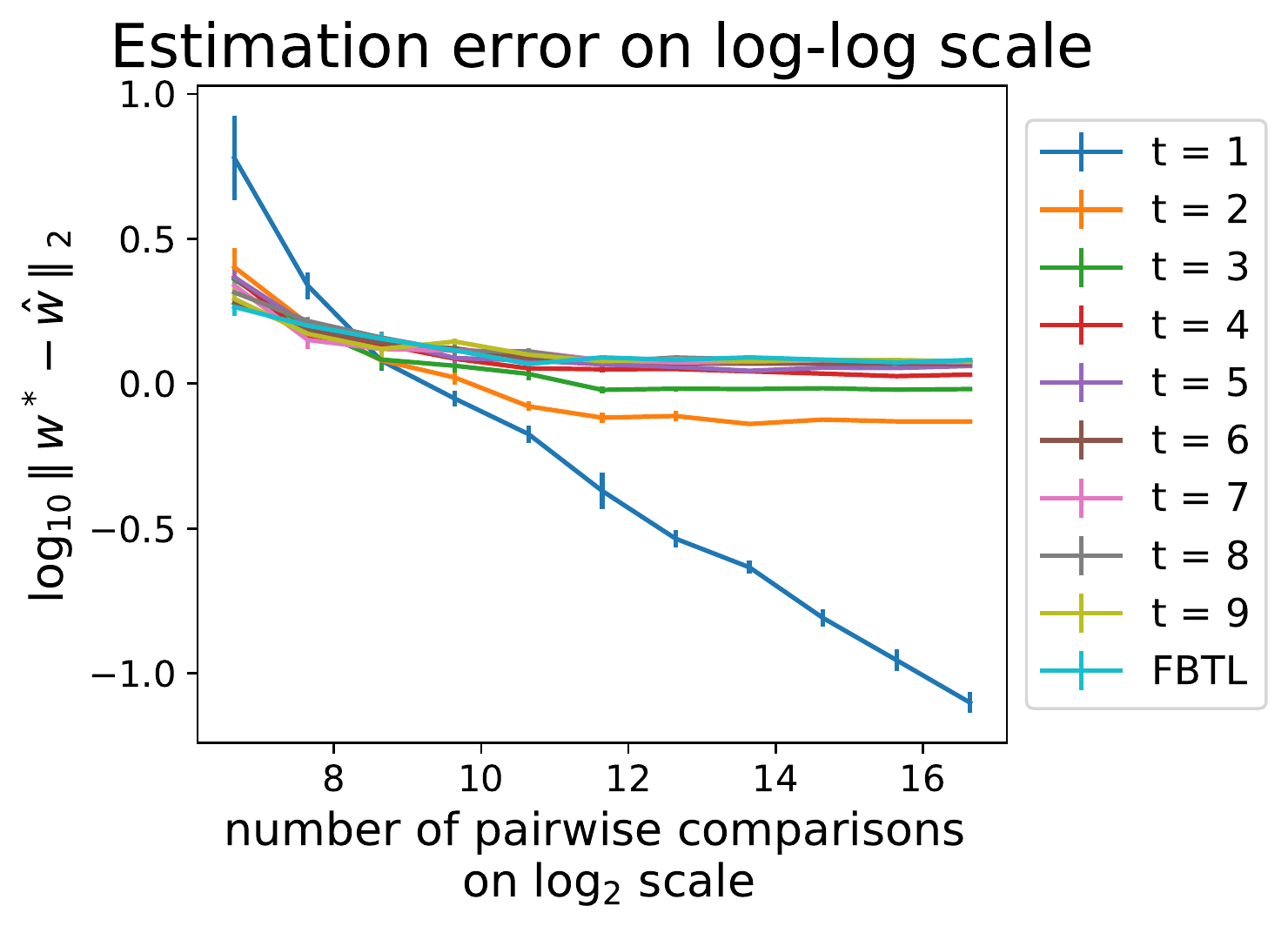}
\includegraphics[width=.45\textwidth]{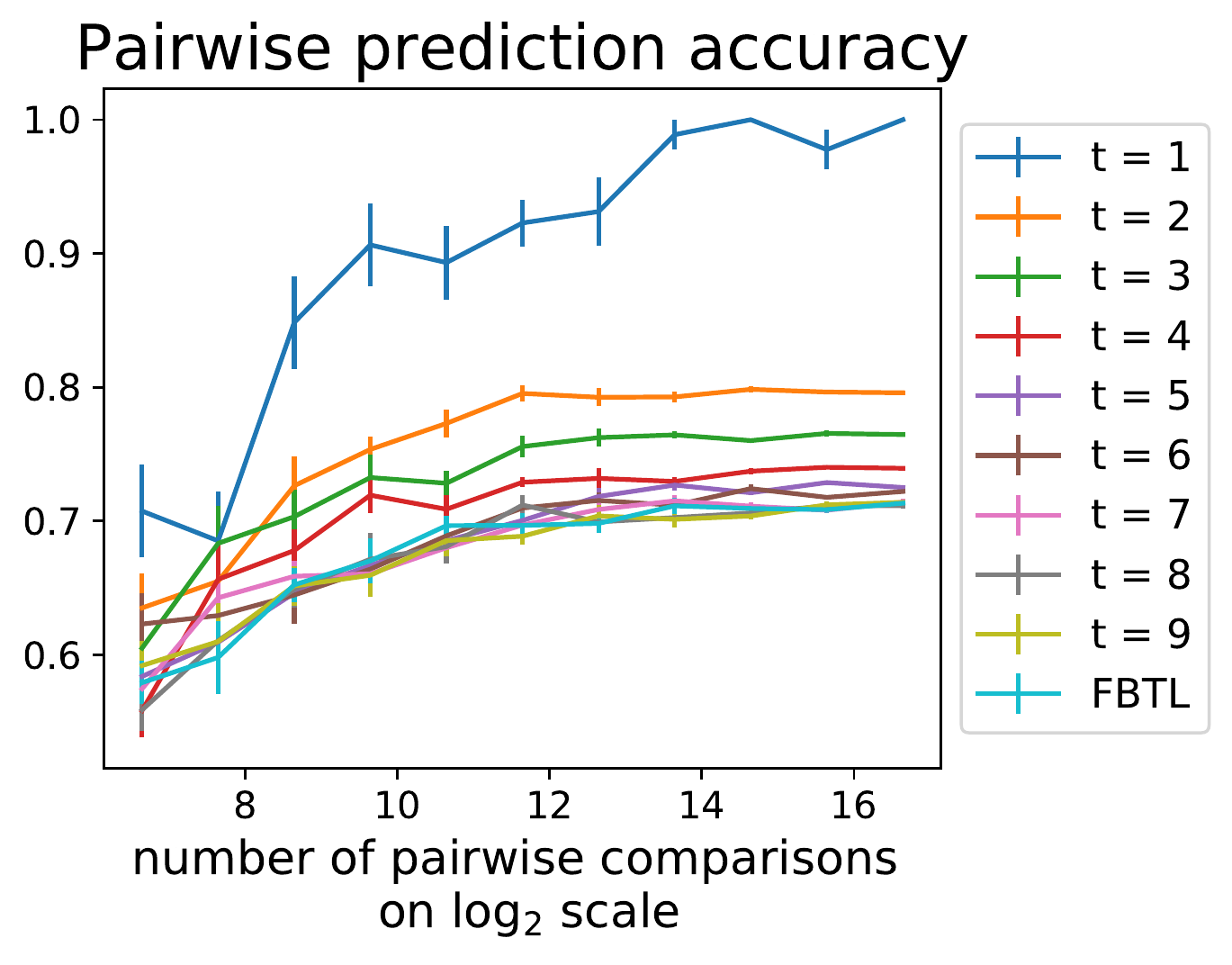}
\\ 
\includegraphics[width=.45\textwidth]{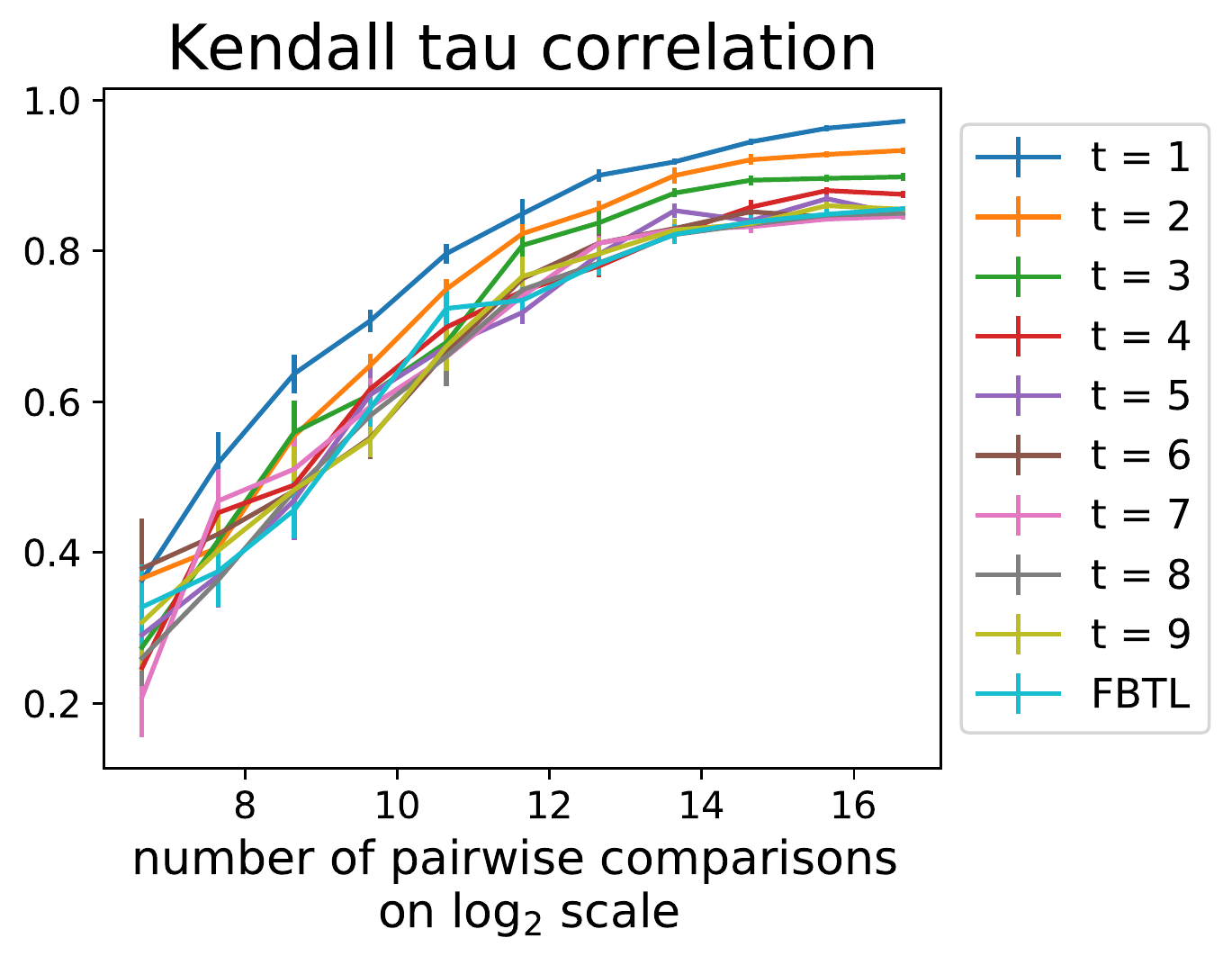}

\caption{These plots investigate model misspecification. The true generative model for the pairwise preference data is the salient feature preference model with the top-$1$ selection function. The coordinates of $U$ and $w$ are sampled from a Gaussian as described in the main text. The MLEs for the salient feature preference model with the top-$t$ selection function for $t \in [d]$ is shown. }
\label{fig:modelMisspecfication1}

\end{figure}

\begin{figure}[htp]

\centering
\includegraphics[width=.5\textwidth]{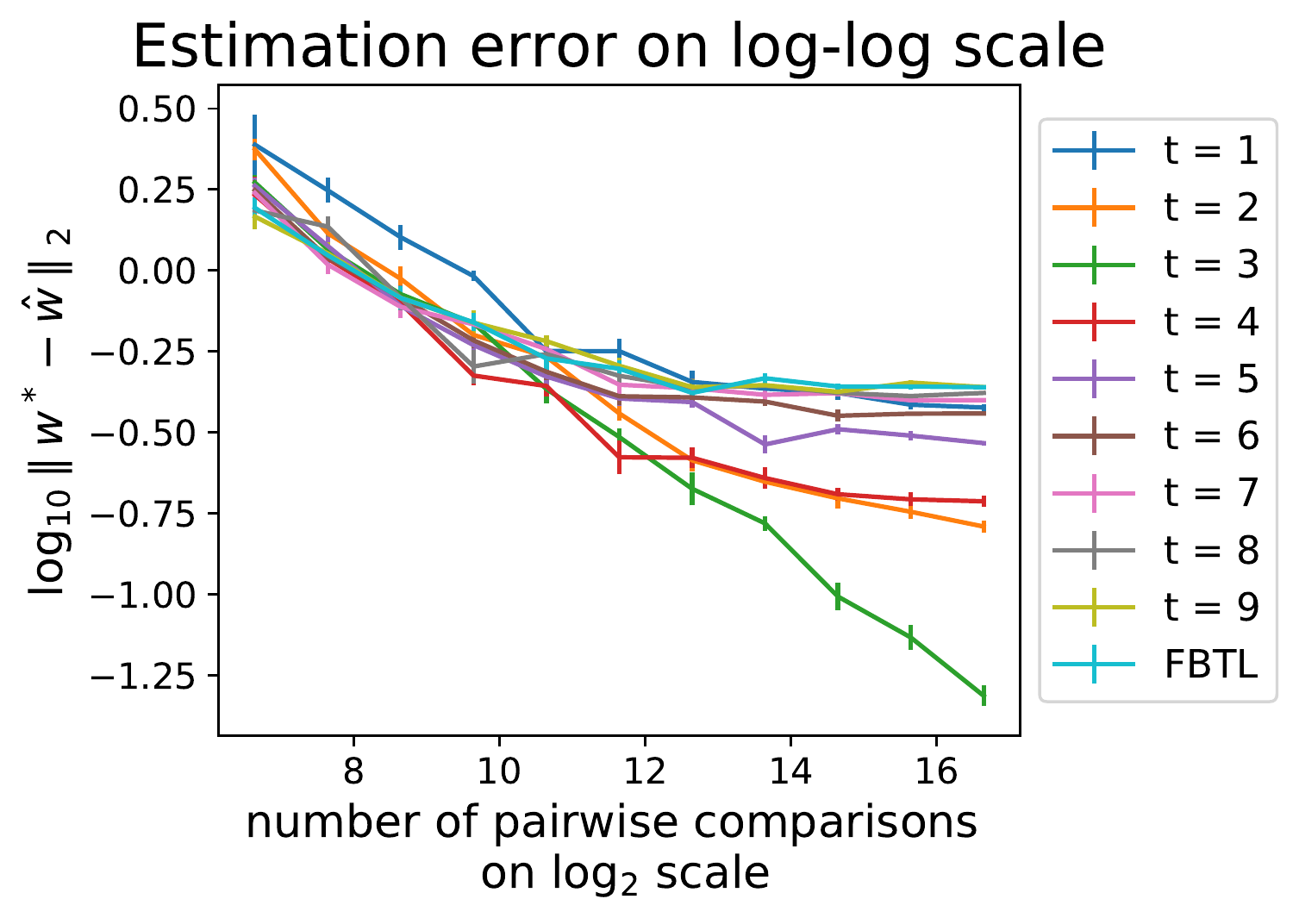}
\includegraphics[width=.45\textwidth]{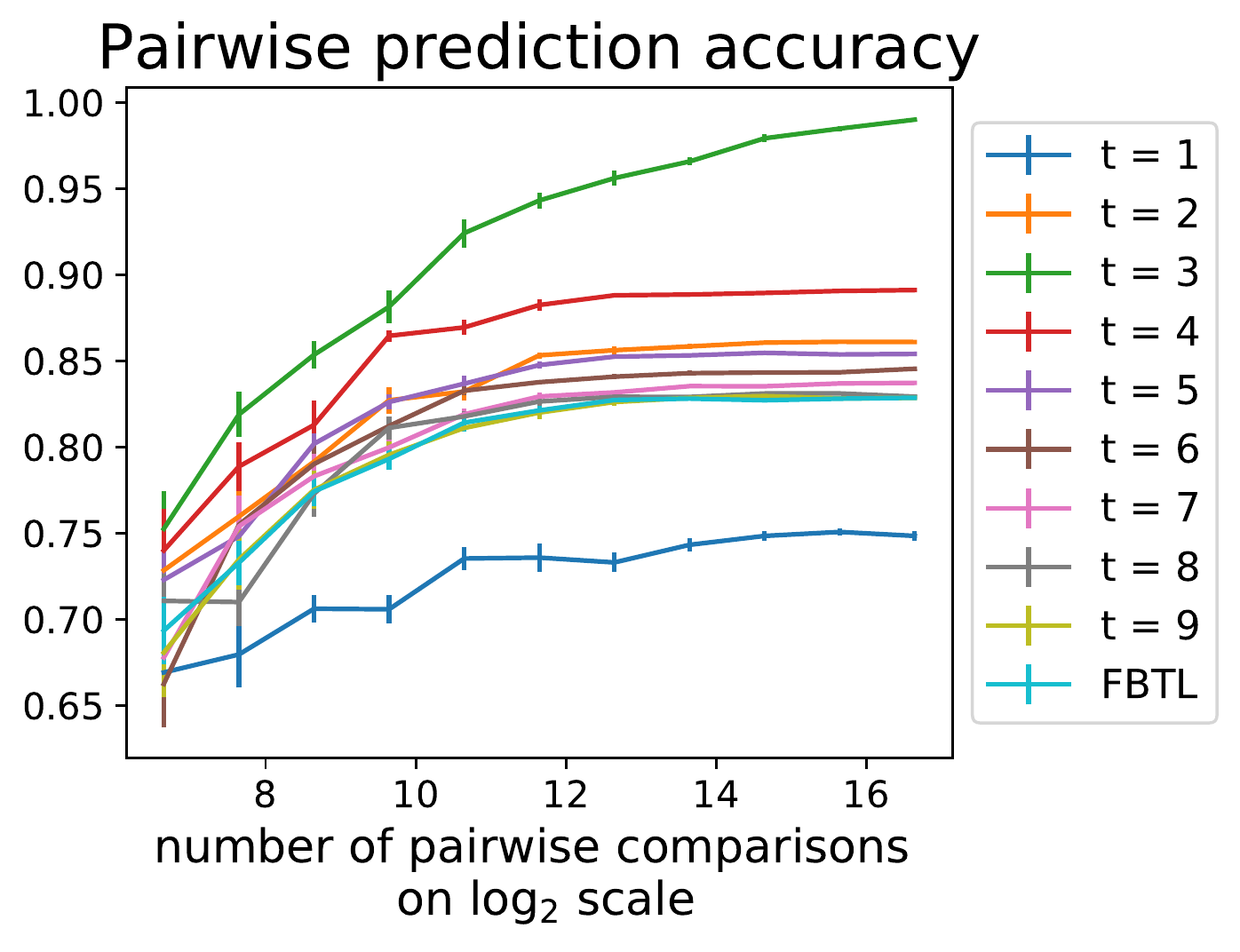}
\includegraphics[width=.45\textwidth]{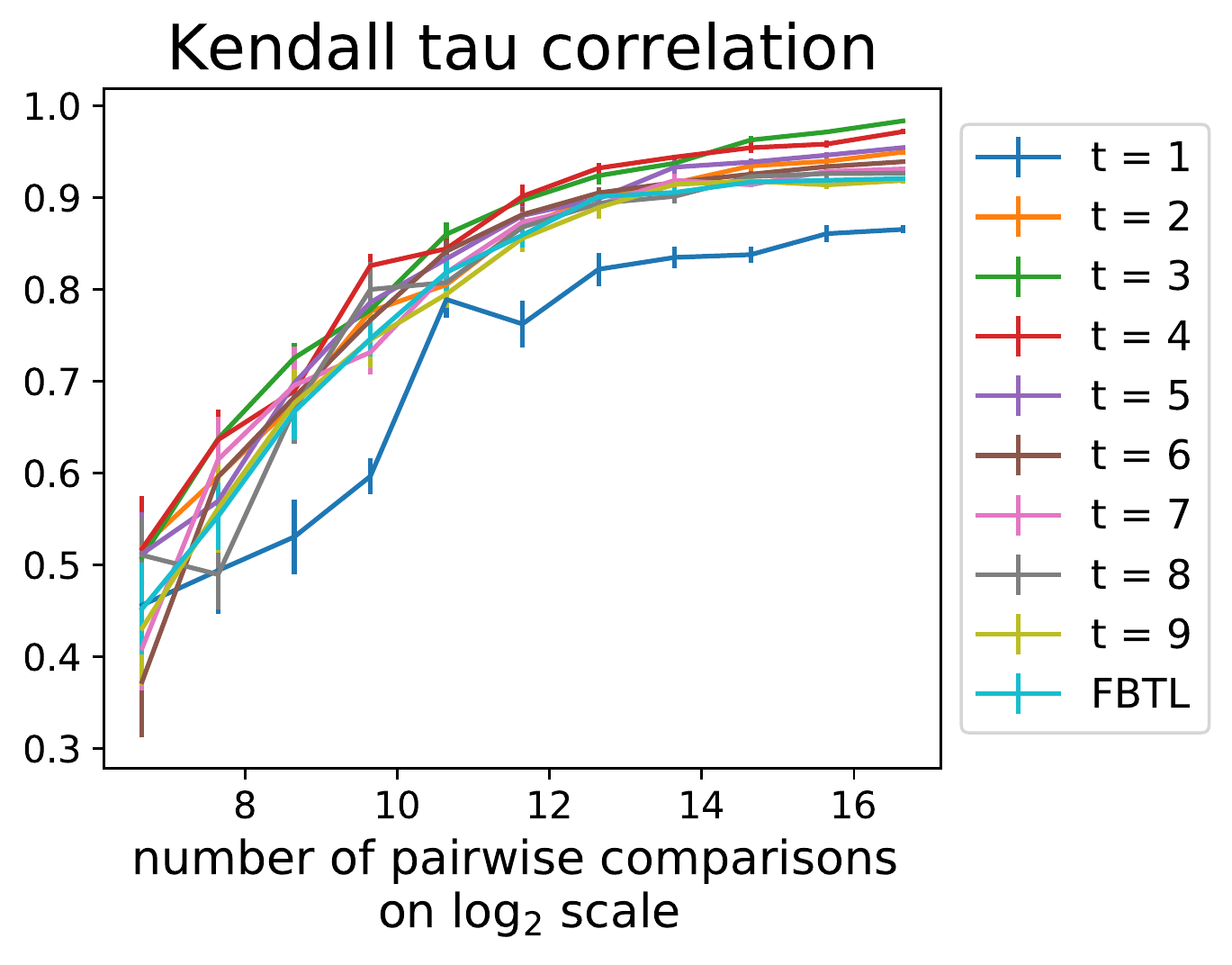}

\caption{These plots investigate model misspecification. The true generative model for the pairwise preference data is the salient feature preference model with the top-$3$ selection function. The coordinates of $U$ and $w$ are sampled from a Gaussian as described in the main text. The MLEs for the salient feature preference model with the top-$t$ selection function for $t \in [d]$ is shown. }
\label{fig:modelMisspecfication3}

\end{figure}

Third, see Figures \ref{fig:modelMisspecfication1} and \ref{fig:modelMisspecfication3} for plots investigating model misspecification. In particular, we use the same experimental set-up as in Section \ref{sec:syntheticExperiments} except that in Figure \ref{fig:modelMisspecfication3} the salient feature preference model with the top-$3$ selection function is used to generate the preference data. We fit the MLE for the salient feature preference model for the top-$t$ selection function for all $t \in [d]$ for both plots. The FBTL model is equivalent to when $t = 10$.

In Figure \ref{fig:modelMisspecfication1}, we see that the model is very sensitive to the choice of $t$. As we would expect, $t=2$ has the second smallest error when the number of samples exceed $2^{10}$.

In Figure \ref{fig:modelMisspecfication3}, we see that the model is still sensitive to the choice of $t$, but not as sensitive as in Figure \ref{fig:modelMisspecfication1}. In this case, we can not only overestimate $t$, i.e. $t>3$, but underestimate $t$, i.e. $t<3$. We see that $t = 2$ and $t = 4$--the two values of $t$ closest to the truth of $t=3$--have roughly the same error. Interestingly, $t=1$ has the worst performance.

\section{Real Data Experiments}
Code is available at \url{https://github.com/Amandarg/salient_features}.

\subsection{Algorithm implementation}
\label{appendix:algorithmImplementation}
In this section, we provide relevant details about how each algorithm is implemented. 
\begin{itemize}
    \item \textbf{RankNet:} We use the RankNet implementation found at \url{https://github.com/airalcorn2/RankNet}, which uses Keras. However, we use the Adam optimizer with default parameters except with a learning rate of 0.0001. We also add an $\ell_2$ penalty to the weights.
    
    \item \textbf{Salient feature preference model and FBTL:}  We use \texttt{sklearn}'s logistic regression solver. In particular, we set \texttt{tol} $= 1e-10$ and \texttt{max\_iter} $= 10000$. Furthermore, we do not fit an intercept. We use the default \texttt{liblinear} solver for real data experiments, and the \texttt{sag} solver for synthetic data experiments since we do not use regularization. All other parameters use the default values.
    
    \item \textbf{Ranking SVM:}  We use \texttt{sklearn}'s \texttt{LinearSVC} solver with the same parameters as above. In particular, we do not fit an intercept.
\end{itemize}

The synthetic experiments were ran on a 2016 MacBook Pro with a 2.6 GhZ Quad-Core Intel Core i7 processor. The real data experiments were ran on the University of Michigan's Great Lakes Cluster \footnote{\url{https://arc-ts.umich.edu/greatlakes/}}.

\subsection{District compactness experiments}
\label{appendix:district_experiments}
We refer the reader to \cite{districtCompactness} for the full details about the district compactness data, but provide relevant details here. We obtained the data by contacting the authors.

\subsection{Pairwise comparison description}
There were three pairwise comparison studies. Due to data collection issues, only two of these pairwise comparison studies, called \texttt{shiny2pairs} and \texttt{shiny3pairs}, are available. In \texttt{shiny2pairs}, there are 3,576 pairwise for 298 people who each answered 12 pairwise comparisons. In \texttt{shiny3pairs}, there are 1,800 pairwise comparisons for 90 people who each answered 20 pairwise comparisons. There is no overlap in the districts used in \texttt{shiny2pairs} and \texttt{shiny3pairs}. 


\subsection{$k$-wise rankings for $k>2$ description}

There are 8 sets of $k$-wise ranking data. In many cases, the feature data for some districts are missing entirely, so in our own experiments, we throw out any district without feature data. Recall, we use the $k$-wise ranking data for validation and testing, so we also remove any districts present in the training set.

\begin{itemize}
    \item \texttt{Shiny1} contains rankings for 298 people on 20 districts, but the feature information for 10 districts are missing. The people are composed of undergraduate students, PhD students, law students, consultants, legislators involved in the redistricting process, and judges.
    \item \texttt{Shiny2} contains rankings on 20 districts for 103 people collected on Mturk. The feature information on 10 of the districts are missing however.
    \item \texttt{Mturk} contains another set of Mturk experiments collected on 100 districts and 13 people, which we use as our validation set. However, 34 of the districts also had pairwise comparison information collected about them, so we throw these out. 
    \item \texttt{UG1-j1}, \texttt{UG1-j2}, \texttt{UG1-j3}, \texttt{UG1-j4}, and \texttt{UG1-j5} are 4 sets of $20$-wise ranking data for 4 undergraduates at Harvard. The initial task was to rank 100 districts at once, but the resulting data set contains 5 sets of rankings on 20 districts. Out of the 100 districts used across the 5 sets of rankings, there are 38 districts with missing feature information.
\end{itemize}

\begin{figure}%
    \centering
    \includegraphics[width=9cm]{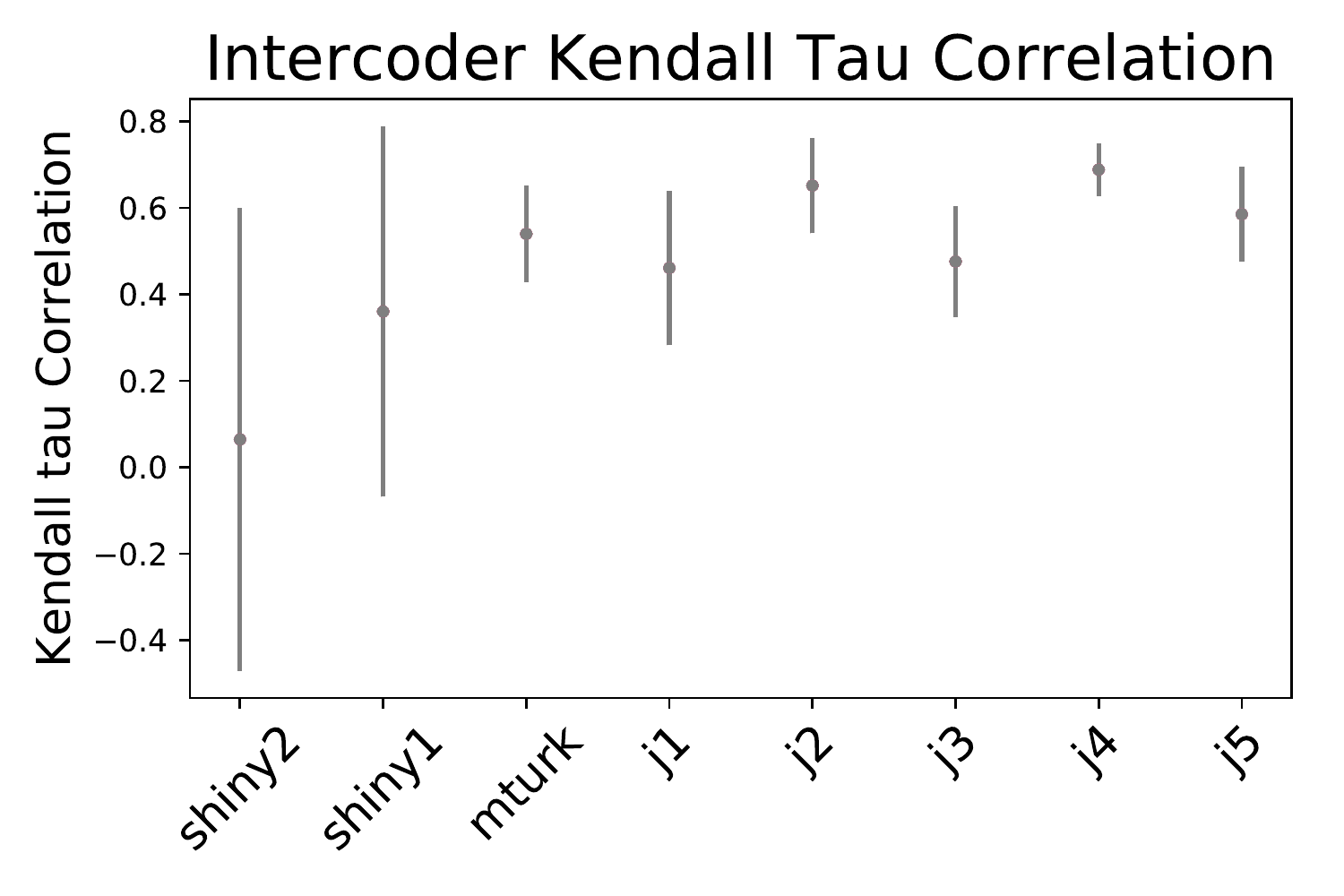} 
  \caption{For each of the $k$-wise ranking data sets, the average agreement between people in terms of the Kendall tau correlation is shown.}%
    \label{fig:intercoder}%
\end{figure}

See Figure \ref{fig:intercoder} which depicts the average Kendall tau correlation between pairs of rankings in a $k$-wise ranking data set and the standard deviation. Recall the Kendall tau correlation, $KT(\cdot, \cdot)$, is defined in Equation \eqref{eq:kt}. This plot shows roughly how much people agree with each other, where higher values mean more agreement. In particular, suppose there are $N$ $k$-wise rankings given by $\sigma_1, \dots, \sigma_N$. Then the average Kendall tau correlation for the $N$ rankings is

\[\frac{1}{2 \binom{N}{2}} \sum_{(i,j) \in [N]\times[N]} \mathrm{KT}(\sigma_i, \sigma_j)\] and refer to this quantity as the average intercoder Kendall tau correlation. We see that people typically disagree on \texttt{shiny2} and \texttt{shiny1}, whereas people tend to agree more often on the rest of the $k$-wise data sets perhaps because there are fewer people. 

The districts used in \texttt{shiny1} and \texttt{shiny2} are the same, and these districts also comprise one of the \texttt{UG1} data sets as well. However, the districts in \texttt{mturk} are disjoint from the rest of the $k$-wise ranking sets. In addition, \texttt{mturk} has relatively low intercoder variability. For these two reasons, we decided to use mturk as our validation set. We decided to keep \texttt{shiny1} and \texttt{shiny2} separate since the original authors did and also since they are comprised of different groups of people resulting in different behavior, e.g., \texttt{shiny1} has a higher average intercoder Kendall tau correlation than \texttt{shiny2}.

\subsection{Data preprocessing}
We remove pairwise comparisons that were asked fewer than 5 times resulting in 5,150 pairwise comparisons over 94 unique pairs on 122 districts. There are 8 sets of $k$-wise comparison data that we use for validation and testing. We remove any districts in the $k$-wise ranking data that are present in the training data. We standardize the features of the districts by subtracting the mean and dividing by the standard deviation, where we use the mean and standard deviation from the training set. Standardizing the features is important for the salient feature preference model with the top-$t$ selection function, so that each feature is roughly on the same scale. Otherwise, the top-$t$ selection function might just choose the coordinates with the largest magnitude, and not the coordinates truly with the most variability. 

\subsection{Experiment details}
\label{appendix:districtHyperparameters}

The hyperparameters for the salient feature preference model with the top-$t$ selection function are $t$ and the $\ell_2$ regularization parameter $\mu$. The hyperparameter for FBTL is the $\ell_2$ regularization parameter $\mu$. For Ranking SVM, the only hyperparameter is $C$ which controls the penalty for violating the margin. We vary $t \in [d]$ where $d=27$ since there are 27 features. We vary $\mu$ and $C$ in $\{.00001, .0001, .001, .01,.1,1,10,100, 1000, 10000, 100000, 1000000\}$.

The hyperparameters for RankNet include the $\ell_2$ regularization parameter $\mu$ and number of nodes in the hidden layer. We use one hidden layer. We varied the number of nodes in the single hidden unit in in $\{5*i: i \in [19]\}$. We use a batch size of 250, and we use 800 epochs. Initially, we varied $\mu$ also in $\{.00001, .0001, .001, .01,.1,1,10,100, 1000, 10000, 100000, 1000000\}$, but as we will discuss in the next section we decided to vary $\mu$ in $\{.00001, .0001, .001, .01,.1,1,10\}$.

\subsection{Best performing hyperparameters}
Again, the validation set that was use is the \texttt{mturk} ranking data. Given $\hat{w}$, an estimate of $w^*$, we estimate the ranking by sorting each item's features with its inner product with $\hat{w}$. Then we pick the best hyperparameters by the largest average Kendall tau correlation of the estimated ranking with each individual ranking in \texttt{mturk}. 

For FBTL, the best performing hyperparameter is $\mu = 100000$. The average Kendall tau correlation of the estimated ranking to each individual ranking in \texttt{mturk} is 0.38 with a standard deviation of 0.05. The pairwise comparison accuracy on the training set is $56\%$, which is defined in Section \ref{sec:additionalExperiments} of the Supplement. Although the regularization strength is large, the norm of the estimated judgement vector is .015. The largest coordinate of the judgement vector in absolute value is .005 and the smallest is .0001.

For the salient feature preference model with the top-$t$ selection function the best performing hyperparameters are $t = 2$ and $\mu = .001$. The average Kendall tau correlation of the estimated ranking to each individual ranking in \texttt{mturk} is 0.54 with a standard deviation of 0.06. The pairwise comparison accuracy on the training set is $69\%$. 

\begin{figure}%
    \centering
    \includegraphics[width=12cm]{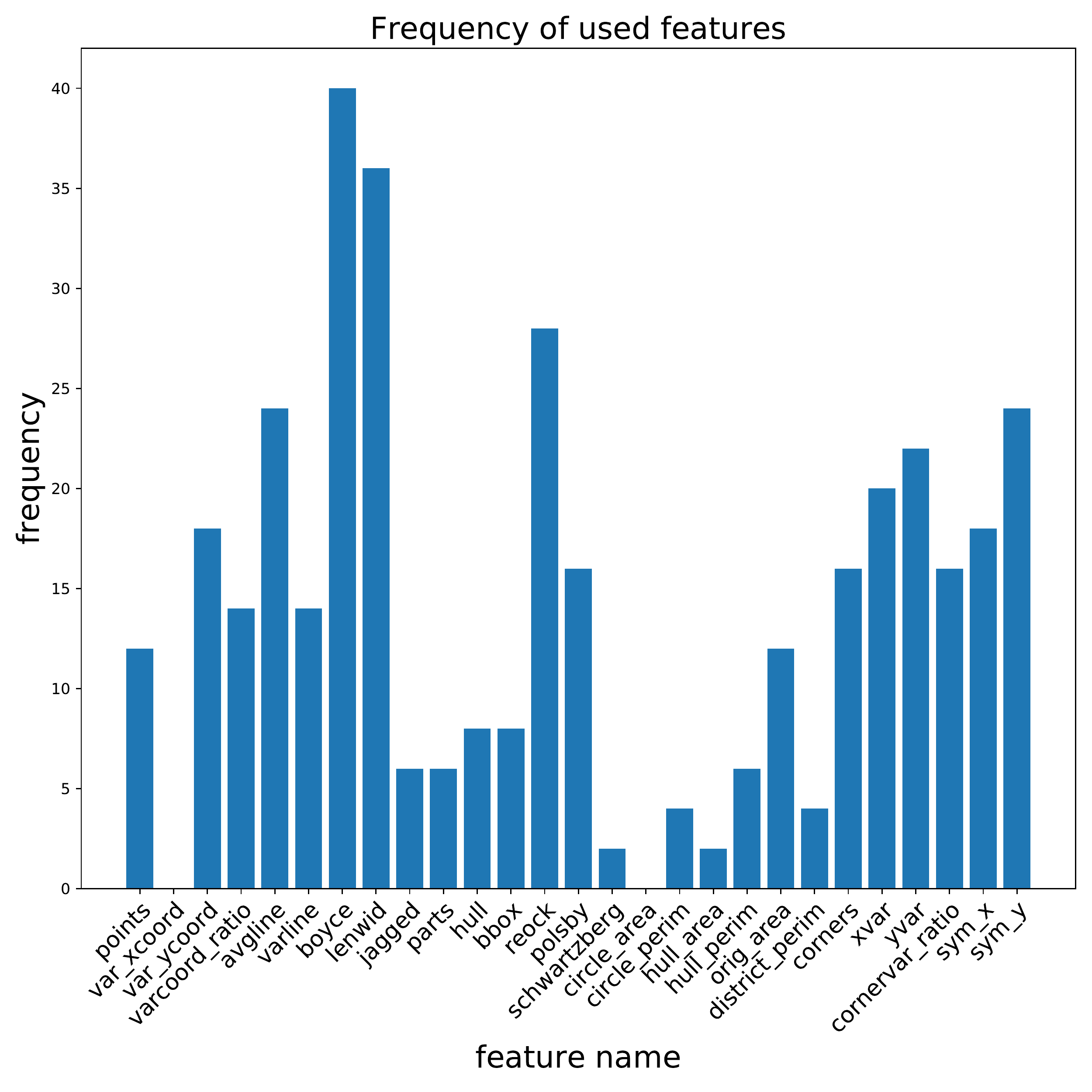} 
  \caption{The frequency that the top-$2$ selection function chooses each feature over unique pairwise comparisons in the training data.}%
    \label{fig:distributionOfFeatures}%
\end{figure}

Figure \ref{fig:distributionOfFeatures} shows how often each of the 27 features are selected by the top-$2$ selection function over unique pairwise comparisons in the training data. Notice that \texttt{var xcoord} and \texttt{circle area} are never selected. The learned weights for those features in the FBTL model when all the features are used are 2 of the top 3 features with the smallest weights, so these features play a relatively insignificant role when all the features are used any way.

For RankNet, the best hyperparameters on the validation set are $\mu = .1$ and 75 nodes in the hidden layer. The average Kendall tau correlation of the estimated ranking to each individual ranking in \texttt{mturk} is 0.407 with a standard deviation of 0.05. The pairwise comparison accuracy on the training set is $59\%$. As we discussed in the previous section, we initially searched over larger values of $\mu$. The best performing hyperparameters were $\mu = 10000$ and 40 nodes in the hidden layer. The pairwise comparison training accuracy was higher ($69\%$) and the average Kendall tau correlation on the validation set was also higher (.48 with a standard deviation of .05). However, these hyperparameters were very unstable, i.e. training on the same data with the same hyperparameters sometimes gave a completely different model where the average Kendall tau correlation on the validation set or some of the test sets were sometimes negative. 

For Ranking SVM, the best hyperparameter on the validation set is $C = 1000000$. The average Kendall tau correlation of the estimated ranking to each individual ranking in \texttt{mturk} is 0.38 with a standard deviation of 0.05. The pairwise comparison accuracy on the training set is $56\%$. Although $C$ is large, the norm of the estimate of the judgement vector is $.006$, the largest entry in absolute value is .002, and the smallest is .0006, so it is finding a non-zero estimate for the judgement vector.

\subsection{Zappos experiments}
\label{appendix:zappos}
We refer the reader to \cite{finegrained, semjitter} for the full details about the \texttt{UT Zappos50k} data set but provide relevant details here. The data can be found at \url{http://vision.cs.utexas.edu/projects/finegrained/utzap50k/}.

\subsection{Pairwise comparison data description}
The \texttt{UT Zappos50K} data set consists of pairwise comparisons on images of shoes and 960 extracted color and vision features for each shoe \cite{finegrained, semjitter}. Given images of two different shoes and an attribute from $\{$``open,'' ``pointy,'' ``sporty,'' ``comfort''$\}$, respondents were asked to pick which shoe exhibits the attribute more. The data consists of both easier, coarse questions, i.e. based on comfort, pick between a slipper or high-heel, and also harder, fine grained questions i.e. based on comfort, pick between two slippers. Each pairwise comparison is asked to 5 different people, and the confidence of each person's answer is also collected.

There are 2,863 unique pairwise comparisons involving 5,319 shoes for open, 2,700 unique pairwise comparisons involving 5,028 shoes for pointy, 2,766 unique pairwise comparisons involving 5,144 shoes for sporty, and 2,756 unique pairwise comparisons involving 5,129 shoes for comfort. For each attribute, $86\%$ of unique pairwise comparisons involve an item that is in no other pairwise comparison regarding that attribute. Also, for each attribute, nearly $93\%$ of items only appear in one pairwise comparison. In light of this, an algorithm like \cite{chen2016predicting} will likely not work well since (1) this model requires learning a set of parameters for each item and (2) the model does not work for unseen items, i.e., we must ensure that items in testing also appear in training to evaluate the model.

Furthermore, for each of the attributes, there are no triplets of items $(i,j,k)$ where pairwise comparison data has been collected on $i$ vs. $j$, $j$ vs. $k$, and $k$ vs. $i$. Therefore, we cannot even test if there are intransitivities in this data.

\subsection{Data pre-processing}
Respondents were given the option to declare a tie between two items. We do not train on any of these pairwise comparisons. To be clear, we use both the ``coarse" and ``fine-grained" comparisons during training. We standardize the features by subtracting the mean and dividing by the standard deviation, where we use the mean and standard deviation of the training set for each attribute since we train a model for each attribute.

\subsection{Experiment details}
\label{appendix:zapposHyperparameters}

The hyperparameters for the salient feature preference model with the top-$t$ selection function are $t$ and the $\ell_2$ regularization parameter $\mu$. The hyperparameter for FBTL is the $\ell_2$ regularization parameter $\mu$. For Ranking SVM, the only hyperparameter is $C$ which controls the penalty for violating the margin. We vary $t \in \{10*i: i \in [99]\}$ since there are 990 features. We vary $\mu$ and $C$ in $\{.000001, .00001, .0001, .001, .01, .1\}$. For RankNet, the hyperparameters are $\mu$ and the number of nodes in the hidden layer. We vary $\mu$ in $\{.05, .1, .15\}$ and the nodes in $\{50, 250, 500\}$. We choose these values of $\mu$ to try since on validation sets, it appeared that any value less than $.05$ was over fitting (train accuracy was in the 90\%s but validation accuracy was in the 70\%s) and values above .15 were not learning a good model (train accuracy was in the 60\%s). We only search over these hyperparameters due to time constraints. We use ten 70\% train, 15\% validation, and 15\% test split. 

\subsection{Best performing hyperparameters}
Because the pairwise comparisons are either ``coarse" or ``fine-grained," we pick the best hyperparameters based on the average of the pairwise comparison accuracy on the ``coarse" questions and the ``fine-grained" questions on the validation set. See Table \ref{table:zapposT} for statistics about the best performing $t$ for the salient feature preference model with the top-$t$ selection function on the validation set over 10 train/validation/test splits. See Tables \ref{table:salientMu}, \ref{table:BTLMu}, \ref{table:RankNetMu} for statistics about the best performing $\mu$ for the salient feature preference model, FBTL model, and RankNet on the validation set over the 10 train/validation/test splits. See Table \ref{table:RankingSVMC} for statistics about the best performing $C$ for Ranking SVM on the validation set over the over the 10 train/validation/test splits. See Table \ref{table:RankNetNodes} for the best performing number of nodes in the hidden layer on the validation set over the 10 splits. We also report the average pairwise accuracy, which has been defined in the main text, on the validation set for all algorithms in Table \ref{table:zapposValidation}. 

 \begin{table}
  \centering
  \caption{Statistics about the best performing $t$ for the salient feature preference model with the top-$t$ selection function on the validation set over 10 train/validation/test splits for \texttt{UT Zappos50k}.}
  \label{table:zapposT}

  \begin{tabular}{lllll}
    \toprule
    \cmidrule(r){1-2}
Attribute:                     & open & pointy & sporty & comfort \\
    \midrule
Min & 440 & 310 & 110 & 40\\
Max & 830 & 980 & 850 & 950 \\
Average &  663 & 614 & 550 & 563\\
Standard deviation & 150 & 198 & 238 & 305\\ 
  \end{tabular}
\end{table}

 \begin{table}
  \centering
  \caption{Statistics about the best performing $\mu$ for the salient feature preference model on the validation set over 10 train/validation/test splits for \texttt{UT Zappos50k}.}
  \label{table:salientMu}

  \begin{tabular}{lllll}
    \toprule
    \cmidrule(r){1-2}
Attribute:                     & open & pointy & sporty & comfort \\
    \midrule
Min & 1000 & 100 & 1000 & 10 \\ 
Max & 10000 & 100000 & 10000 & 10000 \\ 
Average & 4600.0 & 12520.0 & 5500.0 & 5311.0 \\ 
Standard deviation & 4409.08 & 29389.65 & 4500.0 & 4700.46
  \end{tabular}
\end{table}

 \begin{table}
  \centering
  \caption{Statistics about the best performing $\mu$ for FBTL on the validation set over 10 train/validation/test splits for \texttt{UT Zappos50k}.}
  \label{table:BTLMu}

  \begin{tabular}{lllll}
    \toprule
    \cmidrule(r){1-2}
Attribute:                     & open & pointy & sporty & comfort \\
    \midrule
Min &  1000 & 100 & 1000 & 10 \\ 
Max &  100000 & 100000 & 100000 & 100000 \\ 
Average & 15400 & 12520 & 17200 & 24211 \\ 
Standard deviation & 28517 & 29389 & 27827 & 38131
  \end{tabular}
\end{table}

 \begin{table}
  \centering
  \caption{Statistics about the best performing $C$ for Ranking SVM on the validation set over 10 train/validation/test splits for \texttt{UT Zappos50k}.}
  \label{table:RankingSVMC}

  \begin{tabular}{lllll}
    \toprule
    \cmidrule(r){1-2}
Attribute:                     & open & pointy & sporty & comfort \\
    \midrule
Min &  10000 & 1000 & 10000 & 100 \\ 
Max &  100000 & 1000000 & 1000000 & 1000000 \\ 
Average & 70000 & 124300 & 163000 & 144010  \\ 
Standard deviation & 42426 & 294261 & 281888 & 288619
  \end{tabular}
\end{table}

 \begin{table}
  \centering
  \caption{Statistics about the best performing $\mu$ for RankNet on the validation set over 10 train/validation/test splits for \texttt{UT Zappos50k}.}
  \label{table:RankNetMu}

  \begin{tabular}{lllll}
    \toprule
    \cmidrule(r){1-2}
Attribute:                     & open & pointy & sporty & comfort \\
    \midrule
Min &  .05 & .05 & .05 & .05 \\ 
Max &  .15 & .1 & .15 & .15 \\ 
Average &  .075 & .055 & .085 & .105  \\ 
Standard deviation & .033 & .015 & .039 & .041
  \end{tabular}
\end{table}

 \begin{table}
  \centering
  \caption{Statistics about the best performing number of nodes in the hidden layer for RankNet on the validation set over 10 train/validation/test splits for \texttt{UT Zappos50k}.}
  \label{table:RankNetNodes}

  \begin{tabular}{lllll}
    \toprule
    \cmidrule(r){1-2}
Attribute:                     & open & pointy & sporty & comfort \\
    \midrule
Min &  50 & 50 & 50 & 250 \\ 
Max &  500 & 500 & 250 & 500 \\ 
Average &  335 & 205 & 190 & 350  \\ 
Standard deviation & 178.95 & 201.84 & 91.65 & 122.47
  \end{tabular}
\end{table}

 \begin{table}
  \centering
  \caption{Average pairwise prediction accuracy over 10 train/validation/test splits on the validation sets by attribute for \texttt{UT Zappos50k}. $C$ stands for coarse and $F$ stands for fine grained. The number in parenthesis is the standard deviation.}
  \label{table:zapposValidation}

  \begin{tabular}{lllllllll}
    \toprule
    \cmidrule(r){1-2}
Model:                     & open-$C$ & pointy-$C$ & sporty-$C$ & comfort-$C$ & open-$F$ & pointy-$F$ & sporty-$F$ & comfort-$F$ \\
    \midrule
Salient features      & 0.75 (.01) & 0.8 (.01) & 0.79 (.02) & 0.77 (.03)  & 0.64 (.03) & 0.6 (.03) & 0.62 (.03) & 0.66 (.03) \\
FBTL & 0.75 (.02) & 0.8 (.01) & 0.79 (.01) & 0.77 (.02) & 0.63 (.03) & 0.59 (.03) & 0.6 (.02) & 0.62 (.03) \\
Ranking SVM      & 0.75 (.02) & 0.8 (.02) & 0.8 (.01) & 0.77 (.02)  & 0.62 (.04) & 0.59 (.03) & 0.6 (.02) & 0.62 (.04) \\
RankNet  &  0.75 (.02) & 0.78 (.03) & 0.78 (.01) & 0.76 (.02) & 0.67 (.03) & 0.61 (.04) & 0.61 (.02) & 0.64 (.03) \\
    \bottomrule
  \end{tabular}
\end{table}


\end{document}